\documentclass[10pt,twocolumn,letterpaper]{article}

\usepackage[pagenumbers]{cvpr}
\usepackage[dvipsnames]{xcolor}
\definecolor{cvprblue}{rgb}{0.21,0.49,0.74}
\usepackage[pagebackref,breaklinks,colorlinks,citecolor=cvprblue]{hyperref}
\usepackage{bm}
\usepackage{multirow}
\usepackage{tabularx}
\usepackage{colortbl}
\newcolumntype{C}{>{\centering\arraybackslash}X}
\newcolumntype{L}{>{\raggedright\arraybackslash}X}
\newcolumntype{R}{>{\raggedleft\arraybackslash}X}
\usepackage{algorithm,algorithmic}
\usepackage[accsupp]{axessibility}
\newcommand{\first}[1]{\multicolumn{1}{>{\columncolor[rgb]{1.0,0.75,0.75}}c}{#1}}
\newcommand{\second}[1]{\multicolumn{1}{>{\columncolor[rgb]{1.0,0.9,0.9}}c}{#1}}
\newcommand{\third}[1]{\multicolumn{1}{>{\columncolor[rgb]{1.0,0.95,0.95}}c}{#1}}
\makeatletter
\newcommand\footnoteref[1]{\protected@xdef\@thefnmark{\ref{#1}}\@footnotemark}
\makeatother

\renewcommand{\baselinestretch}{0.96}

\def\paperID{16665}
\def\confName{CVPR}
\def\confYear{2024}

\title{Improving Physics-Augmented Continuum Neural Radiance Field-Based\\ Geometry-Agnostic System Identification with Lagrangian Particle Optimization}

\author{Takuhiro Kaneko\\
  NTT Corporation}

\begin{document}
\maketitle

\begin{abstract}
  Geometry-agnostic system identification is a technique for identifying the geometry and physical properties of an object from video sequences without any geometric assumptions. Recently, physics-augmented continuum neural radiance fields (PAC-NeRF) has demonstrated promising results for this technique by utilizing a hybrid Eulerian--Lagrangian representation, in which the geometry is represented by the Eulerian grid representations of NeRF, the physics is described by a material point method (MPM), and they are connected via Lagrangian particles. However, a notable limitation of PAC-NeRF is that its performance is sensitive to the learning of the geometry from the first frames owing to its two-step optimization. First, the grid representations are optimized with the first frames of video sequences, and then the physical properties are optimized through video sequences utilizing the fixed first-frame grid representations. This limitation can be critical when learning of the geometric structure is difficult, for example, in a few-shot (sparse view) setting. To overcome this limitation, we propose Lagrangian particle optimization (LPO), in which the positions and features of particles are optimized through video sequences in Lagrangian space. This method allows for the optimization of the geometric structure across the entire video sequence within the physical constraints imposed by the MPM. The experimental results demonstrate that the LPO is useful for geometric correction and physical identification in sparse-view settings.\footnote{\label{foot:samples}The project page is available at \url{https://www.kecl.ntt.co.jp/people/kaneko.takuhiro/projects/lpo/}.}
\end{abstract}

\begin{figure}[t]
  \begin{center}
    \includegraphics[width=0.99\columnwidth]{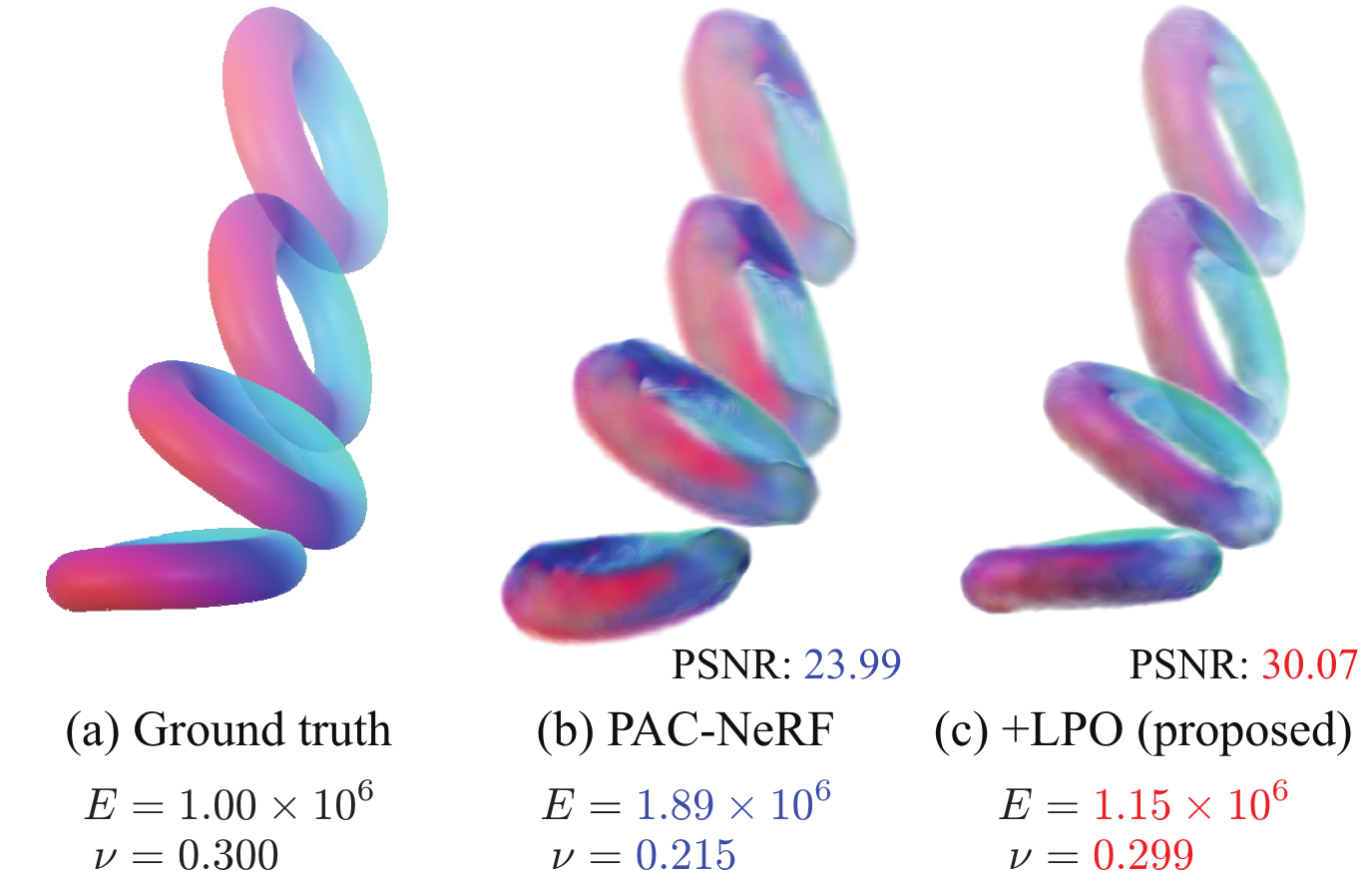}
  \end{center}
  \vspace{-5mm}
  \caption{Impact of the proposed Lagrangian particle optimization (LPO) in sparse-view geometric-agnostic system identification.
    We aim to identify the geometry and physical properties of an object from visual observations \textit{without any geometric assumptions} in \textit{severe} (e.g., sparse-view) settings.
    As shown in (b), a standard PAC-NeRF~\cite{XLiICLR2023} has difficulty learning the geometry in a sparse-view setting (particularly, when the number of views is three).
    Consequently, it also fails to estimate the physical properties (Young’s modulus $E$ and Poisson's ratio $\nu$).
    As shown in (c), LPO is useful for correcting this failure and succeeds in improving the identification of both geometry and physical properties.}
  \label{fig:teaser}
  \vspace{-4mm}
\end{figure}

\section{Introduction}
\label{sec:introduction}

System identification is a technique used to identify the geometry and physical properties of an object based on visual observations.
It has been actively studied in computer vision and physics because of its wide range of applications, including 3D reproduction, environmental understanding, and content creation.
To solve this problem in a realistic environment, it is important to adequately address both the \textit{extrinsic geometric structure} and the \textit{intrinsic physical properties}.
For example, to identify the geometry and physical properties of an object from the observation that it collides with the ground, it is necessary to understand not only the appearance and shape change in the \textit{geometry} but also other internal factors in \textit{physics} which are necessary to explain this phenomenon without any discrepancies over time.

Generally, this problem is ill-posed and challenging to resolve.
However, the emergence of powerful geometric representations by neural fields~\cite{YXieCGF2022}, 2D--3D connections by neural rendering~\cite{ATewariCGF2022}, and trainable physical simulations using differentiable physical simulators~\cite{CBanerjeeArXiv2023} provide clues for solving this problem.
For instance, previous studies~\cite{MJaquesICLR2020,KMJatavallabhulaICLR2021,RMaICLR2022} succeeded in estimating physical properties such as velocity, mass, friction, inertia, and elasticity directly from video sequences by combining differentiable physical simulators with differentiable renderers.

Although these studies have yielded promising results in terms of physical property estimation, their applicability is restricted by the assumption that the geometric structure of a scene is \textit{completely known}, which makes it difficult to apply them in practical scenarios.
To eliminate this assumption, \textit{geometry-agnostic system identification}, which is a technique for identifying the geometry and physical properties of an object from visual observations \textit{without any geometric assumptions}, has been studied.
Specifically, a pioneering model involves \textit{physics-augmented continuum neural radiance fields (PAC-NeRF)}~\cite{XLiICLR2023}, which is an extension of NeRF~\cite{BMildenhallECCV2020} that is enforced to follow the conservation laws of continuum mechanics.
PAC-NeRF obtains this functionality utilizing a hybrid Eulerian--Lagrangian representation, that is, the geometry (volume density and color fields) is represented with the Eulerian grid representations of NeRF~\cite{CSunCVPR2022}, which are transformed into Lagrangian particles in material space, and physical simulation is conducted on the particles using a material point method (MPM) (particularly, differentiable MPM (DiffMPM)~\cite{YHuICLR2020}).
Because all modules are differentiable, PAC-NeRF can be trained directly with multi-view video sequences and can optimize the geometry and physics without explicit supervision.

Although PAC-NeRF has enabled the tackling of new tasks, its notable limitation is that its performance is sensitive to the learning of the geometry from the first frames of the video sequences owing to its two-step optimization.
First, the Eulerian grid representations were trained with the first frames of the video sequences, and then the physical properties were optimized across the video sequences by utilizing the frozen first-frame grid representations.
This limitation makes it difficult to apply it to cases in which geometry learning is difficult, for example, in a few-shot (sparse-view) setting, as shown in Figure~\ref{fig:teaser}(b).

One possible solution is to train the Eulerian grid representations using all video sequences and not just the first frame.
However, a critical problem with this approach is that Eulerian grid representations cannot be trained with explicit physical constraints on the particles because they are optimized in Eulerian (i.e., world or grid) space and cannot reflect all physical phenomena occurring in Lagrangian (i.e., material or particle) space; for example, they cannot propagate gradients calculated for particle positions.

To solve this problem, we propose \textit{Lagrangian particle optimization (LPO)}, in which the positions and features (i.e., volume density and color) of the particles are optimized \textit{in Lagrangian space and not in Eulerian space}.
Contrary to the abovementioned possible solution, this method allows for the optimization of the geometric structure across video sequences with \textit{explicit} physical constraints on the particles imposed by the MPM.
Thus, the gradients calculated for the particle positions are reflected.

Furthermore, LPO is useful not only for geometry correction, but also for corrections to the physical identification because this task is closely related, that is, an accurate geometry is useful for accurately identifying the physical properties, and vice versa.
This motivated us to utilize the geometry corrected by LPO to reidentify the physical properties and propose \textit{iterative geometry--physics optimization} for gradually seeking the optimal states.
Figure~\ref{fig:teaser}(c) shows the effectiveness of this method.

We evaluated the effectiveness of LPO in sparse-view settings.
In particular, we first applied LPO to a pretrained PAC-NeRF and demonstrated that LPO is useful for correcting geometric structures through video sequences.
Subsequently, we employed LPO in iterative geometry--physics optimization and demonstrated that LPO contributes to improving the performance of physical identification.
Ablation and comparative studies were conducted to determine the importance of each component.

Our contributions can be summarized as follows:
\begin{itemize}
\item To improve the performance of \textit{geometry-agnostic system identification}, we propose \textit{LPO}, which optimizes the position and features of the particles \textit{in Lagrangian space} to optimize the geometric structures across video sequences within the physical constraints of an MPM.
\item We propose \textit{iterative geometry--physics optimization} to utilize the geometry corrected by LPO for reidentifying physical properties in an iterative manner.
\item We experimentally demonstrated that LPO is useful for geometric correction and physical identification in sparse-view settings.
  We also provide detailed analyses and extended results in the Appendices.
  Video samples are available at the \href{https://www.kecl.ntt.co.jp/people/kaneko.takuhiro/projects/lpo/}{project page}.\footnoteref{foot:samples}
\end{itemize}

\section{Related work}
\label{sec:related_work}

\textbf{Neural radiance fields (NeRF).}
Novel view synthesis is a fundamental problem in computer vision and graphics and has been addressed in a large body of research.
Recently, a substantial breakthrough has been achieved with the emergence of neural fields that represent a scene utilizing a coordinate-based network (e.g., a survey paper~\cite{YXieCGF2022}).
NeRF~\cite{BMildenhallECCV2020} is a representative variant of such neural fields and has attracted significant attention because of its geometric consistency and high-fidelity novel view synthesis.
Various studies have been conducted since the emergence of NeRF.
These can be categorized into three topics.
(1) Improvements to the image quality and expanding of the applicable settings, such as wild or few-shot settings (e.g.,~\cite{KZhangArXiv2020,RMartinCVPR2021,JBarronICCV2021,JBarronCVPR2022,BMildenhallCVPR2022,DVerbinCVPR2022,XChenCVPR2022,WHuICCV2023,TBarronICCV2023,MNiemeyerCVPR2022,JZhangNIPS2021,YWeiICCV2021,BRoessleCVPR2022,KDengCVPR2022,AJainICCV2021,JChibaneCVPR2021,AYuCVPR2021,AChenICCV2021,JYangCVPR2023}),
(2) speeding up the training or inference speed (e.g.,~\cite{TNeffCVF2021,MPiala3DV2021,DLindellCVPR2021,AKurzECCV2022,SGarbinICCV2021,PHedmanICCV2021,LLiuNeurIPS2020,LWuCVPR2022,CSunCVPR2022,AYu2021ICCV,SFridovichCVPR2022,SWizadwongsaCVPR2021,EChanCVPR2022,AChenECCV2022,TMullerTOG2022,THuCVPR2022,DRebainCVPR2021,CReiserICCV2021,TKanekoICCV2023,WHuICCV2023,TBarronICCV2023}), and
(3) incorporating other models or functionalities, such as generative models~\cite{IGoodfellowNIPS2014,YSongNeurIPS2019,JHoNeurIPS2020} (e.g.,~\cite{KSchwarzNeurIPS2020,EChanCVPR2021,MNiemeyerCVPR2021,MNiemeyer3DV2021,JGuICLR2022,EChanCVPR2022,YDengCVPR2022,YXueCVPR2022,TKanekoCVPR2022,ISkorokhodovNeurIPS2022,BPooleICLR2023,ERChanICCV2023}) and dynamics/physics (e.g.,~\cite{ZLiCVPR2021,APumarolaCVPR2021,GGafniCVPR2021,ETretschkICCV2021,KParkICCV2021,XGuoICCV2023,SShenECCV2022,JZhangSIGGRAPHAsia2022,MChuTOG2022,SGuanICML2022,XLiICLR2023}).
Among these, this study relates to the first in terms of its application to few-shot settings.
This study relates to the third in terms of seeking physics-informed models.
As studies on NeRF are benefiting from developments in adjacent categories, applying our ideas to other categories and models will be an interesting future research direction.

\smallskip\noindent
\textbf{NeRF with dynamics/physics.}
As mentioned above, NeRFs with dynamics/physics have been actively studied (e.g.,~\cite{ZLiCVPR2021,APumarolaCVPR2021,GGafniCVPR2021,ETretschkICCV2021,KParkICCV2021,XGuoICCV2023,SShenECCV2022,JZhangSIGGRAPHAsia2022,MChuTOG2022,SGuanICML2022,XLiICLR2023}).
These studies have one characteristic in common: they learned time-varying neural fields from video sequences.
These can be roughly categorized into two models.
(1) Non- (or weak) physics-informed models (e.g.,~\cite{ZLiCVPR2021,APumarolaCVPR2021,GGafniCVPR2021,ETretschkICCV2021,KParkICCV2021,XGuoICCV2023,SShenECCV2022,JZhangSIGGRAPHAsia2022}) and 
(2) physics-informed models (e.g.,~\cite{MChuTOG2022,SGuanICML2022,XLiICLR2023}).
The advantage of the first is that it can be applied to scenes or objects that are difficult to describe physically; however, its disadvantage is that it requires a large amount of training data to learn dynamics from scratch, and the learned representations are not necessarily interpretable because they are not based on physics.
The advantage of the second is that it can obtain interpretable representations based on physics; however, the disadvantage is that its application is limited to physically describable objects or scenes.
In particular, physics informed NeRF~\cite{MChuTOG2022} targets smoke scenes and does not address the boundary conditions; therefore, it cannot handle solid or contact materials.
NeuroFluid~\cite{SGuanICML2022} focuses on fluid dynamics grounding and solves it using an intuitive physics-based approach in which formal instruction in physics is not explicitly defined.
In contrast, PAC-NeRF~\cite{XLiICLR2023} is based on a principled and interpretable physical simulator and can be applied to various materials, including Newtonian/non-Newtonian fluids, elastic materials, plasticine, and granular media.
This study focused on the PAC-NeRF and attempted to widen its applicability while prioritizing its interpretability.
However, a Lagrangian particle representation is commonly utilized in physics (e.g., NeuroFluid~\cite{SGuanICML2022}\footnote{Note that NeuroFluid optimizes the \textit{transition probability of particles} under the assumption that the initial particle positions are known, that is, fixed.
  In contrast, LPO optimizes the \textit{initial particle positions} to correct the geometry estimation of the first frames.
  Therefore, NeuroFluid and LPO are complementary.}); therefore, applying our ideas, that is, LPO, to other models is an interesting research topic.

\smallskip\noindent
\textbf{NeRF for few-shot (sparse-view) settings.}
In practice, it is often laborious or impractical to collect multi-view images.
Recent studies~\cite{MNiemeyerCVPR2022,JZhangNIPS2021,YWeiICCV2021,BRoessleCVPR2022,KDengCVPR2022,AJainICCV2021,JChibaneCVPR2021,AYuCVPR2021,AChenICCV2021,SShenECCV2022,JYangCVPR2023,MChuTOG2022} addressed this problem by refining NeRF such that it could be applied to few-shot (sparse-view) settings.
These methods can be roughly divided into three approaches.
(1) Regularization using external models.
For example, normalizing flow~\cite{LDinhICLR2017}-based~\cite{MNiemeyerCVPR2022}, VGG~\cite{KSimonyanICLR2015}-based~\cite{JZhangNIPS2021}, depth-based~\cite{YWeiICCV2021,BRoessleCVPR2022,KDengCVPR2022}, and CLIP~\cite{ARadfordICML2021}-based~\cite{AJainICCV2021} regularizations have been proposed.
(2) Utilization of general and transferable models~\cite{JChibaneCVPR2021,AYuCVPR2021,AChenICCV2021,SShenECCV2022}.
They are trained using external datasets, and in several studies, they are finetuned for each scene.
(3) Introduction of advanced training methods, such as gradual training to prevent overfitting to sparse views, including frequency regularization~\cite{JYangCVPR2023} and layer-by-layer growing strategies~\cite{MChuTOG2022}.
Among these categories, the proposed LPO is categorized as the third one, that is, optimization is conducted for each scene, and external models and datasets are not required.
The main difference between the LPO and previous methods is that LPO attempts to optimize geometric structures through video sequences with the \textit{physical constraints} of an MPM.
However, previous studies have not sufficiently considered these constraints.
Owing to this difference in applications, the proposed method is complementary to, rather than competitive with, other methods.

\smallskip\noindent
\textbf{System identification of soft bodies.}
Soft bodies are not only high-dimensional, but also allow large deformations; therefore, it is challenging to identify their geometry and physical properties.
Typical methods can be categorized into three types.
(1) Gradient-free methods~\cite{BWangTOG2015,TTakahashiTOG2019},
(2) neural network-based methods~\cite{ASanchezICML2020,YLiICLR2019,ZXuRSS2019}, and
(3) differentiable physics-simulation-based methods~\cite{YQiaoNeurIPS2021,TDuTOG2021,MGeilingerTOG2020,EHeidenRSS2021,KMJatavallabhulaICLR2021,RMaICLR2022,HChenCVPR2022,XLiICLR2023}.
The strength of the methods in the first and second categories is that they are flexible; however, they have difficulty achieving a high accuracy owing to their black-box modeling.
Methods in the third category require sophisticated modeling to fill the gap between simulations and the real world; however, they have recently demonstrated promising results owing to advancements in differentiable physical simulators.
The third category can be further divided into two subcategories.
(i) Methods that assume that watertight geometric mesh representations of objects are available~\cite{YQiaoNeurIPS2021,TDuTOG2021,MGeilingerTOG2020,EHeidenRSS2021,KMJatavallabhulaICLR2021,RMaICLR2022} and
(2) methods that do not assume this~\cite{HChenCVPR2022,XLiICLR2023}.
Among these methods, we focused on the final category, PAC-NeRF~\cite{XLiICLR2023}, which prioritizes general and fast characteristics.
Specifically, PAC-NeRF adopts an MPM that can be applied to various materials, including fluids~\cite{CJiangTOG2015}, sand~\cite{GKlarTOG2016}, foam~\cite{YYueTOG2015}, and elastic objects~\cite{HChenCVPR2022}.
In particular, PAC-NeRF utilizes DiffMPM~\cite{YHuICLR2020} to construct a fully differentiable simulation rendering system.
This study is based on this advancement.
In particular, sensitivity to geometry learning from the first frames is one of the bottlenecks of PAC-NeRF.
Therefore, this study focused on alleviating this bottleneck and attempted to widen its applicability to severe (e.g., sparse-view) settings.

\begin{figure*}[t]
  \begin{center}
    \includegraphics[width=\textwidth]{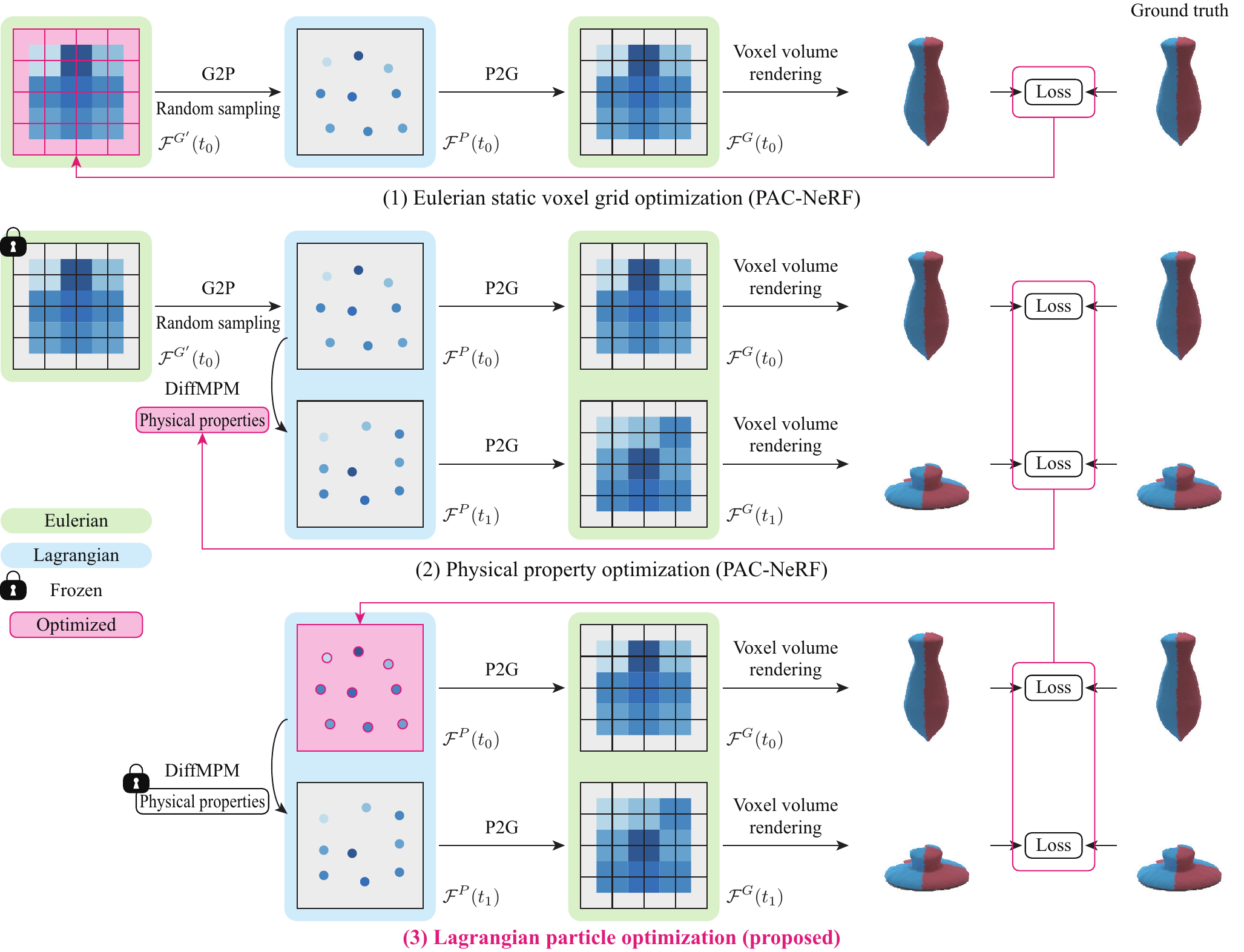}
  \end{center}
  \vspace{-4mm}
  \caption{Optimization  pipelines of PAC-NeRF (1)(2) and the proposed LPO (3).}
  \label{fig:pipelines}
  \vspace{-2mm}
\end{figure*}

\section{Method}
\label{sec:method}

In this section, we first define the problem statement (Section~\ref{subsec:problem}) and then briefly review PAC-NeRF~\cite{XLiICLR2023}, upon which our method is built (Section~\ref{subsec:preliminary}).
Subsequently, we explain the main idea of the proposed method, that is, geometric correction using LPO (Section~\ref{subsec:geometry}), and then introduce its application to physical identification (Section~\ref{subsec:physics}).

\subsection{Problem statement}
\label{subsec:problem}

Given a set of multi-view video sequences, our objectives are as follows: (1) to recover geometric representations\footnote{Here, both appearance and shape are treated as part of the geometric representations.
  We attempted to recover both.} of the target dynamic object (particularly continuum materials) through video sequences and (2) identify its physical properties.
More formally, the training data comprise ground truth color observations, that is, $\hat{\mathbf{C}}(\mathbf{r}, t)$.
Here, $\mathbf{r}(s) \in \mathbb{R}^3$ is a camera ray defined as $\mathbf{r}(s) = \mathbf{o} + s \mathbf{d}$, where $\mathbf{o} \in \mathbb{R}^3$ is the camera origin, $\mathbf{d} \in \mathbb{S}^2$ is the view direction, and $s \in [s_n, s_f]$ is a distance from the camera origin.
During training, $\mathbf{r}$ is sampled from a set of ray collections in the training data, that is, $\hat{\mathcal{R}}$.
$t  \in \mathbb{R}^{+}$ is a time variable, and during training, it is sampled from $\{ t_0, \dots, t_{N-1} \}$, where $N$ is the number of frames.
Given these training data, we attempt to (1) predict the color observation $\mathbf{C}(\mathbf{r}, t)$ that can recover $\hat{\mathbf{C}}(\mathbf{r}, t)$ from the training data and those for novel views, and (2) identify the physical properties.
In particular, to widen its applicability, we addressed a scenario in which input views are sparse, that is, $\hat{\mathcal{R}}$ is limited.

\subsection{Preliminary: PAC-NeRF}
\label{subsec:preliminary}

To solve the abovementioned problem, PAC-NeRF employs three core components: a continuum NeRF, particle-grid interconverters, and a Lagrangian field.

\smallskip\noindent
\textbf{Continuum NeRF.}
PAC-NeRF extends a standard (static) NeRF to a continuum NeRF to address the dynamics of continuum materials.
To achieve this, a standard NeRF is first extended to a dynamic NeRF~\cite{APumarolaCVPR2021} for dynamic scenarios.
In a dynamic NeRF, the volume density and color fields for point $\mathbf{x}$ and view direction $\mathbf{d}$ are defined as $\sigma (\mathbf{x}, t)$ and $\mathbf{c}(\mathbf{x}, \mathbf{d}, t)$, respectively, where the time variable $t$ is introduced to represent the dynamics.
The color of each pixel $\mathbf{C}(\mathbf{r}, t)$ is calculated using volume rendering~\cite{NMaxTVCG1995}
\begin{flalign}
  \label{eq:volume_rendering}
  \mathbf{C}(\mathbf{r}, t) & = \int_{s_n}^{s_f} T(s, t) \sigma (\mathbf{r}(s), t) \mathbf{c}(\mathbf{r}(s), \mathbf{d}, t) ds
                              \nonumber \\
                            & \:\:\:\:\:\:\:\:\:\:\:\:\:\: + \mathbf{c}_{bg} T(s_f, t),
  \\
  T(s, t) & = \exp \left( - \int_{s_n}^s \sigma(\mathbf{r}(u), t) du \right),
\end{flalign}
where $\mathbf{c}_{bg}$ denotes the background color.
The model is trained with pixel-wise loss.
\begin{flalign}
  \label{eq:pixel_loss}
  \mathcal{L}_{pixel} = \frac{1}{N} \sum_{i=0}^{N-1} \frac{1}{|\hat{\mathcal{R}}|} \sum_{\mathbf{r} \in \hat{\mathcal{R}}} \| \mathbf{C}(\mathbf{r}, t_i) - \hat{\mathbf{C}}(\mathbf{r}, t_i) \|_2^2.
\end{flalign}
Furthermore, a dynamic NeRF is extended to a continuum NeRF to represent continuum materials.
To achieve this, conservation laws for the velocity field are imposed on the volume density and color fields.
\begin{flalign}
  \label{eq:conservation_laws}
  \frac{D \sigma}{D t} = 0,\; \frac{D \mathbf{c}}{D t} = \mathbf{0},
\end{flalign}
whereas the material derivative of an arbitrary time-dependent field $\phi(\mathbf{x}, t)$ is defined as $\frac{D \phi}{D t} = \frac{\partial \phi}{\partial t} + \mathbf{v} \cdot \nabla \phi$.
Here, $\mathbf{v}$ is the velocity field and follows momentum conservation for continuum materials:
\begin{flalign}
  \label{eq:momentum_conservation}
  \rho \frac{D \mathbf{v}}{D t} = \nabla \cdot \bm{T} + \rho \mathbf{g},
\end{flalign}
where $\rho$ is the physical density field, $\bm{T}$ is the internal Cauchy stress tensor, and $\mathbf{g}$ is the acceleration of gravity.
This equation is solved using DiffMPM~\cite{YHuICLR2020}.

\smallskip\noindent
\textbf{Particle-grid interconverters.}
As shown in Figure~\ref{fig:pipelines}(1)(2), in PAC-NeRF, a physical simulation is conducted in Lagrangian particle space using DiffMPM~\cite{YHuICLR2020}, whereas neural rendering is performed in Eulerian grid space with a discretized voxel-based NeRF~\cite{CSunCVPR2022}.
To bridge these two spaces, grid-to-particle (G2P) and particle-to-grid (P2G) conversions are conducted.
\begin{flalign}
  \label{eq:p2g_g2p}
  \mathcal{F}_p^P \approx \sum_i w_{ip} \mathcal{F}_i^G,\;
  \mathcal{F}_i^G \approx \frac{\sum_p w_{ip}\mathcal{F}_p^P}{\sum_p w_{ip}},
\end{flalign}
where $\mathcal{F}_{*}^{*}$ is a field (e.g., volume density or color field); $P$ and $G$ denote particle and grid views, respectively; $p$ and $i$ indicate the index of the particle and grid node, respectively; and $w_{ip}$ is the weight of the trilinear shape function defined on $i$ and evaluated at $p$.

\smallskip\noindent
\textbf{Lagrangian field.}
As shown in Figure~\ref{fig:pipelines}(1), during training, the Eulerian voxel field $\mathcal{F}^{G'}(t_0)$ is optimized for the first frames of the video sequences.
Subsequently, this field is converted into a Lagrangian particle field $\mathcal{F}^P(t_0)$ using G2P, in which the positions of the particles are determined by random sampling around voxel grids.
The geometric shape is represented in Lagrangian space by removing the particles whose alpha values are small, that is, the remaining particles represent the shape of the object.
As shown in Figure~\ref{fig:pipelines}(2), the particle field in the next step, that is, $\mathcal{F}^P(t_1)$, where $t_1 = t_0 + \delta t$ and $\delta t$ is the duration of the simulation time step, is calculated based on $\mathcal{F}^P(t_0)$ using DiffMPM~\cite{YHuICLR2020}.
After this process, $\mathcal{F}^P(t_1)$ is converted to the Eulerian voxel field $\mathcal{F}^G(t_1)$ using P2G, and pixels are rendered based on this field using voxel-based volume rendering~\cite{CSunCVPR2022}, in which the volume density and color of a point at position $\mathbf{x}$ are rendered as follows:
\begin{flalign}
  \label{eq:density_grid}
  \sigma(\mathbf{x}, t) & = \mathrm{softplus} (\mathrm{interp}(\mathbf{x}, \sigma^G)),
  \\
  \label{eq:color_grid}
  \mathbf{c}(\mathbf{x}, \mathbf{d}, t) & = \mathrm{MLP} (\mathrm{interp}(\mathbf{x}, \mathbf{c}^G), \mathbf{d}),
\end{flalign}
where $\sigma^G$ and $\mathbf{c}^G$ denote the volume density and color fields, respectively, discretized on fixed voxel grids, and $\mathrm{interp}$ denotes trilinear interpolation.

\subsection{Geometric correction with LPO}
\label{subsec:geometry}

As explained above, in PAC-NeRF, the Eulerian voxel fields $\mathcal{F}^{G'}(t_0)$ are optimized for the first frames of the video sequences (Figure~\ref{fig:pipelines}(1)).
These are fixed while optimizing the physical properties using video sequences (Figure~\ref{fig:pipelines}(2)).
This approach is problematic when the learning of the Eulerian voxel field from the first frames is difficult, for example, in sparse-view settings, because failure of this learning cannot be corrected after the process.
This also means that in dynamic scenes, information over time is useful for covering a lack of information owing to sparse views; however, this approach cannot utilize this information to optimize the volume density and color.

One possible solution is to train $\mathcal{F}^{G’}(t_0)$ with the entire video sequence and not just the first frame.
However, in this approach, the optimization targets are limited to the volume density and color of the grids; therefore, it is difficult to reflect the physics-based optimization that occurs in Lagrangian particle space.\footnote{More precisely, as explained in Section~\ref{subsec:preliminary}, particles are randomly sampled around equally spaced voxel grids, and the Eulerian voxel fields (precisely, alpha values calculated from them) are utilized to mask them with a non-differentiable way.
  Consequently, gradients calculated for the particle positions cannot propagate to $\mathcal{F}^{G'}(t_0)$.}
This can cause excessive modification of the geometry beyond physical constraints.

Alternatively, we developed \textit{Lagrangian particle optimization (LPO)}, in which the geometric structure is optimized \textit{not in Eulerian grid space but in Lagrangian particle space} (Figure~\ref{fig:pipelines}(3)).
More formally, in the Eulerian voxel grid optimization, volume density and color fields of the grids are optimized for the \textit{fixed} voxel grid.
In contrast, in LPO, not only the volume density and color fields of the particles, that is, $\sigma^P$ and $\mathbf{c}^P$, but also the particle position, that is $\mathbf{x}^P$, are \textit{optimized}.\footnote{Note that the values of $\sigma^P$ and $\mathbf{c}^P$ are changed in training but they are time-invariant, i.e., have the same values across sequences.
  Therefore, the conservation laws (Equation~(\ref{eq:conservation_laws})) are preserved.}
Because $\mathbf{x}^P$ is defined in Lagrangian space, it can reflect the physical constraints of the MPM.
Consequently, we can optimize the geometric structures within the physical allowance.

\begin{algorithm}[t]
  \caption{Iterative geometry--physics optimization}
  \label{alg:iterative_optimization}
  \begin{algorithmic}[1]
    \renewcommand{\algorithmicrequire}{\textbf{Input:}}
    \renewcommand{\algorithmicensure}{\textbf{Output:}}
    \renewcommand{\COMMENT}[2][12.5em]{
      \leavevmode\hfill\makebox[#1][l]{$\triangleright$~{\scriptsize #2}}}
    \REQUIRE Ground-truth color observations $\hat{\mathbf{C}}(\hat{\mathbf{r}}, t)$\\
    $\hat{\mathbf{r}} \in \hat{\mathcal{R}}$: Rays from observed viewpoints
    \\ $t \in \{ t_0, \dots, t_{N-1} \}$: Time
    \ENSURE Physical properties
    \FOR {$i = 1$ to $R$}
    \STATE // (1) Eulerian static voxel grid optimization
    \IF{$i = 1$}
    \STATE Optimize voxel grids with $\hat{\mathbf{C}}(\hat{\mathbf{r}}, t_0)$
    \ELSE
    \STATE Optimize voxel grids with $\hat{\mathbf{C}}(\hat{\mathbf{r}}, t_0)$ and $\breve{\mathbf{C}}(\breve{\mathbf{r}}, t_0)$
    \ENDIF
    \STATE // (2) Physical property optimization
    \STATE Optimize physical properties with $\hat{\mathbf{C}}(\hat{\mathbf{r}}, t)$
    \STATE // (3) Lagrangian particle optimization
    \STATE Optimize particles with $\hat{\mathbf{C}}(\hat{\mathbf{r}}, t)$
    \STATE // (4) Color prediction for unobserved views $\breve{\mathbf{r}} \in \breve{\mathcal{R}}$
    \STATE Predict $\breve{\mathbf{C}}(\breve{\mathbf{r}}, t_0)$
    \ENDFOR
  \end{algorithmic}
\end{algorithm}

\subsection{Physical identification with LPO}
\label{subsec:physics}

Identifying both the geometry and physical properties from limited observations is ill-posed and challenging because there is a chicken-and-egg relationship between the geometric structure and the physical properties, that is, accurate geometry estimation is necessary for accurate physical identification and vice versa.
Considering that particle-based geometric correction is highly challenging when there is a large gap between the ground truth and the predicted pixels owing to the high-dimensional nature (e.g., in extreme cases, particles can diverge), in practice, we apply LPO after physical property optimization, as shown in Figure~\ref{fig:pipelines}.

However, an interesting question is \textit{whether the corrected geometric structures can be utilized to improve the identification of the physical properties}.
We developed \textit{iterative geometry--physics optimization} (Algorithm~\ref{alg:iterative_optimization}) to answer this question.
As described above, we first conduct PAC-NeRF optimization ((1) and (2)) and then apply LPO (3).
Subsequently, we render the images in the first frame based on the updated $\sigma^P$, $\mathbf{c}^P$, and $\mathbf{x}^P$(4).
We denote them as $\breve{\mathbf{C}}(\breve{\mathbf{r}}, t_0)$, where $\breve{\mathbf{r}} \in \breve{\mathcal{R}}$ and $\breve{\mathcal{R}}$ is designed such that it covers the missing area in $\hat{\mathcal{R}}$ in the training data.
We retrain PAC-NeRF from scratch using the combination of the original $\hat{\mathbf{C}}(\hat{\mathbf{r}}, t_0)$ and synthesized $\breve{\mathbf{C}}(\breve{\mathbf{r}}, t_0)$ as training data.
Specifically, to alleviate the negative effect caused by incomplete geometry correction, which can occur at the beginning of the iterative calculation, we only utilize $\breve{\mathbf{C}}(\breve{\mathbf{r}}, t_0)$ for the Eulerian static voxel grid optimization and do not utilize it to optimize the physical properties and particles.

Another possible solution is to update the voxel grids by applying P2G to the updated $\sigma^P$, $\mathbf{c}^P$, and $\mathbf{x}^P$ instead of (4) and (1).
However, we found that the repeated utilization of G2P and P2G tends to produce artifacts (such as those typically occurring in the erosion process) owing to their approximate nature (Equation~(\ref{eq:p2g_g2p})).\footnote{A similar phenomenon was observed in the original study on PAC-NeRF~\cite{XLiICLR2023}.
  Based on this observation, they rendered an image in the first frame by utilizing the voxel grids obtained after the P2G and G2P conversions, that is, $\mathcal{F}^G(t_0)$, instead of the original voxel grids $\mathcal{F}^{G'}(t_0)$ to make the rendering conditions the same as those in other frames.}
Therefore, the full recalculation approach described above was adopted.

\begin{table*}[t]
  \centering
  \scriptsize
  \setlength{\tabcolsep}{0pt}
  \begin{tabularx}{\textwidth}{cCCCCCCCCCCC}
    \toprule
    & PAC-NeRF & PAC-NeRF-3v & +LPO & +LPO-F & +LPO-P & +GO & \!PAC-NeRF-3v$^\dag$\! & +LPO & +LPO-F & +LPO-P & +GO
    \\ \cmidrule(lr){1-1} \cmidrule(lr){2-2} \cmidrule(lr){3-7} \cmidrule(lr){8-12} 
    PSNR$\uparrow$
    & 35.99 & 27.39 & \first{29.22} & 28.22 & \second{28.89} & 28.33 & 28.47 & \first{30.11} & 29.31 & \second{29.87} & 29.39
    \\
    SSIM$\uparrow$
    & 0.989 & 0.978 & \first{0.980} & \second{0.979} & \first{0.980} & \second{0.979} & 0.980 & \first{0.982} & \second{0.981} & \second{0.981} & \second{0.981}
    \\
    LPIPS$\downarrow$
    & 0.020 & 0.034 & \first{0.032} & \second{0.033} & \first{0.032} & \second{0.033} & 0.033 & \first{0.031} & \second{0.032} & \first{0.031} & \first{0.031}
    \\ \bottomrule
  \end{tabularx}
  \vspace{-3mm}
  \caption{Comparison of PSNR$\uparrow$, SSIM$\uparrow$, and LPIPS$\downarrow$ on \textbf{geometric correction}.
    The scores for each scene are provided in Table~\ref{tab:comparison_geometry_ex} in Appendix~\ref{sec:analyses_main_experiments}.
    The qualitative comparisons are provided in Figures~\ref{fig:results_newtonian}--\ref{fig:results_sand} in Appendix~\ref{sec:qualitative_results}.}
  \label{tab:comparison_geometry}
  \vspace{-3mm}
\end{table*}

\section{Experiments}
\label{sec:experiments}

\subsection{Experimental setup}
\label{subsec:experimental_setup}

Two experiments were conducted to investigate the effectiveness of the proposed LPO.
First, we evaluated the effectiveness of LPO for geometric correction (Section~\ref{subsec:experiment_geometry}) and then we investigated the utility of LPO for physical identification (Section~\ref{subsec:experiment_physics}).
The main results of these experiments are presented here, and the detailed analyses, extended results, and implementation details are provided in the Appendices.
Video samples are available at the \href{https://www.kecl.ntt.co.jp/people/kaneko.takuhiro/projects/lpo/}{project page}.\footnoteref{foot:samples}
In this section, we provide the experimental setup common to both experiments and other setups in subsequent sections.

\smallskip\noindent
\textbf{Dataset.}
We investigated the benchmark performance on the dataset provided by the original study on PAC-NeRF~\cite{XLiICLR2023}.\footnote{Note that although only one dataset was used, this dataset is useful for assessing the versatility because it includes a wide variety of materials, surpassing the scope of previous studies (e.g., elastic objects only in~\cite{HChenCVPR2022}).}
This dataset comprised nine scenes and various continuum materials, including Newtonian fluids (\textit{Droplet} and \textit{Letter} with fluid viscosity $\mu$ and bulk modulus $\kappa$), non-Newtonian fluids (\textit{Cream} and \textit{Toothpaste} with shear modulus $\mu$, bulk modulus $\kappa$, yield stress $\tau_Y$, and plastic viscosity $\eta$), elastic materials (\textit{Torus} and \textit{Bird} with Young's modulus $E$ and Poisson's ratio $\nu$), plasticine (\textit{Playdoh} and \textit{Cat} with $E$, $\nu$, and $\tau_Y$), and granular media (\textit{Trophy} with friction angle $\theta_{fric}$).
In each scene, the objects fall freely under the influence of gravity and undergo collisions.
The ground-truth simulation data were generated using the MLS-MPM framework~\cite{YHuTOG2018}.
A photorealistic simulation engine rendered objects under diverse environmental lighting conditions and ground textures.
Each scene was captured from 11 viewpoints with cameras evenly spaced on the upper hemisphere, including the object.
To evaluate our method under sparse-view settings, three views were used for training and the remaining eight views were used for testing.
To show the robustness to the view settings, we provide the results for the other view settings in Appendix~\ref{sec:view_experiments}.

\smallskip\noindent
\textbf{Assumptions and preprocessing.}
For a fair comparison, we follow the assumptions and preprocessing used in PAC-NeRF~\cite{XLiICLR2023}.
It was assumed that the intrinsic and extrinsic parameters of the set of static cameras were previously known.
Moreover, it was assumed that collision objects, such as the ground plane, were previously known.
As mentioned in~\cite{XLiICLR2023}, this is not difficult to estimate from the observed images.
For preprocessing, a video matting framework~\cite{SLinCVPR2021} was applied to remove static background objects and focus the rendering on the target object.

\subsection{Evaluation of geometric correction}
\label{subsec:experiment_geometry}

\smallskip\noindent
\textbf{Compared models.}
In the first experiment, we investigated the effectiveness of LPO on the geometric correction (i.e., the method described in Section~\ref{subsec:geometry}).
Three models were used as the baseline.
(I) \textit{PAC-NeRF} was the same as that in the original~\cite{XLiICLR2023} and was trained with all views, including the eight views that were used for testing.
We examined this model to determine the upper bound of its performance.
(II) \textit{PAC-NeRF-3v} was trained using three views.
The training settings were the same as those for (I) except that the number of views varied.
(III) \textit{PAC-NeRF-3v$^\dag$} was an improved variant of (II) for sparse-view settings.
An interesting question is whether LPO, which is a few-shot learning method for \textit{dynamic} scenes, can be used with other few-shot learning methods, such as those for \textit{static} scenes.
To answer this question, we adjusted the Eulerian static voxel grid optimization (Figure~\ref{fig:pipelines}(1)) of PAC-NeRF-3v, such that it became robust to sparse views.
Specifically, we scheduled a surface regularizer to reduce unexpected artifacts, introduced a view-invariant pixel-wise loss to compensate for the lack of views, and adjusted the training length to prevent overfitting.
Details are provided in Appendix~\ref{sec:pacnerf3vdag_detail}.\footnote{\label{foot:previous_few_shot_nerf}In the preliminary experiments, we examined the previous representative few-shot learning methods (e.g., DietNeRF~\cite{AJainICCV2021} and FreeNeRF~\cite{JYangCVPR2023}).
  However, we found that they were less stable than PAC-NeRF-3v, possibly because in our experimental settings, the number of views was small (three) despite the wide range of the views (upper hemisphere), and explicit voxel representations were more useful than the fully implicit representation in \cite{AJainICCV2021,JYangCVPR2023}.
  However, this study and previous studies are complementary.
  Further investigations will be important in future work.}
We applied LPO to both (II) and (III) to investigate the effect of the initial Eulerian static voxel grid optimization.
Hereafter, we denote these variants by \textit{+LPO}.
Furthermore, three comparative and ablation models were evaluated to determine the importance of each component.
(i) \textit{+LPO-F} and (ii) \textit{+LPO-P} are variants of LPO, in which only the features and positions of the particles are optimized.
(iii) \textit{+GO} optimized Eulerian voxel grids through the entire video sequence instead of using Lagrangian particles.
(i)--(iii) are used as alternatives to \textit{+LPO}.
These variants were applied to both (II) and (III).

\smallskip\noindent
\textbf{Evaluation metrics.}
We evaluated the performance of the geometric correction using metrics commonly used to assess the performance of novel view synthesis in NeRF studies: the peak signal-to-noise ratio (\textit{PSNR}), structural similarity index measure (\textit{SSIM})~\cite{ZWangTIP2004}, and learned perceptual image patch similarity (\textit{LPIPS})~\cite{RZhangCVPR2018}.
In particular, we calculated the scores using all test data over time.

\smallskip\noindent
\textbf{Results.}
The results are summarized in Table~\ref{tab:comparison_geometry}.
Our findings are threefold.
\textit{(1) PAC-NeRF-3v/3v$^\dag$ vs. +LPO.}
+LPO improved both baselines in terms of all metrics.
In particular, the utility of +LPO for PAC-NeRF-3v$^\dag$ indicates an improvement in the Eulerian static voxel grid optimization and that of LPO are complementary.
\textit{(2) +LPO vs. +LPO-F/P.}
We found that +LPO was superior or comparable to +LPO-F/P.
This is because the feature and position optimizations are complementary.
Intuitively, feature optimization is useful for correcting features of the \textit{appearance} that are invisible in the first frame but visible after the object has moved.
Similarly, position optimization is effective for correcting features of the \textit{shape} that are invisible in the first frame but visible after the object has moved.
In severe (e.g., sparse-view) settings, the utilization of both is important.
\textit{(3) +LPO vs. +GO.}
+LPO outperforms +GO, possibly because +LPO can optimize the geometry adequately within the physical constraints of the MPM, whereas +GO cannot because of the absence of an explicit physical constraint.

\begin{table*}[t]
  \centering
  \scriptsize
  \setlength{\tabcolsep}{2pt}
  \begin{tabularx}{\textwidth}{CCCCCCCCCCCCCCC}
    \toprule
    & & PAC- & PAC- & \multirow{2}{*}{+LPO$^4$} & \multirow{2}{*}{+LPO-F$^4$} & \multirow{2}{*}{+LPO-P$^4$} & \multirow{2}{*}{+GO$^4$} & \multirow{2}{*}{+None$^4$} & PAC- & \multirow{2}{*}{+LPO$^4$} & \multirow{2}{*}{+LPO-F$^4$} & \multirow{2}{*}{+LPO-P$^4$} & \multirow{2}{*}{+GO$^4$} & \multirow{2}{*}{+None$^4$}
    \\
    & & NeRF & NeRF-3v & & & & & & \!NeRF-3v$^\dag$\!
    \\ \cmidrule(lr){1-1} \cmidrule(lr){2-2} \cmidrule(lr){3-3} \cmidrule(lr){4-9} \cmidrule(lr){10-15}
    \multirow{2}{*}{Droplet}
    & $\log_{10}(\mu)$
      & 0.039 & 0.140 & \second{0.112} & \second{0.112} & 0.119 & \first{0.111} & 0.131 & 0.136 & 0.082 & \second{0.068} & \first{0.067} & 0.082 & 0.090
    \\
    & $\log_{10}(\kappa)$
      & 0.017 & \first{1.285} & 1.628 & 1.638 & \second{1.466} & 1.675 & 1.684 & 1.263 & \first{0.106} & 1.139 & \second{0.520} & 1.520 & 1.656
    \\ \cmidrule(lr){1-1} \cmidrule(lr){2-2} \cmidrule(lr){3-3} \cmidrule(lr){4-9} \cmidrule(lr){10-15}
    \multirow{2}{*}{Letter}
    & $\log_{10}(\mu)$
      & 0.041 & 0.674 & \first{0.015} & \second{0.026} & 0.079 & 0.048 & 0.530 & 0.379 & \second{0.010} & 0.087 & \first{0.007} & 0.118 & 0.436
    \\
    & $\log_{10}(\kappa)$
      & 0.039 & 6.772 & \first{0.174} & \second{0.411} & 1.040 & 0.865 & 6.083 & 5.229 & \second{0.060} & \first{0.027} & 0.127 & 0.138 & 5.060
    \\ \cmidrule(lr){1-1} \cmidrule(lr){2-2} \cmidrule(lr){3-3} \cmidrule(lr){4-9} \cmidrule(lr){10-15}
    \multirow{4}{*}{Cream}
    & $\log_{10}(\mu)$
      & 0.090 & 0.311 & \second{0.178} & \first{0.163} & 0.234 & 0.183 & 0.251 & 0.179 & 0.100 & \second{0.073} & \first{0.065} & 0.092 & 0.076
    \\
    & $\log_{10}(\kappa)$
      & 0.132 & 0.215 & 0.158 & 0.249 & \first{0.027} & \second{0.028} & 0.244 & 0.336 & \first{0.121} & 0.197 & \second{0.142} & 0.193 & 0.339
    \\
    & $\log_{10}(\tau_Y)$
      & 0.007 & 0.014 & \first{0.004} & 0.030 & \second{0.005} & 0.006 & 0.025 & 0.009 & 0.006 & \first{0.001} & \second{0.004} & 0.010 & 0.008
    \\
    & $\log_{10}(\eta)$
      & 0.015 & 0.281 & 0.183 & 0.256 & \second{0.083} & \first{0.061} & 0.252 & 0.195 & \first{0.033} & 0.059 & 0.079 & 0.096 & \second{0.051}
    \\ \cmidrule(lr){1-1} \cmidrule(lr){2-2} \cmidrule(lr){3-3} \cmidrule(lr){4-9} \cmidrule(lr){10-15}
    \multirow{4}{*}{Toothpaste}
    & $\log_{10}(\mu)$
      & 0.026 & 1.891 & \first{0.109} & 0.156 & \second{0.119} & 0.164 & 0.215 & 0.252 & \second{0.031} & 0.201 & \first{0.005} & 0.252 & 0.138
    \\
    & $\log_{10}(\kappa)$
      & 0.247 & 1.580 & \second{0.601} & 1.356 & \first{0.597} & 0.648 & 0.729 & 1.436 & 0.673 & 0.630 & 0.899 & \first{0.590} & \second{0.596}
    \\
    & $\log_{10}(\tau_Y)$
      & 0.066 & 0.201 & 0.114 & 0.191 & \second{0.075} & 0.250 & \first{0.061} & 0.199 & 0.093 & \first{0.064} & 0.117 & 0.149 & \second{0.054}
    \\
    & $\log_{10}(\eta)$
      & 0.013 & 0.373 & \first{0.003} & 0.267 & 0.013 & \second{0.007} & 0.047 & 0.212 & 0.009 & 0.016 & \first{0.005} & \second{0.007} & 0.042
    \\ \cmidrule(lr){1-1} \cmidrule(lr){2-2} \cmidrule(lr){3-3} \cmidrule(lr){4-9} \cmidrule(lr){10-15}
    \multirow{2}{*}{Torus}
    & $\log_{10}(E)$
      & 0.019 & 0.277 & 0.061 & \first{0.008} & 0.083 & \second{0.010} & 0.054 & 0.074 & 0.036 & \second{0.026} & 0.039 & \first{0.015} & 0.074
    \\
    & $\nu$
      & 0.023 & 0.085 & \first{0.001} & 0.120 & \second{0.031} & 0.074 & 0.316 & 0.131 & \first{0.007} & 0.040 & \second{0.033} & 0.050 & 0.129
    \\ \cmidrule(lr){1-1} \cmidrule(lr){2-2} \cmidrule(lr){3-3} \cmidrule(lr){4-9} \cmidrule(lr){10-15}
    \multirow{2}{*}{Bird}
    & $\log_{10}(E)$
      & 0.013 & 0.449 & \first{0.067} & 0.240 & \second{0.074} & 0.313 & 0.246 & 0.123 & \first{0.027} & 0.186 & \second{0.039} & 0.231 & 0.296
    \\
    & $\nu$
      & 0.029 & 0.102 & \first{0.001} & 0.372 & \second{0.075} & 0.365 & 0.581 & 0.141 & \second{0.047} & 0.072 & \first{0.008} & 0.132 & 0.145
    \\ \cmidrule(lr){1-1} \cmidrule(lr){2-2} \cmidrule(lr){3-3} \cmidrule(lr){4-9} \cmidrule(lr){10-15}
    \multirow{3}{*}{Playdoh}
    & $\log_{10}(E)$
      & 0.286 & 0.290 & \first{0.116} & 0.580 & \second{0.120} & 0.304 & 0.170 & 0.521 & \first{0.133} & 0.474 & 2.121 & \second{0.232} & 2.148
    \\
    & $\log_{10}(\tau_Y)$
      & 0.038 & 0.283 & \second{0.165} & \first{0.027} & 0.214 & 0.196 & 0.237 & \second{0.110} & 0.173 & \first{0.079} & 1.179 & 0.197 & 0.967
    \\
    & $\nu$
      & 0.076 & 0.495 & \second{0.111} & 0.173 & \first{0.109} & 0.248 & 0.128 & 0.212 & \first{0.063} & 0.133 & 0.110 & 0.127 & \second{0.106}
    \\ \cmidrule(lr){1-1} \cmidrule(lr){2-2} \cmidrule(lr){3-3} \cmidrule(lr){4-9} \cmidrule(lr){10-15}
    \multirow{3}{*}{Cat}
    & $\log_{10}(E)$
      & 0.855 & 1.301 & \first{0.973} & 1.068 & 1.071 & 1.127 & \second{1.066} & 1.192 & \second{0.706} & \first{0.534} & 0.712 & 0.783 & 0.793
    \\
    & $\log_{10}(\tau_Y)$
      & 0.026 & 0.120 & \second{0.107} & 0.117 & 0.113 & \first{0.086} & 0.112 & 0.084 & \second{0.067} & \first{0.014} & 0.076 & 0.069 & 0.072
    \\
    & $\nu$
      & 0.027 & 0.044 & \first{0.004} & \second{0.036} & 0.069 & 0.190 & 0.079 & 0.118 & \first{0.003} & 0.027 & \second{0.007} & 0.028 & 0.063
    \\ \cmidrule(lr){1-1} \cmidrule(lr){2-2} \cmidrule(lr){3-3} \cmidrule(lr){4-9} \cmidrule(lr){10-15}
    \multirow{1}{*}{Trophy}
    & \!\!\!$\theta_{fric}$ [rad]\!\!\!
      & 0.048 & \second{0.053} & 0.055 & \first{0.051} & 0.056 & 0.057 & 0.054 & \first{0.030} & \second{0.039} & 0.041 & 0.043 & 0.044 & \second{0.039}
    \\ \bottomrule
  \end{tabularx}
  \vspace{-3mm}
  \caption{Comparison of the absolute differences between the ground-truth and the estimated physical properties on \textbf{physical identification}.
    The smaller the values, the better was the performance.
    The values of the physical properties (i.e., not the absolute differences) are provided in Tables~\ref{tab:comparison_physics_value1} and \ref{tab:comparison_physics_value2} in Appendix~\ref{sec:analyses_main_experiments}.
    The qualitative comparisons are presented in Figures~\ref{fig:results_newtonian}--\ref{fig:examples_feat} in Appendix~\ref{sec:qualitative_results}.}
  \label{tab:comparison_physics}
  \vspace{-3mm}
\end{table*}

\subsection{Evaluation of physical identification}
\label{subsec:experiment_physics}

\smallskip\noindent
\textbf{Compared models.}
In the second experiment, we verified the usefulness of LPO for physical identification (i.e., the method described in Section~\ref{subsec:physics}).
In addition to the models described in Section~\ref{subsec:experiment_geometry}, we examined \textit{+None}, for which iterative optimization (Algorithm~\ref{alg:iterative_optimization}) was conducted without geometric correction (i.e., Step (3) was skipped).
This model was used to investigate the importance of geometric correction.
For a fair comparison, we set the number of iterations (i.e., $R$ in Algorithm~\ref{alg:iterative_optimization}) to four for all models.
We use the superscript $4$ (e.g., +LPO$^4$) to specify this.

\smallskip\noindent
\textbf{Evaluation metrics.}
To evaluate the performance of the physical identification, we measured the absolute distance between the ground truth and the estimated physical properties.
For an easy comparison, we calculated these distances after adjusting the scale (i.e., either a logarithmic scale or a linear scale) following the study of PAC-NeRF~\cite{XLiICLR2023}.
The smaller the values, the better was the performance.

\smallskip\noindent
\textbf{Results.}
The results are summarized in Table~\ref{tab:comparison_physics}.
Our findings are fourfold.
\textit{(1) PAC-NeRF-3v/3v$^\dag$ vs. +LPO$^4$.}
+LPO$^4$ improved the physical identification of PAC-NeRF-3v/3v$^\dag$ in most cases.
In particular, the effectiveness of +LPO$^4$ for PAC-NeRF-3v$^\dag$ indicates that the improvement in the Eulerian static voxel grid optimization and that of LPO are complementary for physical identification.
\textit{(2) +LPO$^4$ vs. +LPO-F$^4$/P$^4$.}
In some cases, superiority depends on the physical properties because the physical properties interact with each other, and finding the optimal balance is difficult.
However, we found that +LPO-F$^4$/P$^4$ sometimes had obvious difficulties (e.g., PAC-NeRF-3v+LPO-F$^4$ on Bird and PAC-NeRF-3v$^\dag$+LPO-P$^4$ on Playdoh), whereas +LPO$^4$ exhibited good stability.
We consider that the joint optimization of the features and positions is useful for tackling difficult situations.
\textit{(3) +LPO$^4$ vs. +GO$^4$.}
+GO$^4$ experienced difficulties in some cases (e.g., on Bird).
+LPO$^4$ exhibited a more stable performance.
\textit{(4) +LPO$^4$ vs. +None$^4$.}
+None$^4$ sometimes worsened the performance.
The results indicated that simple iterations were insufficient and that geometric correction was essential.

\smallskip\noindent
\textbf{Applications.}
As mentioned above, there is a chicken-and-egg relationship between the geometric structure and physical properties.
An interesting question is \textit{whether the corrected physical properties can be used to reestimate the geometry} (the opposite problem).
To answer this question, we investigated the geometry identification performance after physical identification.
The results are summarized in Table~\ref{tab:comparison_geometry4}.
These results indicate that accurate physical identification is useful for improving the geometry identification performance.

\begin{table}[t]
  \centering
  \scriptsize
  \setlength{\tabcolsep}{2pt}
  \vspace{2mm}
  \begin{tabularx}{\columnwidth}{cCCCCCC}
    \toprule
    & PAC-NeRF-3v & \multirow{2}{*}{+LPO} & \multirow{2}{*}{+LPO$^4$} & PAC-NeRF-3v$^\dag$ & \multirow{2}{*}{+LPO} & \multirow{2}{*}{+LPO$^4$}
    \\ \cmidrule(lr){1-1} \cmidrule(lr){2-4} \cmidrule(lr){5-7}
    PSNR$\uparrow$
    & 27.39 & \second{29.22} & \first{29.65} & 28.47 & \second{30.11} & \first{30.34}
    \\
    SSIM$\uparrow$
    & 0.978 & \second{0.980} & \first{0.982} & 0.980 & \second{0.982} & \first{0.983}
    \\
    LPIPS$\downarrow$
    & 0.034 & \second{0.032} & \first{0.030} & 0.033 & \second{0.031} & \first{0.029}
    \\ \bottomrule
  \end{tabularx}
  \vspace{-3mm}
  \caption{Comparison of PSNR$\uparrow$, SSIM$\uparrow$, and LPIPS$\downarrow$ on \textbf{geometric recorrection}.
    The scores for each scene are provided in Table~\ref{tab:comparison_geometry4_ex} in Appendix~\ref{sec:analyses_main_experiments}.
    The qualitative comparisons are provided in Figures~\ref{fig:results_newtonian}--\ref{fig:examples_feat} in Appendix~\ref{sec:qualitative_results}.}
  \label{tab:comparison_geometry4}
  \vspace{-3mm}
\end{table}

\section{Discussion}
\label{sec:discussion}

Based on the above experiments, we demonstrated promising results for geometric correction and physical identification.
However, our methods have some limitations.
(1) Our methods require a longer training time due to the introduction of LPO (Section~\ref{subsec:geometry})\footnote{The calculation time of LPO (Figure~\ref{fig:pipelines}(3)) is almost identical to that of the main process of physical property optimization (Figure~\ref{fig:pipelines}(2)) because the forward and backward processes are identical with different optimization targets.
  Similarly, the calculation times for +LPO-F, +LPO-P, and +GO were almost identical to those for +LPO, indicating that the improvement was attributable to the ingenuity of the algorithm and not to an increase in the calculation cost.
  The computation times are discussed in detail in Appendix~\ref{subsec:computation_times}.} and the iterative algorithm (Section~\ref{subsec:physics}).\footnote{The total computation time increases linearly when running Algorithm~\ref{alg:iterative_optimization} repeatedly but is adjustable under a quality-and-time trade-off as discussed in Appendix~\ref{subsec:impact_iterations}.}
However, it is not trivial to obtain robustness to sparse views only by increasing the training time because overfitting is a typical factor that causes learning to fail.
(2) LPO is sensitive to the state before applying LPO (e.g., either PAC-NeRF-3v or -3v$^\dag$) because solving geometry-agnostic system identification in sparse-view settings is ill-posed and challenging.
We believe that the advancements in previous few-shot learning methods (Section~\ref{sec:related_work}) and the newly introduced few-shot learning method with physical constraints (LPO) will solve this problem.

\section{Conclusion}
\label{sec:conclusion}

We proposed LPO to improve PAC-NeRF-based geometric-agnostic system identification.
Our core idea is to optimize the geometry \textit{not in Eulerian space but in Lagrangian space}, utilizing the particles to directly reflect the physical constraints of an MPM.
The results demonstrate that LPO is useful for both geometric correction and physical identification.
Although we focused on PAC-NeRF while prioritizing its high generality, Lagrangian particles are commonly employed in physics-informed models (e.g., several studies discussed in Section~\ref{sec:related_work}).
We expect that our ideas can be utilized in other models or tasks.

\clearpage
\renewcommand{\baselinestretch}{1}
{
  \small
  \bibliographystyle{ieeenat_fullname}
  \bibliography{refs}
}

\clearpage
\appendix

\section{Details of PAC-NeRF-3v$^\dag$}
\label{sec:pacnerf3vdag_detail}

\subsection{Method}
\label{subsec:pacnerf3vdag_method}

In the experiments (Section~\ref{sec:experiments}), we used PAC-NeRF-3v$^\dag$ as the stronger baseline.
PAC-NeRF-3v$^\dag$ is an improved variant of PAC-NeRF-3v for sparse-view settings in which Eulerian static voxel grid optimization (Figure~\ref{fig:pipelines}(a)) is improved.
We examined this model to determine whether the proposed LPO, a few-shot learning method for \textit{dynamic} scenes, can be combined with other few-shot learning methods, such as those for \textit{static} scenes.\footnote{As described in the footnote of the main text,\footnoteref{foot:previous_few_shot_nerf} in preliminary experiments, we found that previous representative few-shot learning methods (for example, DietNeRF~\cite{AJainICCV2021} and FreeNeRF~\cite{JYangCVPR2023}) were less stable than the standard PAC-NeRF-3v.
  This is possible because, in our experimental settings, the number of views was small (three) despite the wide range of views (upper hemisphere), and explicit voxel representations were more effective than the fully implicit representation in \cite{AJainICCV2021,JYangCVPR2023}.
  Therefore, we used an improved variant that we developed.
  However, this study and previous studies are not competitive but complementary.
  Therefore, further investigation is important.}
This appendix explains the details of the model.
PAC-NeRF-3v$^\dag$ adopts three modifications: scheduling a surface regularizer, introducing view-invariant pixel-wise loss, and adjusting the training length.

\smallskip\noindent
\textbf{Scheduling of surface regularizer.}
In the original PAC-NeRF~\cite{XLiICLR2023}, a surface regularizer $\mathcal{L}_{surf}$ is applied to regularize the volume density field
\begin{flalign}
  \label{eq:surface_regularizer}
  \mathcal{L}_{surf} = \sum_p \mathrm{clamp}(\alpha_p, 10^{-4}, 10^{-1}) \left( \frac{\Delta x}{2} \right)^2,
\end{flalign}
where $\alpha_p$ indicates the alpha value of a particle with length $1$ and is calculated as $\alpha_p = 1 - \exp(- \mathrm{softplus}(\sigma_p))$, where $\sigma_p$ denotes the volume density of the particle.
This regularizer minimizes the total surface area, making the spread of the particles more compact and tightening their shapes.
Consequently, the quality of the reconstructed geometries improved~\cite{XLiICLR2023}.

In a preliminary experiment, we found that this regularizer was also effective in eliminating unexpected masses that tend to appear in places where there are few clues owing to the lack of views.
However, we also found that this regularizer has a side effect: it removes necessary components when it is too strong.
Based on these observations, we scheduled this regularizer.
\begin{enumerate}
\item At the beginning of the training, the weight of $\mathcal{L}_{surf}$ is initialized to a default value of PAC-NeRF~\cite{XLiICLR2023}.
\item From the beginning of the training, the weight of $\mathcal{L}_{surf}$ is gradually increased for a certain period of time.
\item After a certain period, the weight of $\mathcal{L}_{surf}$ is gradually decreased until it reaches the default value.
\end{enumerate}
We do not impose a large weight on $\mathcal{L}_{surf}$ from the beginning of the training (Step~1) because it can eliminate all particles, leading to learning ``none'' object.
We decreased the weight of $\mathcal{L}_{surf}$ in Step 3 to alleviate the negative effects caused by the introduction of strong regularization.

\smallskip\noindent
\textit{Implementation details.}
In the experiments, we doubled the weight of $\mathcal{L}_{surf}$ in Step 2 every 100 iterations until the weight reached eight times its default value.
In Step 3, we halved the weight of $\mathcal{L}_{surf}$ each time the resolution of voxel grids was scaled.
This halving process was conducted three times; therefore, the weight of $\mathcal{L}_{surf}$ returned to the default value when all halving processes were finished.

\smallskip\noindent
\textbf{View-invariant pixel-wise loss.}
In sparse view settings, it is challenging to distinguish view-dependent from view-independent factors because there are few clues.
Specifically, in NeRF~\cite{BMildenhallECCV2020} (mainly, voxel-based NeRF~\cite{CSunCVPR2022}), the view-dependent and view-independent factors (i.e., colors) are represented by a multilayer perceptron (MLP), which additionally receives a view direction $\mathbf{d}$, and color fields $\mathbf{c}^G$ that do not receive $\mathbf{d}$, respectively (Equation~(\ref{eq:color_grid})).
In sparse-view settings, dividing the roles between the MLP and $\mathbf{c}^G$ is not trivial.
In extreme cases, the MLP can overfit specific views in the training data (in such cases, $\mathbf{c}^G$ no longer plays an essential role in representing colors), making it difficult to represent colors in novel views.
To alleviate this difficulty, we introduce a view-invariant (VI) pixel-wise loss $\mathcal{L}_{pixel}^{VI}$, which is a variant of the pixel-wise loss $\mathcal{L}_{pixel}$ (Equation~(\ref{eq:pixel_loss})) where the color of a sample on a ray, i.e., $\mathbf{c}$, is calculated by the following equation instead of Equation~(\ref{eq:color_grid})
\begin{flalign}
  \label{eq:random_color_grid}
  \tilde{\mathbf{c}}(\mathbf{x}, \tilde{\mathbf{d}}, t) & = \mathrm{MLP} (\mathrm{interp}(\mathbf{x}, \mathbf{c}^G), \tilde{\mathbf{d}}),
\end{flalign}
where $\tilde{\mathbf{d}} \in \mathbb{S}^2$ denotes the view direction randomly sampled from a unit sphere.
Here, we use $\tilde{\mathbf{c}}$ to denote $\mathbf{c}$ and distinguish it from the original $\mathbf{c}$.
This loss encourages MLP and $\mathbf{c}^G$ to capture the colors of the training images independent of the viewing direction.
Consequently, the model makes it possible to avoid extreme cases (i.e., it mitigates the MLP to overfit specific views in the training data and prevents $\mathbf{c}^G$ from losing a role) and provides some colors even for novel views.
The total pixel-wise loss $\mathcal{L}_{pixel}^{\dag}$ is given by
\begin{flalign}
  \label{eq:pixel_loss_dag}
  \mathcal{L}_{pixel}^{\dag} = \mathcal{L}_{pixel} + \lambda \mathcal{L}_{pixel}^{VI},
\end{flalign}
where $\lambda$ is a hyperparameter balancing the two losses.
During training, $\mathcal{L}_{pixel}^{\dag}$ was used instead of $\mathcal{L}_{pixel}$.

\smallskip\noindent
\textit{Implementation details.}
We set $\lambda = 0.1$ in the experiments.

\begin{table*}[t]
  \centering
  \scriptsize
  \setlength{\tabcolsep}{1pt}
  \begin{tabularx}{\textwidth}{ccCCCCC}
    \toprule
    & & (a) PAC-NeRF-3v & (b) PAC-NeRF-3v$^\dag$ & (c) w/o scheduling of $\mathcal{L}_{surf}$ & (d) w/o $\mathcal{L}_{pixel}^{VI}$ & (e) w/o adjustment of TL
    \\ \cmidrule(lr){1-1} \cmidrule(lr){2-2} \cmidrule(lr){3-3} \cmidrule(lr){4-7} 
    \multirow{3}{*}{Droplet}
    & PSNR$\uparrow$
      & 24.42 & \first{26.83} & \second{26.51} & 26.41 & 24.88
    \\
    & SSIM$\uparrow$
      & 0.975 & \first{0.981} & \first{0.981} & \second{0.979} & 0.977
    \\
    & LPIPS$\downarrow$
      & 0.048 & \first{0.045} & 0.050 & 0.048 & \second{0.047}
    \\ \cmidrule(lr){1-1} \cmidrule(lr){2-2} \cmidrule(lr){3-3} \cmidrule(lr){4-7} 
    \multirow{3}{*}{Letter}
    & PSNR$\uparrow$
      & \second{28.56} & \first{29.37} & 28.52 & 27.75 & 27.62
    \\
    & SSIM$\uparrow$
      & \second{0.979} & \first{0.981} & 0.978 & 0.977 & 0.977
    \\
    & LPIPS$\downarrow$
      & \second{0.032} & \first{0.030} & 0.036 & \second{0.032} & 0.034
    \\ \cmidrule(lr){1-1} \cmidrule(lr){2-2} \cmidrule(lr){3-3} \cmidrule(lr){4-7} 
    \multirow{3}{*}{Cream}
    & PSNR$\uparrow$
      & 26.10 & \first{27.09} & \second{26.51} & 25.73 & 25.67
    \\
    & SSIM$\uparrow$
      & \second{0.982} & \first{0.983} & 0.981 & 0.981 & 0.980
    \\
    & LPIPS$\downarrow$
      & 0.027 & \second{0.025} & \second{0.027} & 0.029 & \first{0.024}
    \\ \cmidrule(lr){1-1} \cmidrule(lr){2-2} \cmidrule(lr){3-3} \cmidrule(lr){4-7} 
    \multirow{3}{*}{Toothpaste}
    & PSNR$\uparrow$
      & 28.70 & \first{31.77} & \second{31.70} & 31.60 & 30.76
    \\
    & SSIM$\uparrow$
      & 0.988 & \first{0.991} & \first{0.991} & \first{0.991} & \second{0.990}
    \\
    & LPIPS$\downarrow$
      & 0.013 & \second{0.012} & \first{0.011} & \first{0.011} & \second{0.012}
    \\ \cmidrule(lr){1-1} \cmidrule(lr){2-2} \cmidrule(lr){3-3} \cmidrule(lr){4-7} 
    \multirow{3}{*}{Torus}
    & PSNR$\uparrow$
      & 25.17 & \first{26.77} & 25.33 & 26.15 & \second{26.55}
    \\
    & SSIM$\uparrow$
      & 0.972 & \first{0.974} & 0.972 & \second{0.973} & 0.972
    \\
    & LPIPS$\downarrow$
      & 0.047 & \first{0.043} & 0.047 & \first{0.043} & \second{0.044}
    \\ \cmidrule(lr){1-1} \cmidrule(lr){2-2} \cmidrule(lr){3-3} \cmidrule(lr){4-7} 
    \multirow{3}{*}{Bird}
    & PSNR$\uparrow$
      & 26.29 & \second{27.64} & \first{28.82} & 27.45 & 26.26
    \\
    & SSIM$\uparrow$
      & 0.979 & \first{0.982} & \first{0.982} & \first{0.982} & \second{0.981}
    \\
    & LPIPS$\downarrow$
      & 0.034 & \first{0.029} & 0.033 & 0.031 & \second{0.030}
    \\ \cmidrule(lr){1-1} \cmidrule(lr){2-2} \cmidrule(lr){3-3} \cmidrule(lr){4-7} 
    \multirow{3}{*}{Playdoh}
    & PSNR$\uparrow$
      & 27.87 & \first{30.05} & 29.03 & \second{29.77} & 28.32
    \\
    & SSIM$\uparrow$
      & 0.976 & \first{0.981} & \first{0.981} & \first{0.981} & \second{0.978}
    \\
    & LPIPS$\downarrow$
      & 0.046 & \first{0.042} & 0.046 & \second{0.043} & \second{0.043}
    \\ \cmidrule(lr){1-1} \cmidrule(lr){2-2} \cmidrule(lr){3-3} \cmidrule(lr){4-7} 
    \multirow{3}{*}{Cat}
    & PSNR$\uparrow$
      & 30.82 & \first{31.93} & \second{31.90} & 30.36 & 31.75
    \\
    & SSIM$\uparrow$
      & 0.987 & \first{0.989} & \first{0.989} & 0.987 & 0.988
    \\
    & LPIPS$\downarrow$
      & \second{0.035} & \second{0.035} & \second{0.035} & 0.038 & 0.036
    \\ \cmidrule(lr){1-1} \cmidrule(lr){2-2} \cmidrule(lr){3-3} \cmidrule(lr){4-7} 
    \multirow{3}{*}{Trophy}
    & PSNR$\uparrow$
      & 28.93 & 29.05 & \first{29.23} & \second{29.17} & 28.85
    \\
    & SSIM$\uparrow$
      & \second{0.963} & \second{0.963} & \first{0.964} & \first{0.964} & \first{0.964}
    \\
    & LPIPS$\downarrow$
      & \second{0.038} & 0.039 & 0.039 & \second{0.038} & \first{0.036}
    \\ \cmidrule[0.1em](lr){1-1} \cmidrule[0.1em](lr){2-2} \cmidrule[0.1em](lr){3-3} \cmidrule[0.1em](lr){4-7} 
    \multirow{3}{*}{\texttt{Average}}
    & PSNR$\uparrow$
      & 27.43 & \first{28.94} & \second{28.62} & 28.27 & 27.85
    \\
    & SSIM$\uparrow$
      & 0.978 & \first{0.980} & \first{0.980} & \second{0.979} & \second{0.979}
    \\
    & LPIPS$\downarrow$
      & 0.036 & \first{0.033} & 0.036 & 0.035 & \second{0.034}
    \\ \bottomrule
  \end{tabularx}
  \vspace{-2mm}
  \caption{Comparison of PSNR$\uparrow$, SSIM$\uparrow$, and LPIPS$\downarrow$ on \textbf{static voxel grid optimization}.
    The scores are calculated for the first frames of the video sequences in the test set.
    Here, PAC-NeRF-3V$^\dag$ (b), an improved variant of PAC-NeRF-3v, is compared with the original PAC-NeRF-3v (a) and the ablated models, including PAC-NeRF-3V$^\dag$ without scheduling of $\mathcal{L}_{surf}$ (c), that without $\mathcal{L}_{pixel}^{VI}$ (d), and that without adjustment of TL (e).
    PAC-NeRF-3V$^\dag$ (b) achieved the 20 best and five second-best scores among 27 evaluation items.}
  \label{tab:comparison_static}
\end{table*}

\smallskip\noindent
\textbf{Adjustment of training length.}
As discussed in previous studies~\cite{JYangCVPR2023,MChuTOG2022}, one of the factors that make learning difficult in few-shot settings is overfitting to sparse views in the training data, causing a loss of generalization ability.
In preliminary experiments, we observed a similar tendency in our settings.
Based on this observation, we adjusted the training length (TL).
In particular, the number of iterations was reduced.
Despite its simplicity, we empirically found that this solution works well in severe view settings (e.g. when the number of views is three).

\smallskip\noindent
\textit{Implementation details.}
In the experiments, we reduced the number of iterations to one-third; the default number (6000) was reduced to 2000.
Based on this change, we reduced the timing of scaling the resolution of voxel grids by one-third.

\subsection{Experiments}
\label{subsec:pacnerf3vdag_experiments}

Ablation studies were conducted to confirm the importance of each modification.
Specifically, we compared \textit{PAC-NeRF-3v$^\dag$} with three ablated models: \textit{PAC-NeRF-3v$^\dag$ without scheduling of $\mathcal{L}_{surf}$}, \textit{PAC-NeRF-3v$^\dag$ without $\mathcal{L}_{pixel}^{VI}$}, and \textit{PAC-NeRF-3v$^\dag$ without adjustment of TL}.
We also examined the original \textit{PAC-NeRF-3v}, that is, PAC-NeRF-3v$^\dag$ without all three modifications.

\smallskip\noindent
\textbf{Results.}
Table~\ref{tab:comparison_static} summarizes the results.
We found that PAC-NeRF-3v$^\dag$ achieved the best or second-best scores in most cases (20 best scores and five second-best scores among the 27 evaluation items).
Consequently, PAC-NeRF-3v$^\dag$ achieved the best average scores in terms of all metrics.
These results indicate the importance of each modification.

\begin{table*}[t]
  \centering
  \scriptsize
  \setlength{\tabcolsep}{2pt}
  \begin{tabularx}{\textwidth}{CCCCCCCCCCCCC}
    \toprule
    & & PAC- & PAC- & \multirow{2}{*}{+LPO} & \multirow{2}{*}{+LPO-F} & \multirow{2}{*}{+LPO-P} & \multirow{2}{*}{+GO} & PAC- & \multirow{2}{*}{+LPO} & \multirow{2}{*}{+LPO-F} & \multirow{2}{*}{+LPO-P} & \multirow{2}{*}{+GO}
    \\
    & & NeRF & NeRF-3v & & & & & NeRF-3v$^\dag$
    \\ \cmidrule(lr){1-1} \cmidrule(lr){2-2} \cmidrule(lr){3-3} \cmidrule(lr){4-8} \cmidrule(lr){9-13}
    \multirow{3}{*}{Droplet}
    & PSNR$\uparrow$
      & 35.30 & 25.42 & \first{27.56} & 27.05 & 27.26 & \second{27.55} & 26.40 & \first{28.18} & 27.36 & \second{28.00} & 27.36
    \\
    & SSIM$\uparrow$
      & 0.990 & 0.975 & \first{0.978} & \second{0.977} & \second{0.977} & \first{0.978} & 0.978 & \first{0.980} & \second{0.979} & \first{0.980} & \second{0.979}
    \\
    & LPIPS$\downarrow$
      & 0.029 & 0.047 & \first{0.043} & 0.045 & \second{0.044} & \second{0.044} & 0.046 & \first{0.044} & \second{0.045} & \first{0.044} & 0.046
    \\ \cmidrule(lr){1-1} \cmidrule(lr){2-2} \cmidrule(lr){3-3} \cmidrule(lr){4-8} \cmidrule(lr){9-13}
    \multirow{3}{*}{Letter}
    & PSNR$\uparrow$
      & 36.01 & 28.94 & \first{29.99} & 29.53 & \second{29.90} & 29.56 & 29.59 & \first{30.44} & 30.23 & \second{30.29} & 30.09
    \\
    & SSIM$\uparrow$
      & 0.991 & \second{0.981} & \first{0.982} & \first{0.982} & \first{0.982} & \first{0.982} & \first{0.983} & \second{0.982} & \first{0.983} & \second{0.982} & \first{0.983}
    \\
    & LPIPS$\downarrow$
      & 0.012 & \second{0.028} & 0.029 & \second{0.028} & 0.029 & \first{0.027} & 0.028 & 0.029 & \second{0.027} & 0.030 & \first{0.026}
    \\ \cmidrule(lr){1-1} \cmidrule(lr){2-2} \cmidrule(lr){3-3} \cmidrule(lr){4-8} \cmidrule(lr){9-13}
    \multirow{3}{*}{Cream}
    & PSNR$\uparrow$
      & 36.70 & 26.61 & \first{28.48} & 27.79 & 27.80 & \second{28.39} & 27.43 & \first{28.82} & 28.34 & 28.50 & \second{28.73}
    \\
    & SSIM$\uparrow$
      & 0.993 & \second{0.983} & \second{0.983} & \first{0.984} & \second{0.983} & \first{0.984} & \second{0.983} & \first{0.984} & \first{0.984} & \first{0.984} & \first{0.984}
    \\
    & LPIPS$\downarrow$
      & 0.014 & 0.026 & \first{0.023} & \second{0.024} & 0.025 & \first{0.023} & \second{0.024} & \first{0.023} & \first{0.023} & \second{0.024} & \first{0.023}
    \\ \cmidrule(lr){1-1} \cmidrule(lr){2-2} \cmidrule(lr){3-3} \cmidrule(lr){4-8} \cmidrule(lr){9-13}
    \multirow{3}{*}{Toothpaste}
    & PSNR$\uparrow$
      & 39.46 & 29.23 & \second{31.27} & 30.90 & 30.66 & \first{31.63} & 31.72 & \second{33.54} & 33.15 & 33.15 & \first{33.77}
    \\
    & SSIM$\uparrow$
      & 0.996 & 0.991 & 0.991 & 0.991 & 0.991 & 0.991 & 0.993 & 0.993 & 0.993 & 0.993 & 0.993
    \\
    & LPIPS$\downarrow$
      & 0.006 & \first{0.010} & \first{0.010} & \first{0.010} & \second{0.011} & \first{0.010} & \second{0.010} & \second{0.010} & \second{0.010} & \second{0.010} & \first{0.009}
    \\ \cmidrule(lr){1-1} \cmidrule(lr){2-2} \cmidrule(lr){3-3} \cmidrule(lr){4-8} \cmidrule(lr){9-13}
    \multirow{3}{*}{Torus}
    & PSNR$\uparrow$
      & 34.54 & 23.99 & \first{28.91} & 25.22 & \second{27.82} & 24.92 & 26.65 & \first{30.05} & 29.18 & 29.03 & \second{29.27}
    \\
    & SSIM$\uparrow$
      & 0.988 & 0.970 & \first{0.978} & 0.971 & \second{0.977} & 0.970 & 0.974 & \first{0.980} & \second{0.978} & \second{0.978} & \second{0.978}
    \\
    & LPIPS$\downarrow$
      & 0.026 & 0.048 & \first{0.038} & 0.046 & \second{0.040} & 0.045 & 0.040 & \first{0.034} & \second{0.036} & 0.037 & \second{0.036}
    \\ \cmidrule(lr){1-1} \cmidrule(lr){2-2} \cmidrule(lr){3-3} \cmidrule(lr){4-8} \cmidrule(lr){9-13}
    \multirow{3}{*}{Bird}
    & PSNR$\uparrow$
      & 35.55 & 24.82 & \first{27.20} & 25.47 & \second{27.18} & 25.30 & 24.91 & \first{28.51} & 25.73 & \second{28.43} & 26.18
    \\
    & SSIM$\uparrow$
      & 0.991 & \second{0.978} & \first{0.980} & \second{0.978} & \first{0.980} & \second{0.978} & 0.979 & \first{0.983} & \second{0.980} & \first{0.983} & \second{0.980}
    \\
    & LPIPS$\downarrow$
      & 0.019 & 0.036 & \first{0.032} & \second{0.035} & \first{0.032} & 0.036 & 0.036 & \first{0.028} & \second{0.033} & \first{0.028} & \second{0.033}
    \\ \cmidrule(lr){1-1} \cmidrule(lr){2-2} \cmidrule(lr){3-3} \cmidrule(lr){4-8} \cmidrule(lr){9-13}
    \multirow{3}{*}{Playdoh}
    & PSNR$\uparrow$
      & 36.43 & 27.70 & \second{28.39} & 27.85 & \first{28.41} & 27.80 & 29.66 & \second{30.07} & 29.71 & \first{30.13} & 29.63
    \\
    & SSIM$\uparrow$
      & 0.991 & \first{0.978} & \first{0.978} & \second{0.977} & \first{0.978} & \second{0.977} & 0.982 & 0.982 & 0.982 & 0.982 & 0.982
    \\
    & LPIPS$\downarrow$
      & 0.026 & 0.042 & \first{0.041} & 0.042 & \first{0.041} & 0.042 & \second{0.038} & \second{0.038} & \second{0.038} & \second{0.038} & \first{0.037}
    \\ \cmidrule(lr){1-1} \cmidrule(lr){2-2} \cmidrule(lr){3-3} \cmidrule(lr){4-8} \cmidrule(lr){9-13}
    \multirow{3}{*}{Cat}
    & PSNR$\uparrow$
      & 37.53 & 30.45 & \first{31.00} & 30.53 & \second{30.99} & 30.19 & 31.11 & \second{31.41} & 30.98 & \first{31.48} & 30.55
    \\
    & SSIM$\uparrow$
      & 0.993 & \first{0.987} & \first{0.987} & \first{0.987} & \first{0.987} & \second{0.986} & 0.988 & 0.988 & 0.988 & 0.988 & 0.988
    \\
    & LPIPS$\downarrow$
      & 0.016 & 0.033 & \second{0.032} & \first{0.031} & \first{0.031} & \first{0.031} & \second{0.034} & \first{0.032} & \first{0.032} & \first{0.032} & \first{0.032}
    \\ \cmidrule(lr){1-1} \cmidrule(lr){2-2} \cmidrule(lr){3-3} \cmidrule(lr){4-8} \cmidrule(lr){9-13}
    \multirow{3}{*}{Trophy}
    & PSNR$\uparrow$
      & 32.43 & 29.33 & \first{30.16} & 29.69 & \second{30.01} & 29.66 & 28.80 & \first{29.92} & 29.10 & \second{29.78} & 28.96
    \\
    & SSIM$\uparrow$
      & 0.967 & \second{0.963} & \first{0.964} & \second{0.963} & \first{0.964} & \second{0.963} & 0.962 & \first{0.964} & \second{0.963} & \first{0.964} & \second{0.963}
    \\
    & LPIPS$\downarrow$
      & 0.034 & 0.039 & \first{0.037} & 0.039 & \second{0.038} & 0.039 & 0.039 & \first{0.037} & 0.039 & \second{0.038} & 0.039
    \\ \cmidrule[0.1em](lr){1-1} \cmidrule[0.1em](lr){2-2} \cmidrule[0.1em](lr){3-3} \cmidrule[0.1em](lr){4-8} \cmidrule[0.1em](lr){9-13}
    \multirow{3}{*}{\texttt{Average}}
    & PSNR$\uparrow$
      & 35.99 & 27.39 & \first{29.22} & 28.22 & \second{28.89} & 28.33 & 28.47 & \first{30.11} & 29.31 & \second{29.87} & 29.39
    \\
    & SSIM$\uparrow$
      & 0.989 & 0.978 & \first{0.980} & \second{0.979} & \first{0.980} & \second{0.979} & 0.980 & \first{0.982} & \second{0.981} & \second{0.981} & \second{0.981}
    \\
    & LPIPS$\downarrow$
      & 0.020 & 0.034 & \first{0.032} & \second{0.033} & \first{0.032} & \second{0.033} & 0.033 & \first{0.031} & \second{0.032} & \first{0.031} & \first{0.031}
    \\ \bottomrule
  \end{tabularx}
  \vspace{-2mm}
  \caption{Comparison of PSNR$\uparrow$, SSIM$\uparrow$, and LPIPS$\downarrow$ for each scene on \textbf{geometric correction}.
    This table is an extended version of Table~\ref{tab:comparison_geometry}.
    The qualitative comparisons are provided in Figures~\ref{fig:results_newtonian}--\ref{fig:results_sand}.}
  \label{tab:comparison_geometry_ex}
\end{table*}

\section{Detailed analyses on main experiments}
\label{sec:analyses_main_experiments}

The main text focuses on representative results because of space limitations.
This appendix provides the extended results that further clarify the effectiveness of the proposed method.
In particular, we provide the scores for each scene (Appendix~\ref{subsec:each_scene}), discuss the computation times (Appendix~\ref{subsec:computation_times}), and discuss the impact of the number of iterations in Algorithm~\ref{alg:iterative_optimization} (Appendix~\ref{subsec:impact_iterations}).

\begin{table*}[t]
  \centering
  \scriptsize
  \setlength{\tabcolsep}{1pt}
  \begin{tabularx}{\textwidth}{ccCCCCCCCCCCCCCC}
    \toprule
    & & Ground truth & PAC-NeRF & PAC-NeRF-3v & +LPO$^4$ & +LPO-F$^4$ & +LPO-P$^4$ & +GO$^4$ & +None$^4$
    \\ \cmidrule(lr){1-1} \cmidrule(lr){2-2} \cmidrule(lr){3-3} \cmidrule(lr){4-4} \cmidrule(){5-10}
    \multirow{2}{*}{Droplet}
    & $\mu$
      & $2.00 \times 10^2$ & $2.19 \times 10^2$ & $2.76 \times 10^2$ & \second{$2.59 \times 10^2$} & \second{$2.59 \times 10^2$} & $2.63 \times 10^2$ & \first{$2.58 \times 10^2$} & $2.70 \times 10^2$
    \\
    & $\kappa$
      & $1.00 \times 10^5$ & $9.62 \times 10^4$ & \first{$5.19 \times 10^3$} & $2.36 \times 10^3$ & $2.30 \times 10^3$ & \second{$3.42 \times 10^3$} & $2.11 \times 10^3$ & $2.07 \times 10^3$
    \\ \cmidrule(lr){1-1} \cmidrule(lr){2-2} \cmidrule(lr){3-3} \cmidrule(lr){4-4} \cmidrule(){5-10}
    \multirow{2}{*}{Letter}
    & $\mu$
      & $1.00 \times 10^2$ & $9.10 \times 10^1$ & $2.12 \times 10^1$ & \first{$9.67 \times 10^1$} & \second{$9.42 \times 10^1$} & $1.20 \times 10^2$ & $1.12 \times 10^2$ & $2.95 \times 10^1$
    \\
    & $\kappa$
      & $1.00 \times 10^5$ & $9.14 \times 10^4$ & $1.69 \times 10^{-2}$ & \first{$6.69 \times 10^4$} & \second{$3.88 \times 10^4$} & $9.12 \times 10^3$ & $1.37 \times 10^4$ & $8.26 \times 10^{-2}$
    \\ \cmidrule(lr){1-1} \cmidrule(lr){2-2} \cmidrule(lr){3-3} \cmidrule(lr){4-4} \cmidrule(){5-10}
    \multirow{4}{*}{Cream}
    & $\mu$
      & $1.00 \times 10^4$ & $1.23 \times 10^4$ & $2.05 \times 10^4$ & \second{$1.51 \times 10^4$} & \first{$1.46 \times 10^4$} & $1.72 \times 10^4$ & $1.52 \times 10^4$ & $1.78 \times 10^4$
    \\
    & $\kappa$
      & $1.00 \times 10^6$ & $1.35 \times 10^6$ & $1.64 \times 10^6$ & $1.44 \times 10^6$ & $1.78 \times 10^6$ & \first{$9.39 \times 10^5$} & \second{$9.37 \times 10^5$} & $1.75 \times 10^6$
    \\
    & $\tau_Y$
      & $3.00 \times 10^3$ & $3.05 \times 10^3$ & $2.90 \times 10^3$ & \first{$2.97 \times 10^3$} & $2.80 \times 10^3$ & \second{$2.96 \times 10^3$} & $3.04 \times 10^3$ & $2.83 \times 10^3$
    \\
    & $\eta$
      & $10.00$ & $10.36$ & $19.10$ & $15.22$ & $18.02$ & \second{$12.11$} & \first{$8.69$} & $17.85$
    \\ \cmidrule(lr){1-1} \cmidrule(lr){2-2} \cmidrule(lr){3-3} \cmidrule(lr){4-4} \cmidrule(){5-10}
    \multirow{4}{*}{Toothpaste}
    & $\mu$
      & $5.00 \times 10^3$ & $5.31 \times 10^3$ & $3.89 \times 10^5$ & \first{$6.42 \times 10^3$} & $7.16 \times 10^3$ & \second{$3.80 \times 10^3$} & $3.43 \times 10^3$ & $3.05 \times 10^3$
    \\
    & $\kappa$
      & $1.00 \times 10^5$ & $5.66 \times 10^4$ & $2.63 \times 10^3$ & \second{$2.51 \times 10^4$} & $4.41 \times 10^3$ & \first{$2.53 \times 10^4$} & $2.25 \times 10^4$ & $1.87 \times 10^4$
    \\
    & $\tau_Y$
      & $2.00 \times 10^2$ & $2.33 \times 10^2$ & $3.18 \times 10^2$ & $1.54 \times 10^2$ & $3.11 \times 10^2$ & \second{$1.68 \times 10^2$} & $1.12 \times 10^2$ & \first{$1.74 \times 10^2$}
    \\
    & $\eta$
      & $10.00$ & $9.71$ & $4.23$ & \first{$9.93$} & $5.41$ & $10.31$ & \second{$10.16$} & $8.97$
    \\ \cmidrule(lr){1-1} \cmidrule(lr){2-2} \cmidrule(lr){3-3} \cmidrule(lr){4-4} \cmidrule(){5-10}
    \multirow{2}{*}{Torus}
    & $E$
      & $1.00 \times 10^6$ & $1.05 \times 10^6$ & $1.89 \times 10^6$ & $1.15 \times 10^6$ & \first{$1.02 \times 10^6$} & $1.21 \times 10^6$ & \second{$1.02 \times 10^6$} & $8.83 \times 10^5$
    \\
    & $\nu$
      & $0.300$ & $0.323$ & $0.215$ & \first{$0.299$} & $0.420$ & \second{$0.331$} & $0.374$ & $-0.016$
    \\ \cmidrule(lr){1-1} \cmidrule(lr){2-2} \cmidrule(lr){3-3} \cmidrule(lr){4-4} \cmidrule(){5-10}
    \multirow{2}{*}{Bird}
    & $E$
      & $3.00 \times 10^5$ & $2.91 \times 10^5$ & $8.43 \times 10^5$ & \first{$3.50 \times 10^5$} & $1.73 \times 10^5$ & \second{$3.56 \times 10^5$} & $1.46 \times 10^5$ & $1.70 \times 10^5$
    \\
    & $\nu$
      & $0.300$ & $0.329$ & $0.402$ & \first{$0.301$} & $-0.072$ & \second{$0.225$} & $-0.065$ & $-0.281$
    \\ \cmidrule(lr){1-1} \cmidrule(lr){2-2} \cmidrule(lr){3-3} \cmidrule(lr){4-4} \cmidrule(){5-10}
    \multirow{3}{*}{Playdoh}
    & $E$
      & $2.00 \times 10^6$ & $3.87 \times 10^6$ & $3.90 \times 10^6$ & \first{$2.61 \times 10^6$} & $7.61 \times 10^6$ & \second{$2.64 \times 10^6$} & $4.03 \times 10^6$ & $2.96 \times 10^6$
    \\
    & $\tau_Y$
      & $1.54 \times 10^4$ & $1.68 \times 10^4$ & $2.96 \times 10^4$ & \second{$2.25 \times 10^4$} & \first{$1.64 \times 10^4$} & $2.52 \times 10^4$ & $2.42 \times 10^4$ & $2.66 \times 10^4$
    \\
    & $\nu$
      & $0.300$ & $0.224$ & $-0.195$ & \second{$0.189$} & $0.127$ & \first{$0.191$} & $0.052$ & $0.172$
    \\ \cmidrule(lr){1-1} \cmidrule(lr){2-2} \cmidrule(lr){3-3} \cmidrule(lr){4-4} \cmidrule(){5-10}
    \multirow{3}{*}{Cat}
    & $E$
      & $1.00 \times 10^6$ & $1.39 \times 10^5$ & $5.00 \times 10^4$ & \first{$1.06 \times 10^5$} & $8.55 \times 10^4$ & $8.49 \times 10^4$ & $7.46 \times 10^4$ & \second{$8.58 \times 10^4$}
    \\
    & $\tau_Y$
      & $3.85 \times 10^3$ & $3.62 \times 10^3$ & $5.08 \times 10^3$ & \second{$4.93 \times 10^3$} & $5.04 \times 10^3$ & $4.99 \times 10^3$ & \first{$4.69 \times 10^3$} & $4.99 \times 10^3$
    \\
    & $\nu$
      & $0.300$ & $0.327$ & $0.344$ & \first{$0.304$} & \second{$0.336$} & $0.231$ & $0.110$ & $0.379$
    \\ \cmidrule(lr){1-1} \cmidrule(lr){2-2} \cmidrule(lr){3-3} \cmidrule(lr){4-4} \cmidrule(){5-10}
    \multirow{1}{*}{Trophy}
    & $\theta_{fric}$
      & $40.00^\circ$ & $37.28^\circ$ & \second{$36.97^\circ$} & $36.84^\circ$ & \first{$37.07^\circ$} & $36.79^\circ$ & $36.73^\circ$ & $36.90^\circ$
    \\ \bottomrule
  \end{tabularx}
  \vspace{-2mm}
  \caption{Comparison of the values of the physical properties for each scene on \textbf{physical identification} when PAC-NeRF-3v was used as a baseline.
    The absolute differences between the ground truth and the estimated physical properties are provided in Table~\ref{tab:comparison_physics}.
    The qualitative comparisons are provided in Figures~\ref{fig:results_newtonian}--\ref{fig:examples_feat}.}
  \vspace{4mm}
  \label{tab:comparison_physics_value1}
\end{table*}

\begin{table*}[t]
  \centering
  \scriptsize
  \setlength{\tabcolsep}{1pt}
  \begin{tabularx}{\textwidth}{ccCCCCCCCC}
    \toprule
    & & Ground truth & PAC-NeRF & PAC-NeRF-3v$^\dag$ & +LPO$^4$ & +LPO-F$^4$ & +LPO-P$^4$ & +GO$^4$ & +None$^4$
    \\ \cmidrule(lr){1-1} \cmidrule(lr){2-2} \cmidrule(lr){3-3} \cmidrule(lr){4-4} \cmidrule(){5-10}
    \multirow{2}{*}{Droplet}
    & $\mu$
      & $2.00 \times 10^2$ & $2.19 \times 10^2$ & $2.73 \times 10^2$ & $2.41 \times 10^2$ & \second{$2.34 \times 10^2$} & \first{$2.33 \times 10^2$} & $2.42 \times 10^2$ & $2.46 \times 10^2$
    \\
    & $\kappa$
      & $1.00 \times 10^5$ & $9.62 \times 10^4$ & $5.45 \times 10^3$ & \first{$1.28 \times 10^5$} & $7.26 \times 10^3$ & \second{$3.02 \times 10^4$} & $3.02 \times 10^3$ & $2.21 \times 10^3$
    \\ \cmidrule(lr){1-1} \cmidrule(lr){2-2} \cmidrule(lr){3-3} \cmidrule(lr){4-4} \cmidrule(){5-10}
    \multirow{2}{*}{Letter}
    & $\mu$
      & $1.00 \times 10^2$ & $9.10 \times 10^1$ & $4.18 \times 10^1$ & \second{$1.02 \times 10^2$} & $8.18 \times 10^1$ & \first{$9.84 \times 10^1$} & $7.62 \times 10^1$ & $3.66 \times 10^1$
    \\
    & $\kappa$
      & $1.00 \times 10^5$ & $9.14 \times 10^4$ & $5.91 \times 10^{-1}$ & \second{$8.70 \times 10^4$} & \first{$1.06 \times 10^5$} & $1.34 \times 10^5$ & $7.28 \times 10^4$ & $8.72 \times 10^{-1}$
    \\ \cmidrule(lr){1-1} \cmidrule(lr){2-2} \cmidrule(lr){3-3} \cmidrule(lr){4-4} \cmidrule(){5-10}
    \multirow{4}{*}{Cream}
    & $\mu$
      & $1.00 \times 10^4$ & $1.23 \times 10^4$ & $1.51 \times 10^4$ & $1.26 \times 10^4$ & \second{$1.18 \times 10^4$} & \first{$1.16 \times 10^4$} & $1.24 \times 10^4$ & $1.19 \times 10^4$
    \\
    & $\kappa$
      & $1.00 \times 10^6$ & $1.35 \times 10^6$ & $2.17 \times 10^6$ & \first{$1.32 \times 10^6$} & $1.57 \times 10^6$ & \second{$1.39 \times 10^6$} & $1.56 \times 10^6$ & $2.18 \times 10^6$
    \\
    & $\tau_Y$
      & $3.00 \times 10^3$ & $3.05 \times 10^3$ & $2.94 \times 10^3$ & $3.04 \times 10^3$ & \first{$2.99 \times 10^3$} & \second{$2.97 \times 10^3$} & $2.93 \times 10^3$ & $2.95 \times 10^3$
    \\
    & $\eta$
      & $10.00$ & $10.36$ & $15.67$ & \first{$10.80$} & $11.44$ & $12.01$ & $12.47$ & \second{$11.25$}
    \\ \cmidrule(lr){1-1} \cmidrule(lr){2-2} \cmidrule(lr){3-3} \cmidrule(lr){4-4} \cmidrule(){5-10}
    \multirow{4}{*}{Toothpaste}
    & $\mu$
      & $5.00 \times 10^3$ & $5.31 \times 10^3$ & $2.80 \times 10^3$ & \second{$4.66 \times 10^3$} & $3.15 \times 10^3$ & \first{$4.94 \times 10^3$} & $2.80 \times 10^3$ & $3.64 \times 10^3$
    \\
    & $\kappa$
      & $1.00 \times 10^5$ & $5.66 \times 10^4$ & $3.67 \times 10^3$ & $2.12 \times 10^4$ & $2.34 \times 10^4$ & $1.26 \times 10^4$ & \first{$2.57 \times 10^4$} & \second{$2.53 \times 10^4$}
    \\
    & $\tau_Y$
      & $2.00 \times 10^2$ & $2.33 \times 10^2$ & $3.16 \times 10^2$ & $1.62 \times 10^2$ & \first{$1.72 \times 10^2$} & $1.53 \times 10^2$ & $1.42 \times 10^2$ & \second{$1.77 \times 10^2$}
    \\
    & $\eta$
      & $10.00$ & $9.71$ & $6.13$ & $10.20$ & $9.65$ & \first{$10.12$} & \second{$10.16$} & $9.07$
    \\ \cmidrule(lr){1-1} \cmidrule(lr){2-2} \cmidrule(lr){3-3} \cmidrule(lr){4-4} \cmidrule(){5-10}
    \multirow{2}{*}{Torus}
    & $E$
      & $1.00 \times 10^6$ & $1.05 \times 10^6$ & $8.43 \times 10^5$ & $1.09 \times 10^6$ & \second{$1.06 \times 10^6$} & $1.09 \times 10^6$ & \first{$9.66 \times 10^5$} & $1.18 \times 10^6$
    \\
    & $\nu$
      & $0.300$ & $0.323$ & $0.431$ & \first{$0.307$} & $0.340$ & \second{$0.267$} & $0.350$ & $0.429$
    \\ \cmidrule(lr){1-1} \cmidrule(lr){2-2} \cmidrule(lr){3-3} \cmidrule(lr){4-4} \cmidrule(){5-10}
    \multirow{2}{*}{Bird}
    & $E$
      & $3.00 \times 10^5$ & $2.91 \times 10^5$ & $3.99 \times 10^5$ & \first{$3.19 \times 10^5$} & $4.61 \times 10^5$ & \second{$3.28 \times 10^5$} & $5.10 \times 10^5$ & $5.93 \times 10^5$
    \\
    & $\nu$
      & $0.300$ & $0.329$ & $0.441$ & \second{$0.347$} & $0.372$ & \first{$0.308$} & $0.432$ & $0.445$
    \\ \cmidrule(lr){1-1} \cmidrule(lr){2-2} \cmidrule(lr){3-3} \cmidrule(lr){4-4} \cmidrule(){5-10}
    \multirow{3}{*}{Playdoh}
    & $E$
      & $2.00 \times 10^6$ & $3.87 \times 10^6$ & $6.63 \times 10^6$ & \first{$2.72 \times 10^6$} & $5.95 \times 10^6$ & $1.51 \times 10^4$ & \second{$3.41 \times 10^6$} & $1.42 \times 10^4$
    \\
    & $\tau_Y$
      & $1.54 \times 10^4$ & $1.68 \times 10^4$ & \second{$1.99 \times 10^4$} & $2.30 \times 10^4$ & \first{$1.85 \times 10^4$} & $2.33 \times 10^5$ & $2.42 \times 10^4$ & $1.43 \times 10^5$
    \\
    & $\nu$
      & $0.300$ & $0.224$ & $0.088$ & \first{$0.237$} & $0.167$ & $0.410$ & $0.173$ & \second{$0.406$}
    \\ \cmidrule(lr){1-1} \cmidrule(lr){2-2} \cmidrule(lr){3-3} \cmidrule(lr){4-4} \cmidrule(){5-10}
    \multirow{3}{*}{Cat}
    & $E$
      & $1.00 \times 10^6$ & $1.39 \times 10^5$ & $6.42 \times 10^4$ & \second{$1.97 \times 10^5$} & \first{$2.93 \times 10^5$} & $1.94 \times 10^5$ & $1.65 \times 10^5$ & $1.61 \times 10^5$
    \\
    & $\tau_Y$
      & $3.85 \times 10^3$ & $3.62 \times 10^3$ & $4.67 \times 10^3$ & \second{$4.50 \times 10^3$} & \first{$3.97 \times 10^3$} & $4.59 \times 10^3$ & $4.52 \times 10^3$ & $4.54 \times 10^3$
    \\
    & $\nu$
      & $0.300$ & $0.327$ & $0.418$ & \first{$0.303$} & 0.327 & \second{$0.293$} & $0.272$ & $0.363$
    \\ \cmidrule(lr){1-1} \cmidrule(lr){2-2} \cmidrule(lr){3-3} \cmidrule(lr){4-4} \cmidrule(){5-10}
    \multirow{1}{*}{Trophy}
    & $\theta_{fric}$
      & $40.00^\circ$ & $37.28^\circ$ & \first{$38.31^\circ$} & $37.75^\circ$ & $37.66^\circ$ & $37.51^\circ$ & $37.46^\circ$ & \second{$37.79^\circ$}
    \\ \bottomrule
  \end{tabularx}
  \vspace{-2mm}
  \caption{Comparison of the values of the physical properties for each scene on \textbf{physical identification} when PAC-NeRF-3v$^\dag$ was used as a baseline.
    The absolute differences between the ground truth and the estimated physical properties are provided in Table~\ref{tab:comparison_physics}.
    The qualitative comparisons are provided in Figures~\ref{fig:results_newtonian}--\ref{fig:examples_feat}.}
  \label{tab:comparison_physics_value2}
\end{table*}

\begin{table*}[t]
  \centering
  \scriptsize
  \setlength{\tabcolsep}{2pt}
  \begin{tabularx}{\textwidth}{CCCCCCCCCCCCCCC}
    \toprule
    & & PAC- & PAC- & \multirow{2}{*}{+LPO$^4$} & \multirow{2}{*}{+LPO-F$^4$} & \multirow{2}{*}{+LPO-P$^4$} & \multirow{2}{*}{+GO$^4$} & \multirow{2}{*}{+None$^4$} & PAC- & \multirow{2}{*}{+LPO$^4$} & \multirow{2}{*}{+LPO-F$^4$} & \multirow{2}{*}{+LPO-P$^4$} & \multirow{2}{*}{+GO$^4$} & \multirow{2}{*}{+None$^4$}
    \\
    & & NeRF & NeRF-3v & & & & & & \!\!\!NeRF-3v$^\dag$\!\!\!
    \\ \cmidrule(lr){1-1} \cmidrule(lr){2-2} \cmidrule(lr){3-3} \cmidrule(lr){4-9} \cmidrule(lr){10-15}
    \multirow{3}{*}{Droplet}
    & PSNR$\uparrow$
      & 35.30 & 25.42 & \first{28.21} & \second{26.52} & \first{28.21} & 26.03 & 25.42 & 26.40 & \second{28.91} & 27.85 & \first{29.03} & 26.23 & 27.52
    \\
    & SSIM$\uparrow$
      & 0.990 & 0.975 & \first{0.981} & 0.978 & \second{0.980} & 0.978 & 0.977 & 0.978 & \first{0.983} & \second{0.982} & \first{0.983} & 0.980 & 0.981
    \\
    & LPIPS$\downarrow$
      & 0.029 & 0.047 & \first{0.042} & 0.046 & \second{0.045} & 0.046 & 0.047 & 0.046 & \first{0.041} & \second{0.042} & 0.043 & 0.045 & 0.043
    \\ \cmidrule(lr){1-1} \cmidrule(lr){2-2} \cmidrule(lr){3-3} \cmidrule(lr){4-9} \cmidrule(lr){10-15}
    \multirow{3}{*}{Letter}
    & PSNR$\uparrow$
      & 36.01 & 28.94 & \first{30.07} & \first{30.07} & 29.83 & \second{30.00} & 29.25 & 29.59 & 30.26 & \first{31.09} & \second{30.50} & 29.70 & 29.80
    \\
    & SSIM$\uparrow$
      & 0.991 & 0.981 & \first{0.983} & \first{0.983} & \second{0.982} & \first{0.983} & \first{0.983} & \second{0.983} & \second{0.983} & \first{0.984} & 0.982 & 0.982 & \first{0.984}
    \\
    & LPIPS$\downarrow$
      & 0.012 & \second{0.028} & \first{0.026} & \first{0.026} & 0.029 & \first{0.026} & \first{0.026} & 0.028 & 0.026 & \first{0.024} & 0.029 & \second{0.025} & \second{0.025}
    \\ \cmidrule(lr){1-1} \cmidrule(lr){2-2} \cmidrule(lr){3-3} \cmidrule(lr){4-9} \cmidrule(lr){10-15}
    \multirow{3}{*}{Cream}
    & PSNR$\uparrow$
      & 36.70 & 26.61 & \first{29.65} & \second{29.02} & 28.83 & 28.95 & 26.57 & 27.43 & \first{30.23} & \second{29.87} & 29.65 & 29.79 & 27.88
    \\
    & SSIM$\uparrow$
      & 0.993 & 0.983 & \first{0.985} & \first{0.985} & \second{0.984} & \first{0.985} & 0.983 & 0.983 & \first{0.985} & \first{0.985} & \first{0.985} & \first{0.985} & \second{0.984}
    \\
    & LPIPS$\downarrow$
      & 0.014 & 0.026 & \first{0.022} & \second{0.023} & 0.024 & \first{0.022} & 0.026 & 0.024 & \first{0.021} & \second{0.022} & \second{0.022} & \first{0.021} & 0.024
    \\ \cmidrule(lr){1-1} \cmidrule(lr){2-2} \cmidrule(lr){3-3} \cmidrule(lr){4-9} \cmidrule(lr){10-15}
    \multirow{3}{*}{Toothpaste}
    & PSNR$\uparrow$
      & 39.46 & 29.23 & 32.18 & \second{32.41} & 31.87 & \first{32.87} & 28.96 & 31.72 & 34.00 & \first{34.51} & 33.94 & \second{34.40} & 31.49
    \\
    & SSIM$\uparrow$
      & 0.996 & \second{0.991} & \second{0.991} & \first{0.992} & \second{0.991} & \first{0.992} & \second{0.991} & \first{0.993} & \second{0.992} & \first{0.993} & \first{0.993} & \first{0.993} & \first{0.993}
    \\
    & LPIPS$\downarrow$
      & 0.006 & \first{0.010} & \second{0.011} & \first{0.010} & \second{0.011} & \first{0.010} & \first{0.010} & \second{0.010} & \second{0.010} & \first{0.009} & \second{0.010} & \first{0.009} & \first{0.009}
    \\ \cmidrule(lr){1-1} \cmidrule(lr){2-2} \cmidrule(lr){3-3} \cmidrule(lr){4-9} \cmidrule(lr){10-15}
    \multirow{3}{*}{Torus}
    & PSNR$\uparrow$
      & 34.54 & 23.99 & \first{30.07} & 27.06 & \second{29.37} & 25.36 & 24.82 & 26.65 & \first{30.48} & 30.01 & \second{30.23} & 28.93 & 26.74
    \\
    & SSIM$\uparrow$
      & 0.988 & 0.970 & \first{0.982} & 0.975 & \second{0.981} & 0.970 & 0.973 & 0.974 & \first{0.983} & \second{0.982} & \second{0.982} & 0.979 & 0.975
    \\
    & LPIPS$\downarrow$
      & 0.026 & 0.048 & \first{0.032} & 0.039 & \second{0.035} & 0.041 & 0.044 & 0.040 & \first{0.031} & \second{0.032} & 0.033 & 0.033 & 0.040
    \\ \cmidrule(lr){1-1} \cmidrule(lr){2-2} \cmidrule(lr){3-3} \cmidrule(lr){4-9} \cmidrule(lr){10-15}
    \multirow{3}{*}{Bird}
    & PSNR$\uparrow$
      & 35.55 & 24.82 & \first{27.97} & 24.86 & \second{27.71} & 24.20 & 23.98 & 24.91 & \first{28.98} & 26.51 & \second{28.74} & 26.08 & 24.95
    \\
    & SSIM$\uparrow$
      & 0.991 & 0.978 & \first{0.980} & \second{0.979} & \first{0.980} & 0.977 & 0.978 & 0.979 & \first{0.982} & \second{0.980} & \first{0.982} & 0.979 & \second{0.980}
    \\
    & LPIPS$\downarrow$
      & 0.019 & \second{0.036} & \first{0.034} & 0.039 & \first{0.034} & 0.041 & 0.040 & 0.036 & \first{0.031} & 0.033 & \second{0.032} & 0.034 & 0.036
    \\ \cmidrule(lr){1-1} \cmidrule(lr){2-2} \cmidrule(lr){3-3} \cmidrule(lr){4-9} \cmidrule(lr){10-15}
    \multirow{3}{*}{Playdoh}
    & PSNR$\uparrow$
      & 36.43 & 27.70 & \first{29.32} & 28.05 & \second{29.01} & 28.02 & 27.94 & 29.66 & \first{30.01} & \second{29.45} & 29.25 & 28.94 & 28.02
    \\
    & SSIM$\uparrow$
      & 0.991 & 0.978 & \first{0.981} & 0.978 & \second{0.980} & 0.979 & 0.979 & \second{0.982} & \first{0.983} & \second{0.982} & \second{0.982} & \second{0.982} & 0.981
    \\
    & LPIPS$\downarrow$
      & 0.026 & \second{0.042} & \first{0.040} & 0.043 & 0.043 & \second{0.042} & \second{0.042} & \first{0.038} & \first{0.038} & \second{0.039} & 0.045 & 0.041 & 0.047
    \\ \cmidrule(lr){1-1} \cmidrule(lr){2-2} \cmidrule(lr){3-3} \cmidrule(lr){4-9} \cmidrule(lr){10-15}
    \multirow{3}{*}{Cat}
    & PSNR$\uparrow$
      & 37.53 & \first{30.45} & 29.81 & 29.71 & 29.95 & 28.35 & \second{30.41} & 31.11 & 30.61 & 30.47 & \second{31.47} & 29.21 & \first{31.78}
    \\
    & SSIM$\uparrow$
      & 0.993 & \first{0.987} & \first{0.987} & \first{0.987} & \first{0.987} & \second{0.985} & \first{0.987} & \second{0.988} & \second{0.988} & \second{0.988} & \first{0.989} & 0.987 & \first{0.989}
    \\
    & LPIPS$\downarrow$
      & 0.016 & \second{0.033} & \first{0.028} & \first{0.028} & \first{0.028} & \first{0.028} & \first{0.028} & 0.034 & \first{0.025} & \second{0.026} & \first{0.025} & \second{0.026} & \first{0.025}
    \\ \cmidrule(lr){1-1} \cmidrule(lr){2-2} \cmidrule(lr){3-3} \cmidrule(lr){4-9} \cmidrule(lr){10-15}
    \multirow{3}{*}{Trophy}
    & PSNR$\uparrow$
      & 32.43 & 29.33 & \second{29.57} & 29.13 & \first{29.89} & 28.31 & 29.44 & 28.80 & \second{29.58} & 29.23 & \first{30.01} & 27.87 & 29.18
    \\
    & SSIM$\uparrow$
      & 0.967 & \second{0.963} & \first{0.964} & \second{0.963} & \first{0.964} & \second{0.963} & \second{0.963} & 0.962 & \first{0.964} & \second{0.963} & \first{0.964} & \second{0.963} & \second{0.963}
    \\
    & LPIPS$\downarrow$
      & 0.034 & \second{0.039} & \first{0.037} & \second{0.039} & \second{0.039} & \second{0.039} & \second{0.039} & 0.039 & \first{0.036} & 0.039 & \second{0.038} & 0.039 & 0.039
    \\ \cmidrule[0.1em](lr){1-1} \cmidrule[0.1em](lr){2-2} \cmidrule[0.1em](lr){3-3} \cmidrule[0.1em](lr){4-9} \cmidrule[0.1em](lr){10-15}
    \multirow{3}{*}{\texttt{Average}}
    & PSNR$\uparrow$
      & 35.99 & 27.39 & \first{29.65} & 28.54 & \second{29.41} & 28.01 & 27.42 & 28.47 & \first{30.34} & 29.89 & \second{30.31} & 29.02 & 28.60
    \\
    & SSIM$\uparrow$
      & 0.989 & 0.978 & \first{0.982} & 0.980 & \second{0.981} & 0.979 & 0.979 & 0.980 & \first{0.983} & \second{0.982} & \second{0.982} & 0.981 & 0.981
    \\
    & LPIPS$\downarrow$
      & 0.020 & 0.034 & \first{0.030} & \second{0.032} & \second{0.032} & 0.033 & 0.034 & 0.033 & \first{0.029} & \second{0.030} & 0.031 & \second{0.030} & 0.032
    \\ \bottomrule
  \end{tabularx}
  \vspace{-2mm}
  \caption{Comparison of PSNR$\uparrow$, SSIM$\uparrow$, and LPIPS$\downarrow$ for each scene on \textbf{geometric recorrection}.
    This table is an extended version of Table~\ref{tab:comparison_geometry4}.
    The qualitative comparisons are provided in Figures~\ref{fig:results_newtonian}--\ref{fig:examples_feat}.}
  \label{tab:comparison_geometry4_ex}
\end{table*}

\subsection{Scores for each scene}
\label{subsec:each_scene}

In this appendix, we provide the scores for each scene.
To evaluate the effectiveness of the proposed method from various perspectives, we describe three experiments in the main text.
(I) Evaluation of the geometric correction (Table~\ref{tab:comparison_geometry}), where the image quality after LPO was applied once, was evaluated.
(II) Evaluation of the physical identification (Table~\ref{tab:comparison_physics}), in which the accuracy of the physical property estimation after Algorithm~\ref{alg:iterative_optimization} was applied, was evaluated. 
(III) Evaluation of the geometric recorrection (Table~\ref{tab:comparison_geometry4}), in which the image quality after Algorithm~\ref{alg:iterative_optimization} was applied, was evaluated.
In this appendix, we discuss these issues in detail.

\smallskip\noindent
\textbf{I. Evaluation of geometric correction.}
First, we discuss the scores for each scene on geometric correction (image quality after LPO was applied once) in detail.
Table~\ref{tab:comparison_geometry_ex} summarizes the results.
This is an extension of the list in Table~\ref{tab:comparison_geometry}.
We discuss the results from three perspectives.

\begin{figure*}[t]
  \begin{center}
    \includegraphics[width=\textwidth]{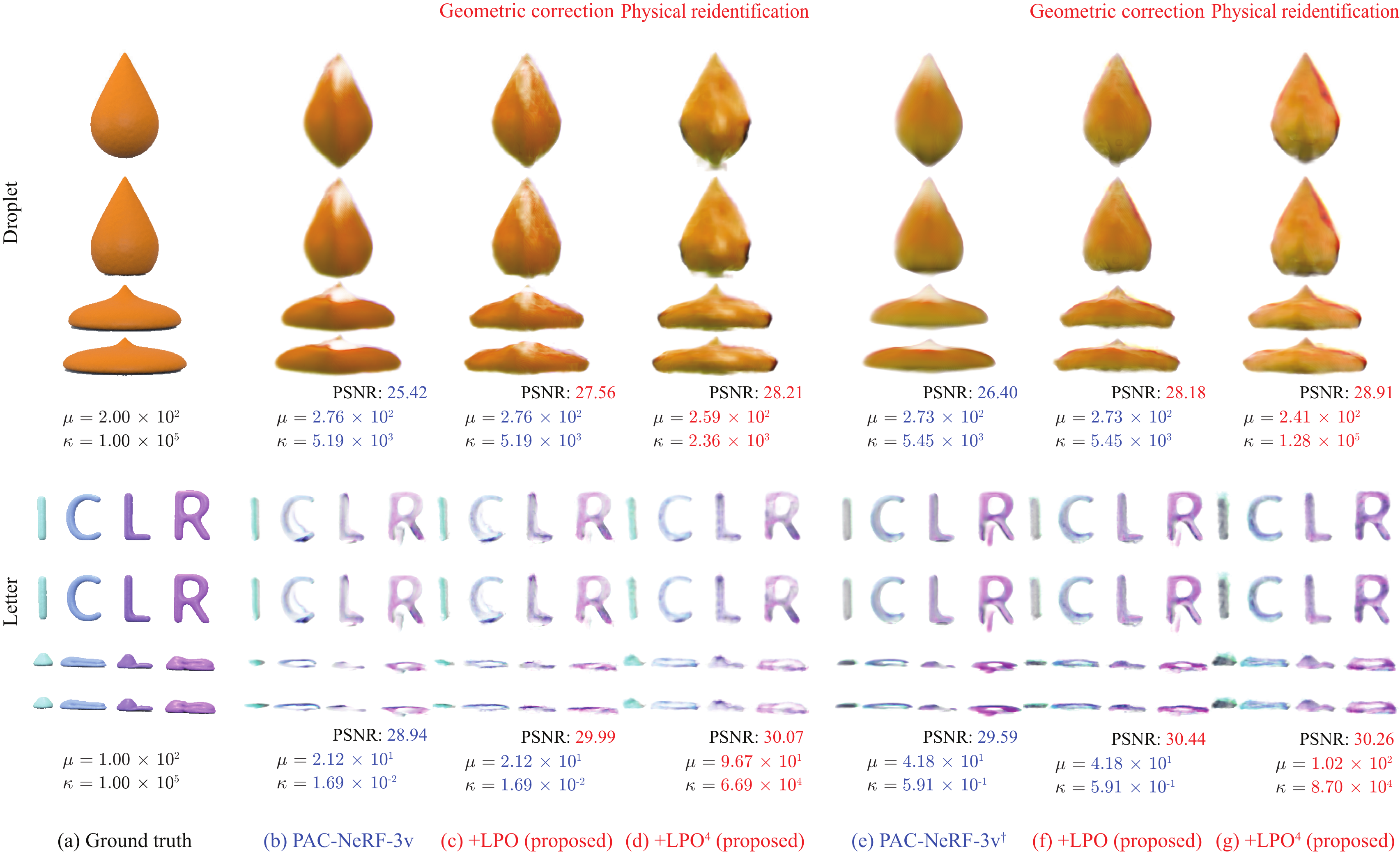}
  \end{center}
  \caption{Qualitative comparisons among PAC-NeRF-3v/3v$^\dag$, +LPO, and +LPO$^4$ on Newtonian fluids (Droplet and Letter).
    Blue fonts indicate the scores obtained by the baselines (PAC-NeRF-3v/3v$^\dag$).
    Red fonts indicate the scores obtained by the proposed methods (+LPO and +LPO$^4$).
    Given the initial estimation by the baseline (b)(e), +LPO first corrects the geometric structures (including appearance and shape) (c)(f).
    By repeatedly conducting physical identification and geometric correction via Algorithm~\ref{alg:iterative_optimization}, +LPO$^4$ reidentifies physical properties and recorrects geometric structures (d)(g).
    In the Droplet scene, the bottom of the droplet is sharply pointed, and its tip is whitened in the baseline (b)(e).
    They are gradually mitigated by applying +LPO (c)(f) and +LPO$^4$ (d)(g).
    In the Letter scene, +LPO (c)(f) and +LPO$^4$ (d)(g) succeed in gradually eliminating artifacts existing in the vicinity of the left line of the letter ``R.'', which arise in the baselines (b)(e).}
  \label{fig:results_newtonian}
\end{figure*}

\begin{figure*}[t]
  \begin{center}
    \includegraphics[width=\textwidth]{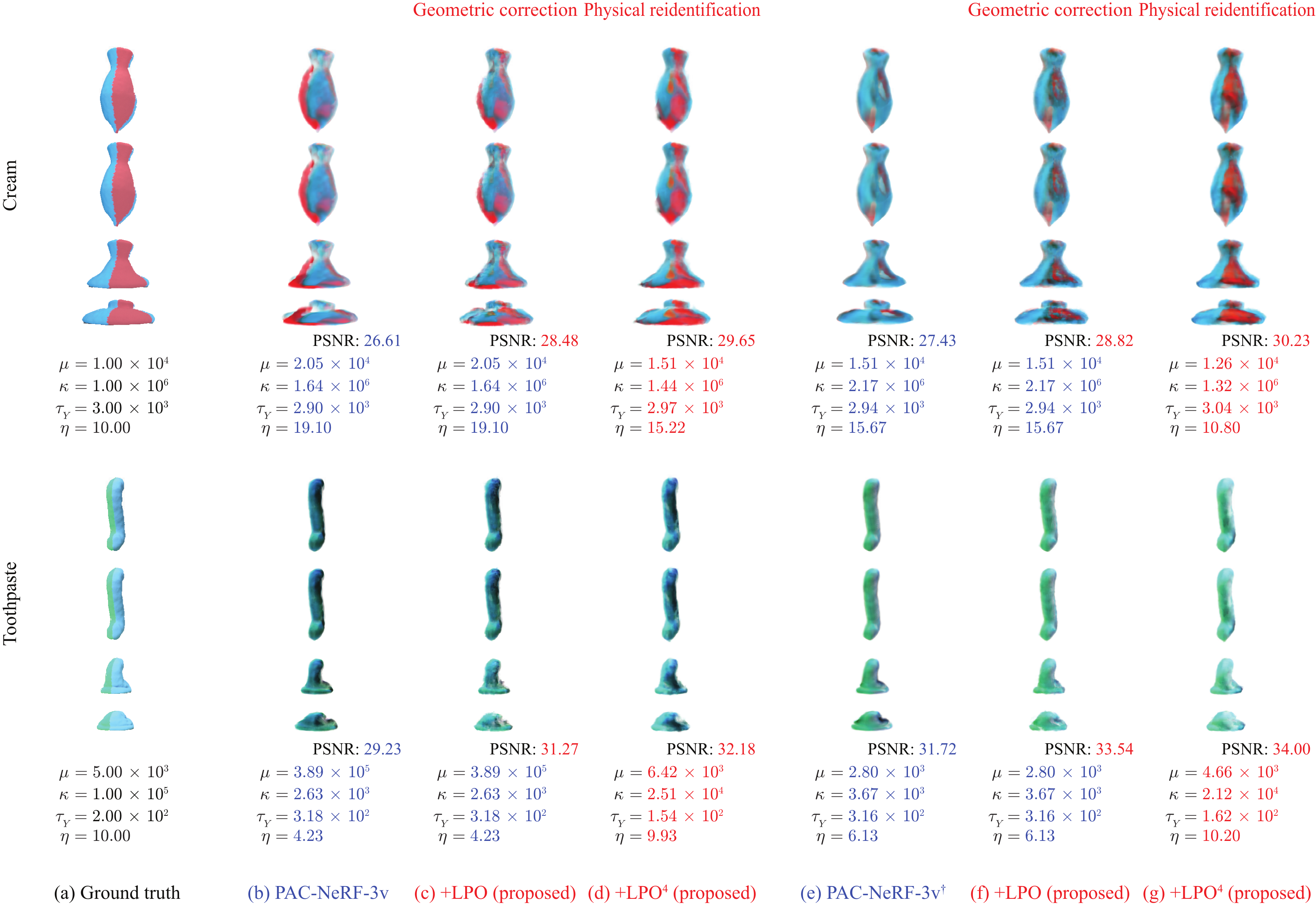}
  \end{center}
  \caption{Qualitative comparisons among PAC-NeRF-3v/3v$^\dag$, +LPO, and +LPO$^4$ on non-Newtonian fluids (Cream and Toothpaste).
    Blue fonts indicate the scores obtained by the baselines (PAC-NeRF-3v/3v$^\dag$).
    Red fonts indicate the scores obtained by the proposed methods (+LPO and +LPO$^4$).
    Given the initial estimation by the baseline (b)(e), +LPO first corrects the geometric structures (including appearance and shape) (c)(f).
    By repeatedly conducting physical identification and geometric correction via Algorithm~\ref{alg:iterative_optimization}, +LPO$^4$ reidentifies physical properties and recorrects geometric structures (d)(g).
    In the Cream scene, the baselines (b)(e) fail to color the materials correctly.
    This failure is alleviated by +LPO (c)(f) and further mitigated by +LPO$^4$ (d)(g).
    In the Toothpaste scene, the baselines (b)(e) make the material darker color than that in the ground truth (a).
    +LPO (c)(f) makes the material brighter, and +LPO$^4$ (d)(g) obtains the color closer to the ground truth (a).
    This effect is pronounced when PAC-NeRF-3v$^\dag$ (e) is used as a baseline.}
  \label{fig:results_nonnewtonian}
\end{figure*}

\begin{figure*}[t]
  \begin{center}
    \includegraphics[width=\textwidth]{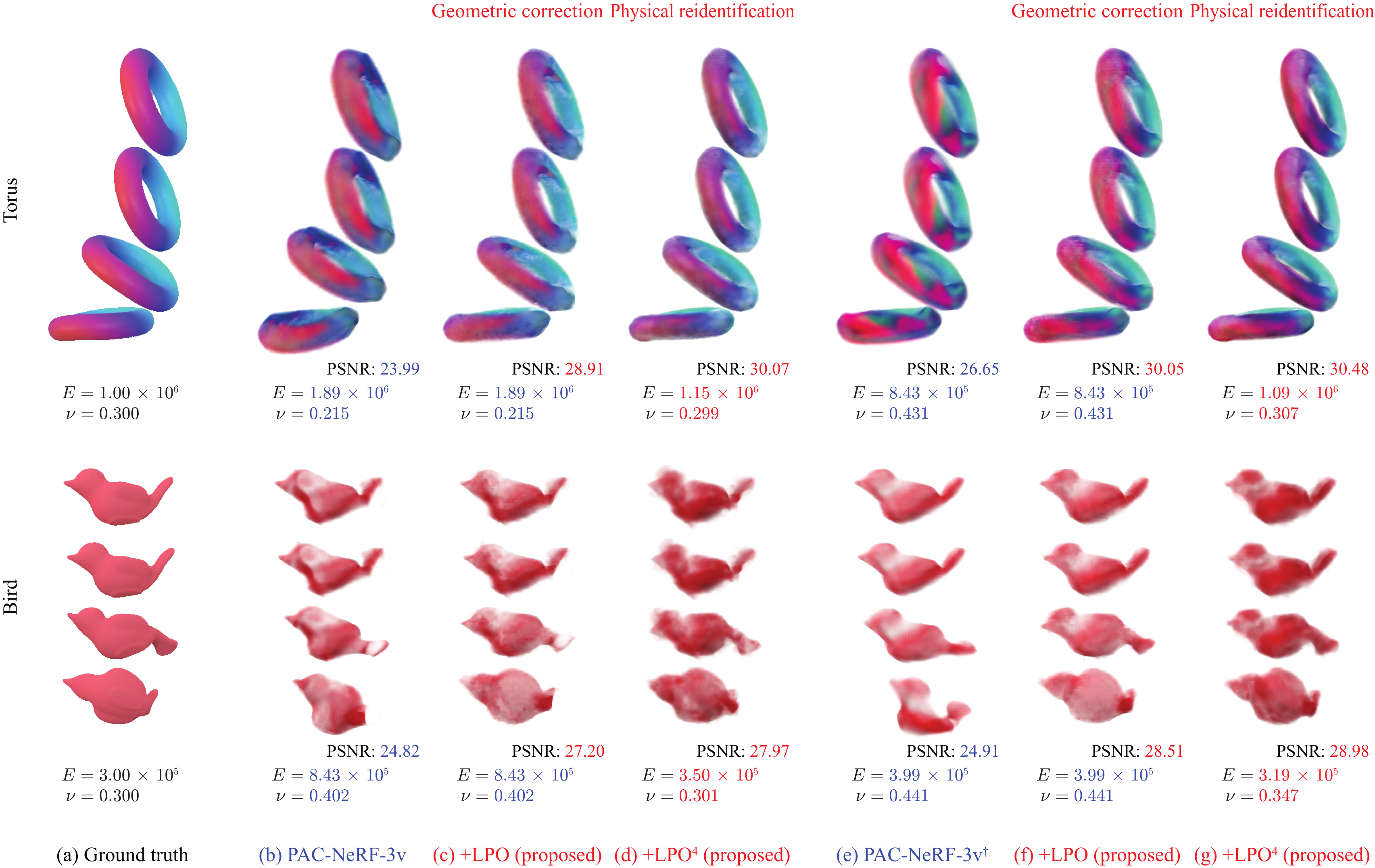}
  \end{center}
  \caption{Qualitative comparisons among PAC-NeRF-3v/3v$^\dag$, +LPO, and +LPO$^4$ on elastic materials (Torus and Bird).
    Blue fonts indicate the scores obtained by the baselines (PAC-NeRF-3v/3v$^\dag$).
    Red fonts indicate the scores obtained by the proposed methods (+LPO and +LPO$^4$).
    Given the initial estimation by the baseline (b)(e), +LPO first corrects the geometric structures (including appearance and shape) (c)(f).
    By repeatedly conducting physical identification and geometric correction via Algorithm~\ref{alg:iterative_optimization}, +LPO$^4$ reidentifies physical properties and recorrects geometric structures (d)(g).
    In the Torus scene, the baselines (b)(e) have difficulty correctly capturing color and shape.
    They are improved by applying +LPO (c)(f), and the fine details are also improved by utilizing +LPO$^4$ (d)(g).
    Also, in the Bird scene, the baselines (b)(e) fail to capture color and shape correctly.
    The shape (e.g., the directions of the tail) is first corrected by +LPO (c)(f), and then the color is corrected by +LPO$^4$ (d)(g).}
  \label{fig:results_elasticity}
\end{figure*}

\begin{figure*}[t]
  \begin{center}
    \includegraphics[width=\textwidth]{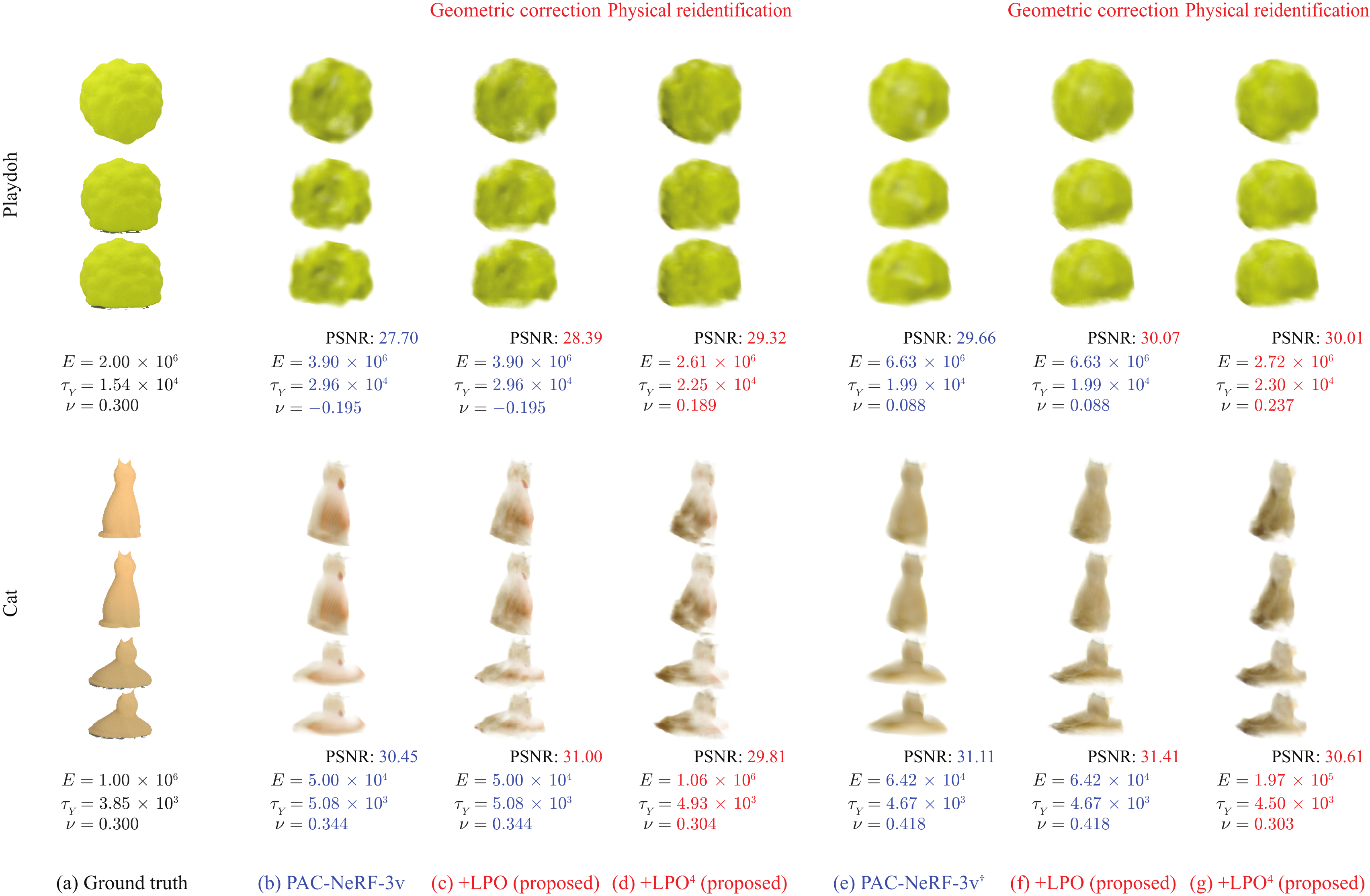}
  \end{center}
  \caption{Qualitative comparisons among PAC-NeRF-3v/3v$^\dag$, +LPO, and +LPO$^4$ on plasticine (Playdoh and Cat).
    Blue fonts indicate the scores obtained by the baselines (PAC-NeRF-3v/3v$^\dag$).
    Red fonts indicate the scores obtained by the proposed methods (+LPO and +LPO$^4$).
    Given the initial estimation by the baseline (b)(e), +LPO first corrects the geometric structures (including appearance and shape) (c)(f).
    By repeatedly conducting physical identification and geometric correction via Algorithm~\ref{alg:iterative_optimization}, +LPO$^4$ reidentifies physical properties and recorrects geometric structures (d)(g).
    In the Playdoh scene, the baselines (b)(e) make the playdoh a little crushed compared to the ground truth (a).
    These geometric failures are gradually alleviated by applying +LPO (c)(f) and +LPO$^4$ (d)(g).
    In the Cat scene, the baselines (b)(e) fail to capture the tail of the cat in the lower left corner.
    +LPO (c)(f) and +LPO$^4$ (d)(g) struggle to recover it.}
  \label{fig:results_plasticine}
\end{figure*}

\begin{figure*}[t]
  \begin{center}
    \includegraphics[width=\textwidth]{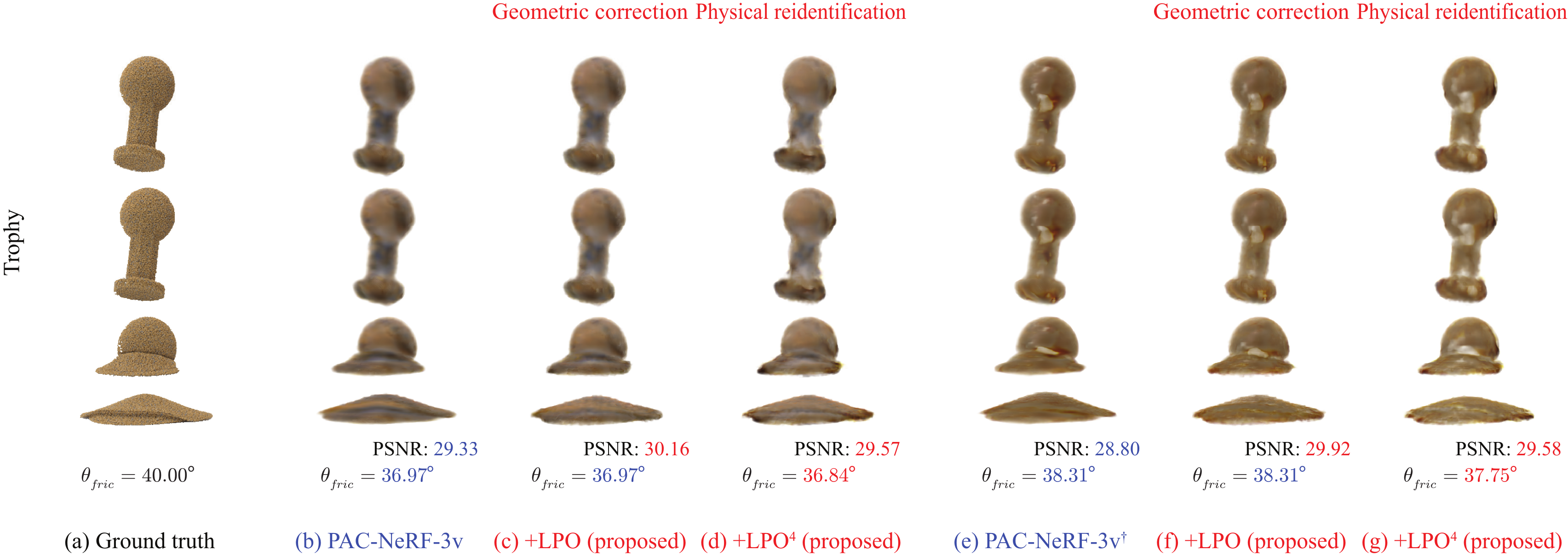}
  \end{center}
  \caption{Qualitative comparisons among PAC-NeRF-3v/3v$^\dag$, +LPO, and +LPO$^4$ on granular media (Trophy).
    Blue fonts indicate the scores obtained by the baselines (PAC-NeRF-3v/3v$^\dag$).
    Red fonts indicate the scores obtained by the proposed methods (+LPO and +LPO$^4$).
    Given the initial estimation by the baseline (b)(e), +LPO first corrects the geometric structures (including appearance and shape) (c)(f).
    By repeatedly conducting physical identification and geometric correction via Algorithm~\ref{alg:iterative_optimization}, +LPO$^4$ reidentifies physical properties and recorrects geometric structures (d)(g).
    In the Trophy scene, the baselines (b)(e) achieve good performance in terms of physical identification (comparable with PAC-NeRF with full-view supervision).
    However, they tend to make the trophy darker than the ground truth (a).
    This misestimation is corrected by +LPO and +LPO$^4$.}
  \label{fig:results_sand}
\end{figure*}

\smallskip\noindent
\textit{(1) PAC-NeRF-3v/3v$^\dag$ vs. +LPO.}
When PAC-NeRF-3v was used as the baseline, +LPO outperformed the baseline in most cases (21 wins, 5 draws, and 1 loss).
Similarly, when PAC-NeRF-3v$^\dag$ was used as the baseline, +LPO outperformed the baseline in most cases (20 wins, 5 draws, and 2 losses).
Consequently, in both cases, +LPO yielded better average scores for all metrics.
These results demonstrate that +LPO is effective for geometric correction independent of the baseline models.
The qualitative comparisons are shown in Figures~\ref{fig:results_newtonian}--\ref{fig:results_sand}.

\smallskip\noindent
\textit{(2) +LPO vs. +LPO-F/P.}
+LPO-F and +LPO-P are ablated variants of +LPO.
In +LPO-F, position (shape) optimization is ablated, and only feature (appearance) optimization is conducted.
In +LPO-P, feature (i.e., appearance) optimization is ablated, and only position (i.e., shape) optimization is conducted.
+LPO outperformed +LPO-F/P in most cases when both PAC-NeRF-3v and PAC-NeRF-3v$^\dag$ were used as the baselines.
Specifically, when PAC-NeRF-3v was used as the baseline, +LPO outperformed +LPO-F with 20 wins, 4 draws, and 3 losses and outperformed +LPO-P with 15 wins, 10 draws, and 2 losses.
When PAC-NeRF-3v$^\dag$ was used as the baseline, +LPO outperformed +LPO-F with 17 wins, 8 draws, and 2 losses, and outperformed +LPO-P with 12 wins, 13 draws, and 2 losses, respectively.
Between +LPO-F and +LPO-P, +LPO-P tends to outperform +LPO-F.
Specifically, when PAC-NeRF-3v was used as the baseline, +LPO-P outperformed +LPO-F with 17 wins, 5 draws, and 5 losses.
When PAC-NeRF-3v$^\dag$ was used as the baseline, +LPO-P outperformed +LPO-F with 13 wins, 9 draws, and 5 losses.
We consider this because shape correction by +LPO-P can correct the failure estimation of the geometry within the physical constraints of MPM.
In contrast, appearance correction by +LPO-F cannot do so and can cause overcorrection beyond the physical constraints, as shown in Figure~\ref{fig:examples_ptcl}.
These results indicate that the feature and position optimizations are complementary rather than competitive.

\smallskip\noindent
\textit{(3) +LPO vs. +GO.}
The difference between these two models is that +LPO conducts optimization in Lagrangian particle space and can optimize not only the features but also the positions of the particles, whereas +GO performs optimization in Eulerian grid space and can optimize the features of the grids, but cannot optimize their positions.
Owing to these characteristics, the performance of +GO is close to that of +LPO-F, which also optimizes the features but not the positions.\footnote{The main differences between these two models are that, in +LPO-F, the positions of particles are fixed during training, while in +GO, the positions of particles are changed in each iteration by random sampling (note that they are not trainable).}
Specifically, when PAC-NeRF-3v was used as the baseline, +LPO outperformed +GO with 18 wins, 5 draws, and 4 losses.
When PAC-NeRF-3v$^\dag$ was used as the baseline, +LPO outperformed +GO with 16 wins, 6 draws, and 5 losses.
These results confirm the importance of position optimization in the +LPO.

\smallskip\noindent
\textbf{II. Evaluation of physical identification.}
Next, we discuss the scores for each scene for physical identification (the accuracy of the physical property estimation after Algorithm~\ref{alg:iterative_optimization} was applied) in detail.
Table~\ref{tab:comparison_physics} in the main text summarizes the absolute differences between the ground truth and the estimated physical properties.
Tables~\ref{tab:comparison_physics_value1} and \ref{tab:comparison_physics_value2} summarize the physical property values when PAC-NeRF-3v and PAC-NeRF-3v$^\dag$ were used as the baselines, respectively.
These results are discussed from four perspectives.

\smallskip\noindent
\textit{(1) PAC-NeRF-3v/3v$^\dag$ vs. +LPO$^4$.}
+LPO$^4$ improved the physical identification of PAC-NeRF-3v/3v$^\dag$ in most cases.
Specifically, when PAC-NeRF-3v was used as the baseline, +LPO$^4$ outperformed the baseline for 21 of the 23 evaluation items.
When PAC-NeRF-3v$^\dag$ was used as the baseline, +LPO$^4$ outperformed the baseline for 21 of 23 evaluation items.
These results indicate that +LPO is effective for physical identification independent of the baseline models.
The qualitative comparisons are shown in Figures~\ref{fig:results_newtonian}--\ref{fig:results_sand}.

\smallskip\noindent
\textit{(2) +LPO$^4$ vs. +LPO-F$^4$/P$^4$.}
When comparing these models, we found that superiority depends on the physical properties.
This is because the physical properties interact, and finding the optimal balance for performance is challenging.
However, +LPO-F$^4$/P$^4$ sometimes encountered apparent difficulties.
For example, the difference in $\log_{10}(E)$ and that in $\nu$ on Bird was large when +LPO-F$^4$ was used with PAC-NeRF-3v and the difference in $\log_{10}(E)$ on Playdoh was large when +LPO-P$^4$ was used with PAC-NeRF-3v$^\dag$.
In contrast, +LPO$^4$ exhibited stable performance.
Owing to this stability, +LPO$^4$ outperformed +LPO-F$^4$ and +LPO-P$^4$ in most cases.
Specifically, when PAC-NeRF-3v was used as the baseline, +LPO$^4$ outperformed +LPO-F$^4$ for 18 items and outperformed +LPO-P$^4$ for 17 of the 23 evaluation items.
When PAC-NeRF-3v$^\dag$ was used as the baseline, +LPO$^4$ outperformed +LPO-F$^4$ for 13 items and outperformed +LPO-P$^4$ for 16 of the 23 evaluation items.
These results indicate that the joint optimization of features and positions in Lagrangian space is useful for obtaining stability and tackling difficult situations.

\smallskip\noindent
\textit{(3) +LPO$^4$ vs. +GO$^4$.}
+GO$^4$ also sometimes suffers from critical difficulties.
For example, the difference in $\log_{10}(E)$ and that in $\nu$ on Bird was large when +GO$^4$ was used with PAC-NeRF-3v.
In contrast, +LPO$^4$ exhibited stable performance and outperformed +GO$^4$ in most cases.
Specifically, when PAC-NeRF-3v was used as the baseline, +LPO$^4$ outperformed +GO$^4$ for 18 of the 23 evaluation items.
When PAC-NeRF-3v$^\dag$ was used as the baseline, +LPO$^4$ outperformed +GO$^4$ for 18 of the 23 evaluation items.

\smallskip\noindent
\textit{(4) +LPO$^4$ vs. +None$^4$.}
+None$^4$ is outperformed by the dynamic optimization methods (i.e., +LPO$^4$, +LPO-F$^4$, +LPO-P$^4$, and +GO$^4$) in most cases.
In particular, when PAC-NeRF-3v was used as the baseline, +LPO$^4$ outperformed +None$^4$ for 20 of the 23 evaluation items.
When PAC-NeRF-3v$^\dag$ was used as the baseline, +LPO$^4$ outperformed +None$^4$ for 19 of the 23 evaluation items.
These results indicate that simple iterative updates without geometric correction in Algorithm~\ref{alg:iterative_optimization} are insufficient to improve physical identification and that it is crucial to correct geometric structures using a dynamic optimization method.

\smallskip\noindent
\textbf{III. Evaluation of geometric recorrection.}
Finally, we discuss the scores for each scene in the geometry recorrection (image quality after Algorithm~\ref{alg:iterative_optimization} was applied) in detail.
Table~\ref{tab:comparison_geometry4_ex} summarizes these results.
This is an extension of the results in Table~\ref{tab:comparison_geometry4}.
We observed tendencies similar to those for geometry correction.
Specifically, +LPO$^4$ outperformed not only the baselines (PAC-NeRF-3v and PAC-NeRF-3v$^\dag$) but also the ablated and comparative models, including +LPO-F$^4$, +LPO-P$^4$, +GO$^4$, and +None$^4$, in most cases.
These results indicate that joint optimization of the features and positions of particles in Lagrangian space is essential not only when geometric correction is conducted once but also when geometric correction is repeatedly conducted along with physical reidentification through Algorithm~\ref{alg:iterative_optimization}.
Qualitative comparisons of PAC-NeRF-3v/3v$^\dag$, +LPO, and +LPO$^4$ are shown in Figure~\ref{fig:results_newtonian}--\ref{fig:results_sand}.
Qualitative comparisons of +LPO$^4$, +LPO-F$^4$, +LPO-P$^4$, +GO$^4$, and +None$^4$ are presented in Figures~\ref{fig:examples_ptcl} and \ref{fig:examples_feat}.

\subsection{Computation times}
\label{subsec:computation_times}

Table~\ref{tab:computation_times} lists the computation times for executing Algorithm~\ref{alg:iterative_optimization} for one iteration on Droplet with PAC-NeRF-3v+LPO.\footnote{Computation times vary with scenes because of the difference in number of frames and object size (which affects the number of particles); however, we observed similar tendencies.}
The total computation time increases linearly when repeatedly running Algorithm~\ref{alg:iterative_optimization}.
However, it is adjustable under a quality-and-time trade-off, as discussed in Appendix~\ref{subsec:impact_iterations}.
The computation time of LPO (3) is almost identical to that of the main process of physical property optimization (2-c) because the forward and backward processes are identical with different optimization targets, as shown in Figures~\ref{fig:pipelines}(2) and (3).
Similarly, the calculation times for +LPO-F, +LPO-P, and +GO were almost identical to those for +LPO, indicating that the performance improvement was attributable to the ingenuity of the algorithm and not to an increase in calculation cost.

\begin{table}[t]
  \centering
  \footnotesize
  \setlength{\tabcolsep}{5pt}
  \setlength{\aboverulesep}{0.5pt}
  \setlength{\belowrulesep}{0.5pt}
  \begin{tabularx}{\columnwidth}{lcr}
    \toprule
    \multicolumn{1}{c}{Process} & \#~Frames & Time (s)
    \\ \midrule
    (1) Eulerian static voxel grid optimization
                                & 1 & 284.6
    \\
    (2) Physical property optimization
    \\
    \: (a) Velocity optimization
                                & 4 & 101.9
    \\
    \: (b) Warm-up optimization with partial frames
                                & 7 & 765.2
    \\
    \: (c) Main optimization with entire frames
                                & 13 & 3153.3
    \\
    (3) Lagrangian particle optimization
                                & 13 & 3225.4
    \\
    (4) Color prediction
                                & 1 & 1.0
    \\ \bottomrule
  \end{tabularx}
  \vspace{-2mm}
  \caption{Computation times of Algorithm~\ref{alg:iterative_optimization} on NVIDIA A100 GPU.
    (1) and (2) are processes driven from the original PAC-NeRF.
    (3) and (4) are processes newly introduced.}
  \label{tab:computation_times}
\end{table}

\begin{table*}[t]
  \centering
  \scriptsize
  \setlength{\tabcolsep}{0pt}
  \begin{tabularx}{\textwidth}{CCCCCCCCCCCC}
    \toprule
    & & PAC-NeRF-3v & +LPO$^1$ & +LPO$^2$ & +LPO$^3$ & +LPO$^4$ & \!\!PAC-NeRF-3v$^\dag$\!\! & +LPO$^1$ & +LPO$^2$ & +LPO$^3$ & +LPO$^4$
    \\ \cmidrule(lr){1-1} \cmidrule(lr){2-2} \cmidrule(lr){3-7} \cmidrule(lr){8-12}
    \multirow{3}{*}{Droplet}
    & PSNR$\uparrow$
      & 25.42 & 27.56 & \second{28.19} & \third{28.06} & \first{28.21} & 26.40 & 28.18 & \second{28.95} & \first{29.14} & \third{28.91}
    \\
    & SSIM$\uparrow$
      & 0.975 & \third{0.978} & \second{0.980} & \first{0.981} & \first{0.981} & 0.978 & \third{0.980} & \second{0.982} & \first{0.983} & \first{0.983}
    \\
    & LPIPS$\downarrow$
      & \third{0.047} & \second{0.043} & \second{0.043} & \first{0.042} & \first{0.042} & 0.046 & \third{0.044} & \second{0.042} & \first{0.041} & \first{0.041}
    \\ \cmidrule(lr){1-1} \cmidrule(lr){2-2} \cmidrule(lr){3-7} \cmidrule(lr){8-12}
    \multirow{3}{*}{Letter}
    & PSNR$\uparrow$
      & 28.94 & \second{29.99} & 29.90 & \third{29.96} & \first{30.07} & 29.59 & \third{30.44} & \first{30.76} & \second{30.70} & 30.26
    \\
    & SSIM$\uparrow$
      & \third{0.981} & \second{0.982} & \first{0.983} & \second{0.982} & \first{0.983} & \second{0.983} & \third{0.982} & \second{0.983} & \first{0.984} & \second{0.983}
    \\
    & LPIPS$\downarrow$
      & \second{0.028} & \third{0.029} & \second{0.027} & 0.031 & \first{0.026} & \second{0.028} & \third{0.029} & \first{0.026} & \first{0.026} & \first{0.026}
    \\ \cmidrule(lr){1-1} \cmidrule(lr){2-2} \cmidrule(lr){3-7} \cmidrule(lr){8-12}
    \multirow{3}{*}{Cream}
    & PSNR$\uparrow$
      & 26.61 & 28.48 & \third{29.18} & \second{29.38} & \first{29.65} & 27.43 & 28.82 & \third{29.69} & \second{29.93} & \first{30.23}
    \\
    & SSIM$\uparrow$
      & \third{0.983} & \third{0.983} & \second{0.984} & \first{0.985} & \first{0.985} & \third{0.983} & \second{0.984} & \first{0.985} & \first{0.985} & \first{0.985}
    \\
    & LPIPS$\downarrow$
      & \third{0.026} & \second{0.023} & \first{0.022} & \second{0.023} & \first{0.022} & 0.024 & \third{0.023} & \second{0.022} & \second{0.022} & \first{0.021}
    \\ \cmidrule(lr){1-1} \cmidrule(lr){2-2} \cmidrule(lr){3-7} \cmidrule(lr){8-12}
    \multirow{3}{*}{Toothpaste}
    & PSNR$\uparrow$
      & 29.23 & 31.27 & \third{31.78} & \second{31.93} & \first{32.18} & 31.72 & 33.54 & \first{34.16} & \second{34.04} & \third{34.00}
    \\
    & SSIM$\uparrow$
      & 0.991 & 0.991 & 0.991 & 0.991 & 0.991 & \first{0.993} & \first{0.993} & \first{0.993} & \second{0.992} & \second{0.992}
    \\
    & LPIPS$\downarrow$
      & \first{0.010} & \first{0.010} & \second{0.011} & \second{0.011} & \second{0.011} & 0.010 & 0.010 & 0.010 & 0.010 & 0.010
    \\ \cmidrule(lr){1-1} \cmidrule(lr){2-2} \cmidrule(lr){3-7} \cmidrule(lr){8-12}
    \multirow{3}{*}{Torus}
    & PSNR$\uparrow$
      & 23.99 & 28.91 & \third{29.87} & \first{30.20} & \second{30.07} & 26.65 & 30.05 & \second{30.90} & \first{30.97} & \third{30.48}
    \\
    & SSIM$\uparrow$
      & 0.970 & \third{0.978} & \second{0.981} & \first{0.982} & \first{0.982} & 0.974 & \third{0.980} & \second{0.983} & \first{0.984} & \second{0.983}
    \\
    & LPIPS$\downarrow$
      & 0.048 & \third{0.038} & \second{0.034} & \first{0.032} & \first{0.032} & 0.040 & \third{0.034} & \second{0.031} & \first{0.030} & \second{0.031}
    \\ \cmidrule(lr){1-1} \cmidrule(lr){2-2} \cmidrule(lr){3-7} \cmidrule(lr){8-12}
    \multirow{3}{*}{Bird}
    & PSNR$\uparrow$
      & 24.82 & \third{27.20} & 26.55 & \second{27.89} & \first{27.97} & 24.91 & 28.51 & \second{28.88} & \third{28.85} & \first{28.98}
    \\
    & SSIM$\uparrow$
      & 0.978 & \second{0.980} & \third{0.979} & \first{0.981} & \second{0.980} & \third{0.979} & \first{0.983} & \first{0.983} & \second{0.982} & \second{0.982}
    \\
    & LPIPS$\downarrow$
      & 0.036 & \first{0.032} & 0.035 & \second{0.033} & \third{0.034} & 0.036 & \first{0.028} & \second{0.030} & \third{0.031} & \third{0.031}
    \\ \cmidrule(lr){1-1} \cmidrule(lr){2-2} \cmidrule(lr){3-7} \cmidrule(lr){8-12}
    \multirow{3}{*}{Playdoh}
    & PSNR$\uparrow$
      & 27.70 & \third{28.39} & \second{29.14} & \first{29.32} & \first{29.32} & 29.66 & \third{30.07} & \second{30.16} & \first{30.26} & 30.01
    \\
    & SSIM$\uparrow$
      & \second{0.978} & \second{0.978} & \first{0.981} & \first{0.981} & \first{0.981} & \second{0.982} & \second{0.982} & \first{0.983} & \first{0.983} & \first{0.983}
    \\
    & LPIPS$\downarrow$
      & \third{0.042} & \second{0.041} & \first{0.040} & \first{0.040} & \first{0.040} & 0.038 & 0.038 & 0.038 & 0.038 & 0.038
    \\ \cmidrule(lr){1-1} \cmidrule(lr){2-2} \cmidrule(lr){3-7} \cmidrule(lr){8-12}
    \multirow{3}{*}{Cat}
    & PSNR$\uparrow$
      & \third{30.45} & \first{31.00} & \second{30.64} & 30.21 & 29.81 & 31.11 & \first{31.41} & \third{31.25} & \second{31.29} & 30.61
    \\
    & SSIM$\uparrow$
      & \second{0.987} & \second{0.987} & \first{0.988} & \second{0.987} & \second{0.987} & \second{0.988} & \second{0.988} & \first{0.989} & \first{0.989} & \second{0.988}
    \\
    & LPIPS$\downarrow$
      & 0.033 & \third{0.032} & \first{0.027} & \first{0.027} & \second{0.028} & 0.034 & 0.032 & \third{0.027} & \first{0.024} & \second{0.025}
    \\ \cmidrule(lr){1-1} \cmidrule(lr){2-2} \cmidrule(lr){3-7} \cmidrule(lr){8-12}
    \multirow{3}{*}{Trophy}
    & PSNR$\uparrow$
      & 29.33 & \first{30.16} & \second{29.99} & \third{29.78} & 29.57 & 28.80 & \second{29.92} & \first{29.97} & \third{29.82} & 29.58
    \\
    & SSIM$\uparrow$
      & \second{0.963} & \first{0.964} & \first{0.964} & \first{0.964} & \first{0.964} & \second{0.962} & \first{0.964} & \first{0.964} & \first{0.964} & \first{0.964}
    \\
    & LPIPS$\downarrow$
      & \second{0.039} & \first{0.037} & \first{0.037} & \first{0.037} & \first{0.037} & \second{0.039} & \third{0.037} & \third{0.037} & \first{0.036} & \first{0.036}
    \\ \cmidrule(lr){1-1} \cmidrule(lr){2-2} \cmidrule(lr){3-7} \cmidrule(lr){8-12}
    \multirow{3}{*}{\texttt{Average}}
    & PSNR$\uparrow$
      & 27.39 & 29.22 & \third{29.47} & \second{29.64} & \first{29.65} & 28.47 & 30.11 & \second{30.52} & \first{30.56} & \third{30.34}
    \\
    & SSIM$\uparrow$
      & 0.978 & \third{0.980} & \second{0.981} & \second{0.981} & \first{0.982} & \third{0.980} & \second{0.982} & \first{0.983} & \first{0.983} & \first{0.983}
    \\
    & LPIPS$\downarrow$
      & 0.034 & \third{0.032} & \second{0.031} & \second{0.031} & \first{0.030} & \third{0.033} & \second{0.031} & \first{0.029} & \first{0.029} & \first{0.029}
    \\ \bottomrule
  \end{tabularx}
  \vspace{-2mm}
  \caption{Comparison of PSNR$\uparrow$, SSIM$\uparrow$, and LPIPS$\downarrow$ on \textbf{geometric (re)correction} when the number of iterations in Algorithm~\ref{alg:iterative_optimization} is changed.
    The scores were calculated using the images in the \textit{test} set.}
  \label{tab:comparison_geometry1-4}
\end{table*}

\begin{table*}[t]
  \centering
  \scriptsize
  \setlength{\tabcolsep}{0pt}
  \begin{tabularx}{\textwidth}{CCCCCCCCCCCC}
    \toprule
    & & PAC-NeRF-3v & +LPO$^1$ & +LPO$^2$ & +LPO$^3$ & +LPO$^4$ & \!\!PAC-NeRF-3v$^\dag$\!\! & +LPO$^1$ & +LPO$^2$ & +LPO$^3$ & +LPO$^4$
    \\ \cmidrule(lr){1-1} \cmidrule(lr){2-2} \cmidrule(lr){3-7} \cmidrule(lr){8-12}
    \multirow{3}{*}{Droplet}
    & PSNR$\uparrow$
      & 32.48 & 36.61 & \third{36.97} & \second{37.12} & \first{37.41} & 32.31 & 36.09 & \third{37.21} & \second{37.83} & \first{37.96}
    \\
    & SSIM$\uparrow$
      & 0.985 & \third{0.990} & \second{0.991} & \second{0.991} & \first{0.992} & \third{0.985} & \second{0.990} & \first{0.992} & \first{0.992} & \first{0.992}
    \\
    & LPIPS$\downarrow$
      & \third{0.034} & \second{0.029} & \second{0.029} & \second{0.029} & \first{0.028} & 0.036 & \third{0.033} & \second{0.029} & \first{0.027} & \first{0.027}
    \\ \cmidrule(lr){1-1} \cmidrule(lr){2-2} \cmidrule(lr){3-7} \cmidrule(lr){8-12}
    \multirow{3}{*}{Letter}
    & PSNR$\uparrow$
      & 32.57 & 35.89 & \third{38.74} & \second{37.62} & \first{39.59} & 32.92 & 36.84 & \third{38.81} & \second{39.09} & \first{39.90}
    \\
    & SSIM$\uparrow$
      & 0.986 & 0.990 & \second{0.994} & \third{0.992} & \first{0.995} & 0.986 & \third{0.992} & \second{0.994} & \first{0.995} & \first{0.995}
    \\
    & LPIPS$\downarrow$
      & 0.020 & 0.020 & \second{0.015} & \third{0.019} & \first{0.013} & 0.021 & 0.019 & \third{0.015} & \second{0.013} & \first{0.012}
    \\ \cmidrule(lr){1-1} \cmidrule(lr){2-2} \cmidrule(lr){3-7} \cmidrule(lr){8-12}
    \multirow{3}{*}{Cream}
    & PSNR$\uparrow$
      & 34.91 & \third{38.45} & \second{39.50} & 37.90 & \first{40.11} & 35.18 & 38.50 & \third{38.70} & \second{39.60} & \first{40.39}
    \\
    & SSIM$\uparrow$
      & 0.991 & 0.990 & \second{0.994} & \third{0.992} & \first{0.995} & 0.990 & \third{0.993} & \second{0.994} & \first{0.995} & \first{0.995}
    \\
    & LPIPS$\downarrow$
      & \third{0.016} & \second{0.014} & \first{0.013} & \second{0.014} & \first{0.013} & 0.017 & \third{0.014} & \third{0.014} & \second{0.013} & \first{0.012}
    \\ \cmidrule(lr){1-1} \cmidrule(lr){2-2} \cmidrule(lr){3-7} \cmidrule(lr){8-12}
    \multirow{3}{*}{Toothpaste}
    & PSNR$\uparrow$
      & 37.41 & 41.49 & \third{42.10} & \second{42.54} & \first{42.91} & 37.36 & 41.43 & \third{42.56} & \second{43.07} & \first{43.41}
    \\
    & SSIM$\uparrow$
      & \second{0.996} & \first{0.997} & \first{0.997} & \first{0.997} & \first{0.997} & \third{0.996} & \second{0.997} & \second{0.997} & \second{0.997} & \first{0.998}
    \\
    & LPIPS$\downarrow$
      & \third{0.006} & \second{0.005} & \second{0.005} & \second{0.005} & \first{0.004} & \third{0.006} & \second{0.005} & \second{0.005} & \first{0.004} & \first{0.004}
    \\ \cmidrule(lr){1-1} \cmidrule(lr){2-2} \cmidrule(lr){3-7} \cmidrule(lr){8-12}
    \multirow{3}{*}{Torus}
    & PSNR$\uparrow$
      & 28.45 & 34.29 & \third{35.96} & \second{36.75} & \first{37.20} & 31.70 & 35.30 & \third{36.66} & \second{37.22} & \first{37.56}
    \\
    & SSIM$\uparrow$
      & 0.977 & \third{0.986} & \second{0.989} & \first{0.990} & \first{0.990} & \third{0.981} & \second{0.987} & \first{0.990} & \first{0.990} & \first{0.990}
    \\
    & LPIPS$\downarrow$
      & 0.041 & \third{0.031} & \second{0.028} & \first{0.027} & \first{0.027} & 0.034 & \third{0.029} & \second{0.027} & \first{0.026} & \first{0.026}
    \\ \cmidrule(lr){1-1} \cmidrule(lr){2-2} \cmidrule(lr){3-7} \cmidrule(lr){8-12}
    \multirow{3}{*}{Bird}
    & PSNR$\uparrow$
      & 30.72 & \third{33.22} & 32.06 & \first{34.91} & \second{34.88} & 30.97 & 33.84 & \third{34.73} & \second{35.47} & \first{35.91}
    \\
    & SSIM$\uparrow$
      & 0.985 & \second{0.987} & \third{0.986} & \first{0.990} & \first{0.990} & 0.987 & \third{0.989} & \second{0.990} & \second{0.990} & \first{0.991}
    \\
    & LPIPS$\downarrow$
      & \third{0.029} & \second{0.027} & 0.030 & \first{0.025} & \first{0.025} & 0.028 & 0.026 & \third{0.025} & \second{0.024} & \first{0.023}
    \\ \cmidrule(lr){1-1} \cmidrule(lr){2-2} \cmidrule(lr){3-7} \cmidrule(lr){8-12}
    \multirow{3}{*}{Playdoh}
    & PSNR$\uparrow$
      & 34.50 & 40.38 & \third{41.47} & \second{41.23} & \first{41.80} & 35.51 & 40.92 & \third{41.22} & \first{41.93} & \second{41.74}
    \\
    & SSIM$\uparrow$
      & \third{0.988} & \second{0.993} & \first{0.994} & \first{0.994} & \first{0.994} & \third{0.989} & \second{0.993} & \first{0.994} & \first{0.994} & \first{0.994}
    \\
    & LPIPS$\downarrow$
      & \third{0.027} & \second{0.022} & \second{0.022} & \first{0.021} & \first{0.021} & 0.027 & \third{0.023} & \second{0.022} & \first{0.021} & \first{0.021}
    \\ \cmidrule(lr){1-1} \cmidrule(lr){2-2} \cmidrule(lr){3-7} \cmidrule(lr){8-12}
    \multirow{3}{*}{Cat}
    & PSNR$\uparrow$
      & 37.25 & 41.34 & \third{41.77} & \second{42.37} & \first{42.94} & 36.66 & 40.43 & \third{42.17} & \second{43.32} & \first{43.49}
    \\
    & SSIM$\uparrow$
      & \third{0.993} & \second{0.996} & \second{0.996} & \first{0.997} & \first{0.997} & 0.993 & \third{0.995} & \second{0.996} & \first{0.997} & \first{0.997}
    \\
    & LPIPS$\downarrow$
      & 0.020 & \third{0.018} & \second{0.017} & \first{0.016} & \first{0.016} & 0.026 & \third{0.024} & \second{0.018} & \first{0.014} & \first{0.014}
    \\ \cmidrule(lr){1-1} \cmidrule(lr){2-2} \cmidrule(lr){3-7} \cmidrule(lr){8-12}
    \multirow{3}{*}{Trophy}
    & PSNR$\uparrow$
      & 32.35 & 34.76 & \third{34.82} & \second{35.01} & \first{35.13} & 31.86 & 34.54 & \third{34.78} & \second{35.00} & \first{35.19}
    \\
    & SSIM$\uparrow$
      & \third{0.966} & \second{0.969} & \second{0.969} & \first{0.970} & \first{0.970} & \third{0.966} & \second{0.969} & \second{0.969} & \first{0.970} & \first{0.970}
    \\
    & LPIPS$\downarrow$
      & 0.033 & 0.031 & 0.031 & 0.031 & 0.031 & \third{0.035} & \second{0.032} & \second{0.032} & \first{0.031} & \first{0.031}
    \\ \cmidrule(lr){1-1} \cmidrule(lr){2-2} \cmidrule(lr){3-7} \cmidrule(lr){8-12}
    \multirow{3}{*}{\texttt{Average}}
    & PSNR$\uparrow$
      & 33.41 & 37.38 & \third{38.16} & \second{38.38} & \first{39.11} & 33.83 & 37.54 & \third{38.54} & \second{39.17} & \first{39.51}
    \\
    & SSIM$\uparrow$
      & 0.985 & \third{0.989} & \second{0.990} & \second{0.990} & \first{0.991} & \third{0.986} & \second{0.990} & \first{0.991} & \first{0.991} & \first{0.991}
    \\
    & LPIPS$\downarrow$
      & 0.025 & \third{0.022} & \second{0.021} & \second{0.021} & \first{0.020} & 0.026 & \third{0.023} & \second{0.021} & \first{0.019} & \first{0.019}
    \\ \bottomrule
  \end{tabularx}
  \vspace{-2mm}
  \caption{Comparison of PSNR$\uparrow$, SSIM$\uparrow$, and LPIPS$\downarrow$ on \textbf{geometric (re)correction} when the number of iterations in Algorithm~\ref{alg:iterative_optimization} is changed.
    The scores were calculated using the images in the \textit{training} set.}
  \vspace{2mm}
  \label{tab:comparison_geometry1-4_train}
\end{table*}

\begin{table*}[t]
  \centering
  \scriptsize
  \setlength{\tabcolsep}{2pt}
  \begin{tabularx}{\textwidth}{ccCCCCCCCCC}
    \toprule
    & & PAC-NeRF-3v & +LPO$^2$ & +LPO$^3$ & +LPO$^4$ & \!PAC-NeRF-3v$^\dag$\! & +LPO$^2$ & +LPO$^3 $ & +LPO$^4$
    \\ \cmidrule(lr){1-1} \cmidrule(lr){2-2} \cmidrule(lr){3-6} \cmidrule(lr){7-10}
    \multirow{2}{*}{Droplet}
    & $\log_{10}(\mu)$
      & 0.140 & \third{0.136} & \first{0.111} & \second{0.112} & 0.136 & \third{0.129} & \second{0.084} & \first{0.082}
    \\
    & $\log_{10}(\kappa)$
      & \first{1.285} & \second{1.447} & 1.784 & \third{1.628} & 1.263 & \third{1.053} & \first{0.097} & \second{0.106}
    \\ \cmidrule(lr){1-1} \cmidrule(lr){2-2} \cmidrule(lr){3-6} \cmidrule(lr){7-10}
    \multirow{2}{*}{Letter}
    & $\log_{10}(\mu)$
      & \third{0.674} & \second{0.023} & 1.264 & \first{0.015} & 0.379 & \second{0.013} & \third{0.053} & \first{0.010}
    \\
    & $\log_{10}(\kappa)$
      & 6.772 & \second{0.325} & \third{1.424} & \first{0.174} & 5.229 & \second{0.507} & \third{0.589} & \first{0.060}
    \\ \cmidrule(lr){1-1} \cmidrule(lr){2-2} \cmidrule(lr){3-6} \cmidrule(lr){7-10}
    \multirow{4}{*}{Cream}
    & $\log_{10}(\mu)$
      & 0.311 & \third{0.200} & \first{0.115} & \second{0.178} & 0.179 & \first{0.031} & \second{0.066} & \third{0.100}
    \\
    & $\log_{10}(\kappa)$
      & \second{0.215} & \third{0.384} & 0.392 & \first{0.158} & 0.336 & \third{0.157} & \first{0.060} & \second{0.121}
    \\
    & $\log_{10}(\tau_Y)$
      & \second{0.014} & 0.032 & \third{0.031} & \first{0.004} & 0.009 & \third{0.007} & \first{0.005} & \second{0.006}
    \\
    & $\log_{10}(\eta)$
      & 0.281 & \third{0.209} & \second{0.198} & \first{0.183} & 0.195 & \third{0.105} & \first{0.026} & \second{0.033}
    \\ \cmidrule(lr){1-1} \cmidrule(lr){2-2} \cmidrule(lr){3-6} \cmidrule(lr){7-10}
    \multirow{4}{*}{Toothpaste}
    & $\log_{10}(\mu)$
      & 1.891 & \third{0.264} & \second{0.246} & \first{0.109} & \third{0.252} & 0.259 & \first{0.026} & \second{0.031}
    \\
    & $\log_{10}(\kappa)$
      & 1.580 & \third{1.439} & \second{1.434} & \first{0.601} & 1.436 & \third{1.382} & \second{0.888} & \first{0.673}
    \\
    & $\log_{10}(\tau_Y)$
      & 0.201 & \third{0.118} & \second{0.116} & \first{0.114} & 0.199 & \third{0.168} & \second{0.137} & \first{0.093}
    \\
    & $\log_{10}(\eta)$
      & 0.373 & \third{0.200} & \second{0.180} & \first{0.003} & 0.212 & \third{0.187} & \first{0.005} & \second{0.009}
    \\ \cmidrule(lr){1-1} \cmidrule(lr){2-2} \cmidrule(lr){3-6} \cmidrule(lr){7-10}
    \multirow{2}{*}{Torus}
    & $\log_{10}(E)$
      & 0.277 & \first{0.055} & \second{0.053} & \third{0.061} & 0.074 & \third{0.049} & \second{0.040} & \first{0.036}
    \\
    & $\nu$
      & \third{0.085} & 0.129 & \second{0.050} & \first{0.001} & 0.131 & \second{0.032} & \third{0.060} & \first{0.007}
    \\ \cmidrule(lr){1-1} \cmidrule(lr){2-2} \cmidrule(lr){3-6} \cmidrule(lr){7-10}
    \multirow{2}{*}{Bird}
    & $\log_{10}(E)$
      & 0.449 & \second{0.136} & \third{0.146} & \first{0.067} & \third{0.123} & 0.158 & \second{0.084} & \first{0.027}
    \\
    & $\nu$
      & \third{0.102} & 0.466 & \second{0.037} & \first{0.001} & 0.141 & \first{0.009} & \third{0.056} & \second{0.047}
    \\ \cmidrule(lr){1-1} \cmidrule(lr){2-2} \cmidrule(lr){3-6} \cmidrule(lr){7-10}
    \multirow{3}{*}{Playdoh}
    & $\log_{10}(E)$
      & \third{0.290} & 0.403 & \second{0.190} & \first{0.116} & 0.521 & \third{0.303} & \second{0.260} & \first{0.133}
    \\
    & $\log_{10}(\tau_Y)$
      & 0.283 & \first{0.075} & \third{0.215} & \second{0.165} & \second{0.110} & \third{0.119} & \first{0.105} & 0.173
    \\
    & $\nu$
      & 0.495 & \first{0.092} & \third{0.236} & \second{0.111} & \third{0.212} & 0.218 & \second{0.082} & \first{0.063}
    \\ \cmidrule(lr){1-1} \cmidrule(lr){2-2} \cmidrule(lr){3-6} \cmidrule(lr){7-10}
    \multirow{3}{*}{Cat}
    & $\log_{10}(E)$
      & 1.301 & \third{1.120} & \second{1.051} & \first{0.973} & 1.192 & \first{0.584} & \second{0.687} & \third{0.706}
    \\
    & $\log_{10}(\tau_Y)$
      & 0.120 & \third{0.119} & \second{0.111} & \first{0.107} & 0.084 & \first{0.062} & \third{0.070} & \second{0.067}
    \\
    & $\nu$
      & \second{0.044} & \third{0.079} & 0.104 & \first{0.004} & 0.118 & \third{0.054} & \second{0.024} & \first{0.003}
    \\ \cmidrule(lr){1-1} \cmidrule(lr){2-2} \cmidrule(lr){3-6} \cmidrule(lr){7-10}
    \multirow{1}{*}{Trophy}
    & $\theta_{fric}$ [rad]
      & \first{0.053} & 0.058 & \first{0.053} & \third{0.055} & \first{0.030} & \second{0.036} & 0.046 & \third{0.039}
    \\ \bottomrule
  \end{tabularx}
  \vspace{-2mm}
  \caption{Comparison of the absolute differences between the ground-truth and estimated physical properties on \textbf{physical identification} when the number of iterations in Algorithm~\ref{alg:iterative_optimization} is changed.
    The smaller the values, the better the performance.}
  \label{tab:comparison_physics1-4}
\end{table*}

\subsection{Impact of the number of iterations}
\label{subsec:impact_iterations}

In the main experiments, we fixed the number of iterations in Algorithm~\ref{alg:iterative_optimization}, that is, $R$, to four for simplicity and fair comparison.
However, it is interesting and important to investigate how the number of iterations affects performance.
To answer this question, we analyze the impact of the number of iterations in this appendix.

\smallskip\noindent
\textbf{Results.}
Table~\ref{tab:comparison_geometry1-4} summarizes the performance changes with geometry (re)correction when the number of iterations was changed.
This table calculates the scores using the images in the \textit{test} set.
Table~\ref{tab:comparison_geometry1-4_train} lists the scores calculated using images in the \textit{training} set.
We investigated these two cases to examine the trade-off between the feasible reconstruction of the training data and the generalization ability for the test data.
Table~\ref{tab:comparison_physics1-4} presents the performance changes with physical identification when the number of iterations is changed.

When PAC-NeRF-3v was used as the baseline, the performances for geometry (re)correction (for both the test and training sets) and physical identification were the best when the number of iterations was four in most cases.
Specifically, as listed in Table~\ref{tab:comparison_geometry1-4}, 6, 8, 13, and 18 itmes achieved the best scores among the 27 evaluation items when the number of iterations was one, two, three, and four, respectively, for geometry (re)correction for the test set.
When scores were the same, they were counted multiple times.
As listed in Table~\ref{tab:comparison_geometry1-4_train}, 1, 3, 11, and 25 items achieved the best scores among the 27 evaluation items when the number of iterations was one, two, three, and four, respectively, for geometry (re)correction for the training set.
As listed in Table~\ref{tab:comparison_physics1-4}, 2, 3, 3, and 16 items achieved the best scores among the 23 evaluation items when the numbers of iterations were one, two, three, and four for physical identification.

In contrast, when PAC-NeRF-3v$^\dag$ was used as the baseline, the performance of the geometry (re)correction for the test set was the best when the number of iterations was three, whereas that for the training set was the best when the number of iterations was four.
The performance of the physical identification was the best when the number of iterations was three or four.
Specifically, as listed in Table~\ref{tab:comparison_geometry1-4}, 5, 10, 15, and 10 items achieved the best scores among the 27 evaluation items when the number of iterations was one, two, three, and four, respectively, for geometry (re)correction for the test set.
As listed in Table~\ref{tab:comparison_geometry1-4_train}, 0, 4, 14, and 26 items achieved the best scores among the 27 evaluation items when the number of iterations was one, two, three, and four, respectively, for geometry (re)correction for the training set.
As listed in Table~\ref{tab:comparison_physics1-4}, 1, 4, 7, and 11 items achieved the best scores among the 23 evaluation items when the numbers of iterations were one, two, three, and four for physical identification.

We consider that these differences arise from differences in the initial static voxel grids.
PAC-NeRF-3v$^\dag$ had better static voxel grids; therefore, fewer updates were required to obtain the best performance (i.e., optimal geometric structures).
Note that it is generally not trivial to determine when to stop iterative updates because there is an intractable trade-off between faithful reproduction of training data and overfitting.
When PAC-NeRF-3v was used as the baseline, the performance improvement continued until four iterations because the geometric reconstruction was more influential than overfitting.
In contrast, when PAC-NeRF-3v$^\dag$ was used as the baseline, the performance improvement (particularly for geometric (re)correction for the test set) was saturated at three iterations because the performance achieved the upper bound in an earlier phase.
From this time on, the overfitting problem became non-negligible.
Conducting further studies on this topic and exploring an improved method (for example, determining the number of iterations adaptively during training) will be interesting future research topics.
Furthermore, the effect of the abovementioned trade-off on the physical identification performance is an open issue.
A detailed investigation of this will be an interesting topic for future research.

\begin{table*}[h]
  \centering
  \scriptsize
  \setlength{\tabcolsep}{0pt}
  \setlength{\aboverulesep}{0.5pt}
  \setlength{\belowrulesep}{0.5pt}
  \newcommand{\spm}[1]{{\scriptsize$\pm$#1}}
  \begin{tabularx}{\textwidth}{ccCCCCCCCCC}
    \toprule
    & PAC-NeRF-3v$^\dag$ & +LPO & +LPO-F & +LPO-P & +GO & +LPO$^4$ & +LPO-F$^4$ & +LPO-P$^4$ & +GO$^4$ & +None$^4$
    \\ \cmidrule(lr){1-1} \cmidrule(lr){2-2} \cmidrule(lr){3-6} \cmidrule(lr){7-11}
    PSNR$\uparrow$
    & 28.18\spm{.43} & \first{29.60\spm{.59}} & 28.89\spm{.45} & \second{29.40\spm{.55}} & 28.91\spm{.45} & \first{29.88\spm{.53}} & 29.34\spm{.51} & \second{29.81\spm{.54}} & 28.79\spm{.44} & 28.49\spm{.45}
    \\
    SSIM$\uparrow$
    & 0.979\spm{.001} & \first{0.981\spm{.001}} & \second{0.980\spm{.001}} & \first{0.981\spm{.001}} & \second{0.980\spm{.001}} & \first{0.982\spm{.001}} & \second{0.981\spm{.001}} & \first{0.982\spm{.001}} & \second{0.981\spm{.001}} & \second{0.981\spm{.001}}
    \\
    LPIPS$\downarrow$
    & 0.036\spm{.002} & \first{0.034\spm{.003}} & \second{0.035\spm{.002}} & \second{0.035\spm{.003}} & \second{0.035\spm{.002}} & \first{0.031\spm{.002}} & \second{0.032\spm{.002}} & 0.033\spm{.003} & 0.033\spm{.002} & 0.034\spm{.002}
    \\ \bottomrule
  \end{tabularx}
  \vspace{-2mm}
  \caption{Comparison of PSNR$\uparrow$, SSIM$\uparrow$, and LPIPS$\downarrow$ averaged over five view settings on \textbf{geometric correction} and \textbf{recorrection}.}
  \label{tab:comparison_selection_views}
\end{table*}

\section{Experiments on other view settings}
\label{sec:view_experiments}

In the main text (Section~\ref{sec:experiments}), we investigate the performance when three specific views are selected as training data.
In this appendix, experiments were conducted using other settings to investigate the versatility of the proposed method.
In particular, we investigated the robustness of the selection of views (Appendix~\ref{subsec:robustness_selection_views}) and the number of views (Appendix~\ref{subsec:robustness_number}).

\subsection{Robustness of the selection of views}
\label{subsec:robustness_selection_views}

In the main text (Section~\ref{sec:experiments}), we investigate the performance when three specific views are selected as training data.
We examined the performance when the training set was changed to investigate the robustness of the selection of views.
Specifically, we investigated performance when five different view sets were used for training.
The number of views used for training was fixed at three, and the remaining eight were used for testing.

\smallskip\noindent
\textbf{Compared models.}
We used \textit{PAC-NeRF-3v$^\dag$} as the baseline and applied \textit{+LPO} and \textit{+LPO$^4$}.
Furthermore, we examined the performances of the ablated and comparative models.
Specifically, in the evaluation of the geometry correction, we examined the performance when \textit{+LPO-F}, \textit{+LPO-P}, and \textit{+GO} were applied to the baseline.
In the evaluation of the physical identification and geometric recorrection, we examined the performance when \textit{+LPO-F$^4$}, \textit{+LPO-P$^4$}, \textit{+GO$^4$}, and \textit{+None$^4$} were applied to the baseline.

\smallskip\noindent
\textbf{Results.}
Table~\ref{tab:comparison_selection_views} summarizes the geometric correction and recorrection results.
We observed tendencies similar to those for the results when three specific views were used for training (see Appendix~\ref{subsec:each_scene}).
Specifically, with respect to geometry correction, +LPO outperformed not only the baseline (PAC-NeRF-3v$^\dag$) but was also superior to or comparable to the ablated and comparative models, including +LPO-F, +LPO-P, and +GO.
Similarly, with respect to geometry recorrection, +LPO$^4$ outperformed not only the baseline (PAC-NeRF-3v$^\dag$) but was also superior to or comparable to the ablated and comparative models, including +LPO-F$^4$, +LPO-P$^4$, +GO$^4$, and +None$^4$.
In terms of physical property identification, +LPO$^4$ outperformed PAC-NeRF-3v$^\dag$, +LPO-F$^4$, +LPO-P$^4$, +GO$^4$, and +None$^4$ in the 20.8$\pm$1.2, 15.2$\pm$2.0, 16.6$\pm$2.0, 17.8$\pm$1.2, and 18.8$\pm$1.9 cases, respectively, across the 23 properties for each view setting.
These results indicate that the joint optimization of features and positions in Lagrangian space is effective for geometry-agnostic system identification, regardless of the selection of views used for training.

\begin{table*}[t]
  \centering
  \scriptsize
  \setlength{\tabcolsep}{2pt}
  \begin{tabularx}{0.8\textwidth}{CCCCCCCC}
    \toprule
    & & PAC-NeRF & PAC-NeRF-6v & +LPO & +LPO-F & +LPO-P & +GO
    \\ \cmidrule(lr){1-1} \cmidrule(lr){2-2} \cmidrule(lr){3-3} \cmidrule(lr){4-8}
    \multirow{3}{*}{Droplet}
    & PSNR$\uparrow$
      & 35.29 & 28.92 & \first{30.58} & 29.82 & \second{30.47} & 29.13
    \\
    & SSIM$\uparrow$
      & 0.989 & 0.982 & \first{0.984} & \second{0.983} & \first{0.984} & \second{0.983}
    \\
    & LPIPS$\downarrow$
      & 0.030 & 0.042 & \first{0.040} & \second{0.041} & \first{0.040} & 0.042
    \\ \cmidrule(lr){1-1} \cmidrule(lr){2-2} \cmidrule(lr){3-3} \cmidrule(lr){4-8}
    \multirow{3}{*}{Letter}
    & PSNR$\uparrow$
      & 36.99 & 31.42 & \first{32.85} & 32.36 & \second{32.62} & 32.32
    \\
    & SSIM$\uparrow$
      & 0.992 & 0.985 & \first{0.987} & \first{0.987} & \second{0.986} & \second{0.986}
    \\
    & LPIPS$\downarrow$
      & 0.011 & \second{0.020} & \first{0.019} & \first{0.019} & \first{0.019} & \second{0.020}
    \\ \cmidrule(lr){1-1} \cmidrule(lr){2-2} \cmidrule(lr){3-3} \cmidrule(lr){4-8}
    \multirow{3}{*}{Cream}
    & PSNR$\uparrow$
      & 36.46 & 30.36 & \first{31.31} & \second{31.13} & 30.89 & 30.80
    \\
    & SSIM$\uparrow$
      & 0.993 & \second{0.986} & \first{0.987} & \first{0.987} & \second{0.986} & \first{0.987}
    \\
    & LPIPS$\downarrow$
      & 0.014 & 0.021 & \first{0.019} & \first{0.019} & \second{0.020} & \first{0.019}
    \\ \cmidrule(lr){1-1} \cmidrule(lr){2-2} \cmidrule(lr){3-3} \cmidrule(lr){4-8}
    \multirow{3}{*}{Toothpaste}
    & PSNR$\uparrow$
      & 38.84 & 34.74 & \first{35.82} & 35.27 & \second{35.58} & 34.79
    \\
    & SSIM$\uparrow$
      & 0.996 & \second{0.993} & \first{0.994} & \second{0.993} & \second{0.993} & \second{0.993}
    \\
    & LPIPS$\downarrow$
      & 0.006 & 0.009 & 0.009 & 0.009 & 0.009 & 0.009
    \\ \cmidrule(lr){1-1} \cmidrule(lr){2-2} \cmidrule(lr){3-3} \cmidrule(lr){4-8}
    \multirow{3}{*}{Torus}
    & PSNR$\uparrow$
      & 34.60 & 29.28 & \first{32.31} & 30.82 & \second{31.80} & 30.80
    \\
    & SSIM$\uparrow$
      & 0.988 & 0.979 & \first{0.984} & 0.981 & \first{0.984} & \second{0.982}
    \\
    & LPIPS$\downarrow$
      & 0.026 & 0.035 & \first{0.030} & \second{0.033} & \first{0.030} & 0.034
    \\ \cmidrule(lr){1-1} \cmidrule(lr){2-2} \cmidrule(lr){3-3} \cmidrule(lr){4-8}
    \multirow{3}{*}{Bird}
    & PSNR$\uparrow$
      & 35.70 & 27.38 & \first{30.48} & 28.67 & \second{30.39} & 28.09
    \\
    & SSIM$\uparrow$
      & 0.992 & 0.981 & \first{0.984} & \second{0.982} & \first{0.984} & 0.981
    \\
    & LPIPS$\downarrow$
      & 0.019 & 0.029 & \first{0.025} & \second{0.027} & \first{0.025} & 0.028
    \\ \cmidrule(lr){1-1} \cmidrule(lr){2-2} \cmidrule(lr){3-3} \cmidrule(lr){4-8}
    \multirow{3}{*}{Playdoh}
    & PSNR$\uparrow$
      & 36.55 & 29.74 & \first{31.01} & 29.93 & \second{30.99} & 29.84
    \\
    & SSIM$\uparrow$
      & 0.991 & \second{0.982} & \first{0.983} & \second{0.982} & \first{0.983} & 0.982
    \\
    & LPIPS$\downarrow$
      & 0.026 & \second{0.039} & \first{0.037} & \second{0.039} & \first{0.037} & \second{0.039}
    \\ \cmidrule(lr){1-1} \cmidrule(lr){2-2} \cmidrule(lr){3-3} \cmidrule(lr){4-8}
    \multirow{3}{*}{Cat}
    & PSNR$\uparrow$
      & 37.10 & 30.77 & \first{31.40} & 30.85 & \second{31.33} & 30.72
    \\
    & SSIM$\uparrow$
      & 0.993 & \first{0.989} & 0.987 & \first{0.989} & \second{0.988} & \second{0.988}
    \\
    & LPIPS$\downarrow$
      & 0.016 & \first{0.024} & 0.026 & \second{0.025} & 0.026 & \second{0.025}
    \\ \cmidrule(lr){1-1} \cmidrule(lr){2-2} \cmidrule(lr){3-3} \cmidrule(lr){4-8}
    \multirow{3}{*}{Trophy}
    & PSNR$\uparrow$
      & 32.23 & 30.22 & \first{30.97} & 30.38 & \second{30.85} & 30.23
    \\
    & SSIM$\uparrow$
      & 0.965 & 0.962 & \first{0.964} & \second{0.963} & \second{0.963} & 0.962
    \\
    & LPIPS$\downarrow$
      & 0.036 & 0.039 & \first{0.037} & 0.039 & \second{0.038} & 0.039
    \\ \cmidrule[0.1em](lr){1-1} \cmidrule[0.1em](lr){2-2} \cmidrule[0.1em](lr){3-3} \cmidrule[0.1em](lr){4-8}
    \multirow{3}{*}{\texttt{Average}}
    & PSNR$\uparrow$
      & 35.97 & 30.31 & \first{31.86} & 31.03 & \second{31.66} & 30.75
    \\
    & SSIM$\uparrow$
      & 0.989 & 0.982 & \first{0.984} & \second{0.983} & \first{0.984} & \second{0.983}
    \\
    & LPIPS$\downarrow$
      & 0.020 & 0.029 & \first{0.027} & \second{0.028} & \first{0.027} & \second{0.028}
    \\ \bottomrule
  \end{tabularx}
  \vspace{-2mm}
  \caption{Comparison of PSNR$\uparrow$, SSIM$\uparrow$, and LPIPS$\downarrow$ for each scene on \textbf{geometric correction} when the number of views in a training set was six.}
  \label{tab:comparison_geometry_view6}
\end{table*}

\begin{table*}[t]
  \centering
  \scriptsize
  \setlength{\tabcolsep}{2pt}
  \begin{tabularx}{0.8\textwidth}{ccCCCCCCC}
    \toprule
    & & PAC-NeRF & PAC-NeRF-6v & +LPO$^4$ & +LPO-F$^4$ & +LPO-P$^4$ & +GO$^4$ & +None$^4$
    \\ \cmidrule(lr){1-1} \cmidrule(lr){2-2} \cmidrule(lr){3-3} \cmidrule(lr){4-9}
    \multirow{2}{*}{Droplet}
    & $\log_{10}(\mu)$
      & 0.039 & 0.026 & \first{0.016} & 0.039 & \second{0.025} & 0.042 & 0.039
    \\
    & $\log_{10}(\kappa)$
      & 0.017 & 0.386 & \first{0.045} & 0.128 & \second{0.076} & 0.404 & 0.144
    \\ \cmidrule(lr){1-1} \cmidrule(lr){2-2} \cmidrule(lr){3-3} \cmidrule(lr){4-9}
    \multirow{2}{*}{Letter}
    & $\log_{10}(\mu)$
      & 0.041 & 0.229 & \first{0.175} & 0.220 & \second{0.199} & 0.222 & 0.224
    \\
    & $\log_{10}(\kappa)$
      & 0.039 & 0.166 & \first{0.006} & \second{0.021} & 0.060 & 0.064 & 0.089
    \\ \cmidrule(lr){1-1} \cmidrule(lr){2-2} \cmidrule(lr){3-3} \cmidrule(lr){4-9}
    \multirow{4}{*}{Cream}
    & $\log_{10}(\mu)$
      & 0.090 & 0.131 & \second{0.056} & \first{0.047} & 0.065 & 0.057 & 0.065
    \\
    & $\log_{10}(\kappa)$
      & 0.132 & 0.835 & 0.146 & \first{0.020} & 0.143 & \second{0.072} & 0.132
    \\
    & $\log_{10}(\tau_Y)$
      & 0.007 & 0.011 & \first{0.001} & \second{0.003} & \first{0.001} & \second{0.003} & 0.012
    \\
    & $\log_{10}(\eta)$
      & 0.015 & 0.287 & \first{0.049} & \second{0.060} & 0.075 & 0.088 & 0.134
    \\ \cmidrule(lr){1-1} \cmidrule(lr){2-2} \cmidrule(lr){3-3} \cmidrule(lr){4-9}
    \multirow{4}{*}{Toothpaste}
    & $\log_{10}(\mu)$
      & 0.026 & \second{0.084} & \first{0.045} & 0.179 & 0.144 & 0.208 & 0.166
    \\
    & $\log_{10}(\kappa)$
      & 0.247 & 0.629 & 0.580 & \first{0.370} & 0.655 & 0.654 & \second{0.573}
    \\
    & $\log_{10}(\tau_Y)$
      & 0.066 & 0.088 & \first{0.036} & 0.090 & \second{0.062} & 0.101 & 0.097
    \\
    & $\log_{10}(\eta)$
      & 0.013 & \second{0.015} & \first{0.006} & 0.022 & \first{0.006} & 0.018 & 0.018
    \\ \cmidrule(lr){1-1} \cmidrule(lr){2-2} \cmidrule(lr){3-3} \cmidrule(lr){4-9}
    \multirow{2}{*}{Torus}
    & $\log_{10}(E)$
      & 0.019 & \second{0.020} & 0.023 & \first{0.019} & 0.025 & 0.032 & 0.025
    \\
    & $\nu$
      & 0.023 & \second{0.024} & \first{0.001} & 0.057 & 0.035 & 0.139 & 0.058
    \\ \cmidrule(lr){1-1} \cmidrule(lr){2-2} \cmidrule(lr){3-3} \cmidrule(lr){4-9}
    \multirow{2}{*}{Bird}
    & $\log_{10}(E)$
      & 0.013 & 0.106 & \first{0.047} & 0.105 & \second{0.058} & 0.119 & 1.769
    \\
    & $\nu$
      & 0.029 & \second{0.133} & \first{0.013} & 0.134 & 0.147 & 0.146 & 0.741
    \\ \cmidrule(lr){1-1} \cmidrule(lr){2-2} \cmidrule(lr){3-3} \cmidrule(lr){4-9}
    \multirow{3}{*}{Playdoh}
    & $\log_{10}(E)$
      & 0.286 & 0.483 & \first{0.252} & 0.311 & \second{0.282} & 0.310 & 0.289
    \\
    & $\log_{10}(\tau_Y)$
      & 0.038 & 0.209 & \first{0.126} & 0.203 & \second{0.171} & 0.184 & 0.180
    \\
    & $\nu$
      & 0.076 & 0.317 & \first{0.110} & 0.135 & 0.141 & \second{0.130} & 0.193
    \\ \cmidrule(lr){1-1} \cmidrule(lr){2-2} \cmidrule(lr){3-3} \cmidrule(lr){4-9}
    \multirow{3}{*}{Cat}
    & $\log_{10}(E)$
      & 0.855 & 2.821 & \second{0.787} & 0.949 & 0.843 & 2.423 & \first{0.745}
    \\
    & $\log_{10}(\tau_Y)$
      & 0.026 & 0.978 & \second{0.008} & \first{0.005} & 0.024 & 0.393 & 0.020
    \\
    & $\nu$
      & 0.027 & 0.071 & \first{0.017} & 0.065 & \second{0.026} & 0.218 & \second{0.026}
    \\ \cmidrule(lr){1-1} \cmidrule(lr){2-2} \cmidrule(lr){3-3} \cmidrule(lr){4-9}
    \multirow{1}{*}{Trophy}
    & $\theta_{fric}$ [rad]
      & 0.048 & 0.052 & \first{0.035} & \second{0.040} & 0.041 & 0.044 & 0.044
    \\ \bottomrule
  \end{tabularx}
  \vspace{-2mm}
  \caption{Comparison of the absolute differences between the ground-truth and estimated physical properties for each scene on \textbf{physical identification} when the number of views in a training set was six.
    The smaller the values, the better the performance.}
  \label{tab:comparison_physics_view6}
\end{table*}

\begin{table*}[t]
  \centering
  \scriptsize
  \setlength{\tabcolsep}{2pt}
  \begin{tabularx}{0.85\textwidth}{CCCCCCCCC}
    \toprule
    & & PAC-NeRF & PAC-NeRF-6v & +LPO$^4$ & +LPO-F$^4$ & +LPO-P$^4$ & +GO$^4$ & +None$^4$
    \\ \cmidrule(lr){1-1} \cmidrule(lr){2-2} \cmidrule(lr){3-3} \cmidrule(lr){4-9}
    \multirow{3}{*}{Droplet}
    & PSNR$\uparrow$
      & 35.29 & 28.92 & \second{30.72} & 29.64 & \first{31.26} & 27.66 & 28.73
    \\
    & SSIM$\uparrow$
      & 0.989 & 0.982 & \first{0.986} & \second{0.984} & \first{0.986} & 0.981 & 0.983
    \\
    & LPIPS$\downarrow$
      & 0.030 & 0.042 & \first{0.037} & 0.041 & \second{0.039} & 0.044 & 0.042
    \\ \cmidrule(lr){1-1} \cmidrule(lr){2-2} \cmidrule(lr){3-3} \cmidrule(lr){4-9}
    \multirow{3}{*}{Letter}
    & PSNR$\uparrow$
      & 36.99 & 31.42 & \first{32.70} & \second{32.69} & 32.66 & 32.13 & 32.25
    \\
    & SSIM$\uparrow$
      & 0.992 & 0.985 & \first{0.989} & \first{0.989} & \second{0.988} & 0.987 & 0.987
    \\
    & LPIPS$\downarrow$
      & 0.011 & 0.020 & \second{0.017} & \first{0.016} & 0.018 & 0.018 & 0.018
    \\ \cmidrule(lr){1-1} \cmidrule(lr){2-2} \cmidrule(lr){3-3} \cmidrule(lr){4-9}
    \multirow{3}{*}{Cream}
    & PSNR$\uparrow$
      & 36.46 & 30.36 & \first{32.11} & \second{31.78} & 31.48 & 30.65 & 30.34
    \\
    & SSIM$\uparrow$
      & 0.993 & 0.986 & \first{0.988} & \first{0.988} & \second{0.987} & \second{0.987} & \second{0.987}
    \\
    & LPIPS$\downarrow$
      & 0.014 & 0.021 & \first{0.018} & \second{0.019} & 0.020 & \second{0.019} & 0.020
    \\ \cmidrule(lr){1-1} \cmidrule(lr){2-2} \cmidrule(lr){3-3} \cmidrule(lr){4-9}
    \multirow{3}{*}{Toothpaste}
    & PSNR$\uparrow$
      & 38.84 & 34.74 & \second{34.91} & 34.44 & \first{35.35} & 33.52 & 34.84
    \\
    & SSIM$\uparrow$
      & 0.996 & \second{0.993} & \first{0.994} & \first{0.994} & \first{0.994} & \second{0.993} & \first{0.994}
    \\
    & LPIPS$\downarrow$
      & 0.006 & 0.009 & 0.009 & 0.009 & 0.009 & 0.009 & 0.009
    \\ \cmidrule(lr){1-1} \cmidrule(lr){2-2} \cmidrule(lr){3-3} \cmidrule(lr){4-9}
    \multirow{3}{*}{Torus}
    & PSNR$\uparrow$
      & 34.60 & 29.28 & \first{32.33} & 31.57 & \second{31.94} & 30.64 & 29.58
    \\
    & SSIM$\uparrow$
      & 0.988 & 0.979 & \first{0.986} & \second{0.984} & \first{0.986} & 0.982 & 0.982
    \\
    & LPIPS$\downarrow$
      & 0.026 & 0.035 & \first{0.028} & 0.031 & \second{0.030} & 0.032 & 0.034
    \\ \cmidrule(lr){1-1} \cmidrule(lr){2-2} \cmidrule(lr){3-3} \cmidrule(lr){4-9}
    \multirow{3}{*}{Bird}
    & PSNR$\uparrow$
      & 35.70 & 27.38 & \first{30.57} & 28.27 & \second{29.67} & 27.84 & 25.96
    \\
    & SSIM$\uparrow$
      & 0.992 & 0.981 & \first{0.985} & 0.981 & \second{0.983} & 0.981 & \second{0.983}
    \\
    & LPIPS$\downarrow$
      & 0.019 & \second{0.029} & \first{0.028} & \second{0.029} & \first{0.028} & 0.030 & 0.035
    \\ \cmidrule(lr){1-1} \cmidrule(lr){2-2} \cmidrule(lr){3-3} \cmidrule(lr){4-9}
    \multirow{3}{*}{Playdoh}
    & PSNR$\uparrow$
      & 36.55 & 29.74 & \second{30.26} & 28.83 & \first{30.90} & 29.03 & 29.65
    \\
    & SSIM$\uparrow$
      & 0.991 & \second{0.982} & \first{0.984} & 0.981 & \first{0.984} & \second{0.982} & \first{0.984}
    \\
    & LPIPS$\downarrow$
      & 0.026 & \second{0.039} & \first{0.037} & 0.041 & \first{0.037} & 0.040 & \second{0.039}
    \\ \cmidrule(lr){1-1} \cmidrule(lr){2-2} \cmidrule(lr){3-3} \cmidrule(lr){4-9}
    \multirow{3}{*}{Cat}
    & PSNR$\uparrow$
      & 37.10 & 30.77 & \second{31.45} & 31.14 & 31.11 & 29.95 & \first{31.79}
    \\
    & SSIM$\uparrow$
      & 0.993 & \first{0.989} & \second{0.988} & \first{0.989} & \second{0.988} & 0.987 & \first{0.989}
    \\
    & LPIPS$\downarrow$
      & 0.016 & \second{0.024} & \second{0.024} & \second{0.024} & 0.026 & 0.025 & \first{0.023}
    \\ \cmidrule(lr){1-1} \cmidrule(lr){2-2} \cmidrule(lr){3-3} \cmidrule(lr){4-9}
    \multirow{3}{*}{Trophy}
    & PSNR$\uparrow$
      & 32.23 & 30.22 & \second{30.44} & 29.69 & \first{30.64} & 29.20 & 30.23
    \\
    & SSIM$\uparrow$
      & 0.965 & \second{0.962} & \first{0.963} & \second{0.962} & \first{0.963} & \second{0.962} & \first{0.963}
    \\
    & LPIPS$\downarrow$
      & 0.036 & 0.039 & \first{0.037} & 0.039 & \second{0.038} & 0.040 & 0.039
    \\ \cmidrule[0.1em](lr){1-1} \cmidrule[0.1em](lr){2-2} \cmidrule[0.1em](lr){3-3} \cmidrule[0.1em](lr){4-9}
    \multirow{3}{*}{\texttt{Average}}
    & PSNR$\uparrow$
      & 35.97 & 30.31 & \first{31.72} & 30.89 & \second{31.67} & 30.07 & 30.37
    \\
    & SSIM$\uparrow$
      & 0.989 & 0.982 & \first{0.985} & 0.983 & \second{0.984} & 0.982 & 0.983
    \\
    & LPIPS$\downarrow$
      & 0.020 & 0.029 & \first{0.026} & 0.028 & \second{0.027} & 0.029 & 0.029
    \\ \bottomrule
  \end{tabularx}
  \vspace{-2mm}
  \caption{Comparison of PSNR$\uparrow$, SSIM$\uparrow$, and LPIPS$\downarrow$ for each scene on \textbf{geometric recorrection} when the number of views in a training set was six.}
  \label{tab:comparison_geometry4_view6}
\end{table*}

\subsection{Robustness of the number of views}
\label{subsec:robustness_number}

In the main experiment (Section~\ref{sec:experiments}) and the above experiments (Appendix~\ref{subsec:robustness_selection_views}), we investigated the performance when the number of views in the training set was three.
To investigate the robustness of the number of views, we examined the performance when the number of views in the training set was changed to six.
Specifically, among the 11 views in the dataset, six were used for training, and the remaining five were used for testing.

\smallskip\noindent
\textbf{Compared models.}
We used \textit{PAC-NeRF-6v}, that is, PAC-NeRF~\cite{XLiICLR2023} trained with six views, as the baseline.
In preliminary experiments, we found that PAC-NeRF-6v demonstrated reasonable performance without the advanced techniques described in Appendix~\ref{sec:pacnerf3vdag_detail} owing to the increase in the number of views.
Therefore, we only used this model as the baseline.
We applied \textit{+LPO} and \textit{+LPO$^4$} to the baseline.
Furthermore, we investigated the performances of the ablated and comparative models.
Specifically, in the evaluation of the geometry correction, we investigated the performance when \textit{+LPO-F}, \textit{+LPO-P}, and \textit{+GO} were applied to the baseline.
In the evaluation of the physical identification and geometric recorrection, we investigated the performance when \textit{+LPO-F$^4$}, \textit{+LPO-P$^4$}, \textit{+GO$^4$}, and \textit{+None$^4$} were applied to the baseline.

\smallskip\noindent
\textbf{Results.}
Tables~\ref{tab:comparison_geometry_view6}, \ref{tab:comparison_physics_view6}, and \ref{tab:comparison_geometry4_view6} summarize the results of geometry correction, physical identification, and geometry reconstruction, respectively.
Similar tendencies were observed when the number of views in the training set was three (as discussed in Section~\ref{sec:experiments} and Appendix~\ref{subsec:robustness_selection_views}).
Specifically, with respect to geometry correction, +LPO outperformed not only the baseline (PAC-NeRF-6v) but also the ablated and comparative models, including +LPO-F, +LPO-P, and +GO, in most cases.
On physical identification and geometry recorrection, +LPO$^4$ outperformed not only the baseline (PAC-NeRF-6v) but also the ablated and comparative models, including +LPO-F$^4$, +LPO-P$^4$, +GO$^4$, and +None$^4$, in most cases.
These results indicate that the joint optimization of features and positions in Lagrangian space is effective for geometry-agnostic system identification, regardless of the number of views used for training.

\clearpage
\clearpage
\section{Qualitative results}
\label{sec:qualitative_results}

This appendix discusses the qualitative results obtained using several frames selected from the video sequences.
We provide video samples at the \href{https://www.kecl.ntt.co.jp/people/kaneko.takuhiro/projects/lpo/}{project page}.\footnoteref{foot:samples}

\subsection{Qualitative comparisons among PAC-NeRF-3v/3v$^\dag$, +LPO, and +LPO$^4$}
\label{subsec:results_lpo_lpo4}

\begin{itemize}
\item Figure~\ref{fig:results_newtonian}:\\
  Qualitative comparisons among PAC-NeRF-3v/3v$^\dag$, +LPO, and +LPO$^4$ on Newtonian fluids (Droplet and Letter).\\
\item Figure~\ref{fig:results_nonnewtonian}:\\
  Qualitative comparisons among PAC-NeRF-3v/3v$^\dag$, +LPO, and +LPO$^4$ on non-Newtonian fluids (Cream and Toothpaste).\\
\item Figure~\ref{fig:results_elasticity}:\\
  Qualitative comparisons among PAC-NeRF-3v/3v$^\dag$, +LPO, and +LPO$^4$ on elastic materials (Torus and Bird).\\
\item Figure~\ref{fig:results_plasticine}:\\
  Qualitative comparisons among PAC-NeRF-3v/3v$^\dag$, +LPO, and +LPO$^4$ on plasticine (Playdoh and Cat).\\
\item Figure~\ref{fig:results_sand}:\\
  Qualitative comparisons among PAC-NeRF-3v/3v$^\dag$, +LPO, and +LPO$^4$ on granular media (Trophy).\\
\end{itemize}

\subsection{Qualitative comparisons among +LPO$^4$, +LPO-F$^4$, +LPO-P$^4$, +GO$^4$, and +None$^4$}
\label{subsec:results_ablations}

\begin{figure*}[t]
  \begin{center}
    \includegraphics[width=\textwidth]{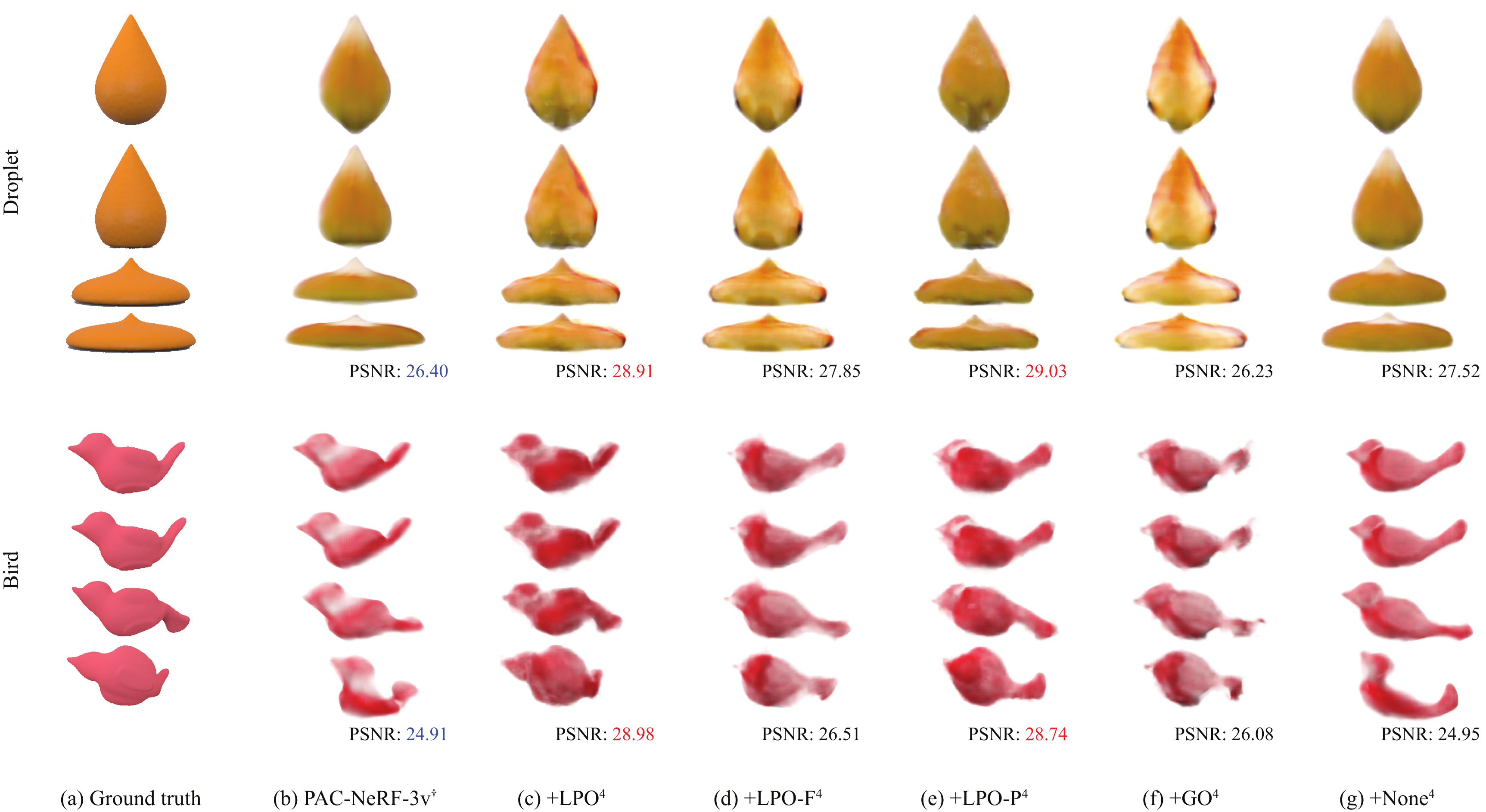}
  \end{center}
  \caption{Qualitative comparisons among +LPO$^4$, +LPO-F$^4$, +LPO-P$^4$, +GO$^4$, and +None$^4$ in the scenes where +LPO-P$^4$ outperformed +LPO-F$^4$ in terms of PSNR.
    In the +LPO-F$^4$, position (that is, shape) optimization was ablated, and only feature (that is, appearance) optimization was conducted.
    In the +LPO-P$^4$, feature (that is, appearance) optimization was ablated, and only position (that is, shape) optimization was conducted.
    In the above scenes, large geometric gaps exist between the ground truth (a) and PAC-NeRF-3v$^\dag$ (b).
    For example, in the second row of the Droplet scene, the bottom of the droplet is flat in the ground truth (a), while that is bulging in PAC-NeRF-3v$^\dag$ (b).
    As shown in (d), only appearance correction by +LPO-F$^4$ is insufficient to correct this geometry failure estimation, and the bottom of the droplet is still bulging.
    Instead, +LPO-F$^4$ attempts to solve this problem by changing the appearance, making the colors overcorrected.
    In contrast, shape correction by +LPO-P$^4$ effectively addresses this failure, and the bottom of the droplet is flat in (e).
    The same correction was also conducted in +LPO$^4$ (c), a combination of +LPO-F$^4$ and +LPO-P$^4$.
    Similarly, in the Bird scene, the directions of the tail are adequately corrected in +LPO$^4$ (c) and +LPO-P$^4$ (e).
    In contrast, those are not sufficiently conducted in +LPO-F$^4$ (d).
    Also, in this case, +LPO-F$^4$ (d) attempts to solve this problem by overcorrecting the colors.
    +GO$^4$ (f), which also only corrects appearance, has the same difficulty as +LPO-F$^4$ (d).
    In the Droplet scene, the bottom of the droplet is bulging; in the Bird scene, the bird's tail is corrupted.
    +None$^4$ (g), which performs Algorithm~\ref{alg:iterative_optimization} without geometry correction, does not have a sufficient correction ability.
    In the Droplet scene, shape and appearance are almost identical to those in PAC-NeRF-3v$^\dag$ (b).
    In the Bird scene, the pose of the bird is not corrected.}
  \label{fig:examples_ptcl}
\end{figure*}

\begin{figure*}[t]
  \begin{center}
    \includegraphics[width=\textwidth]{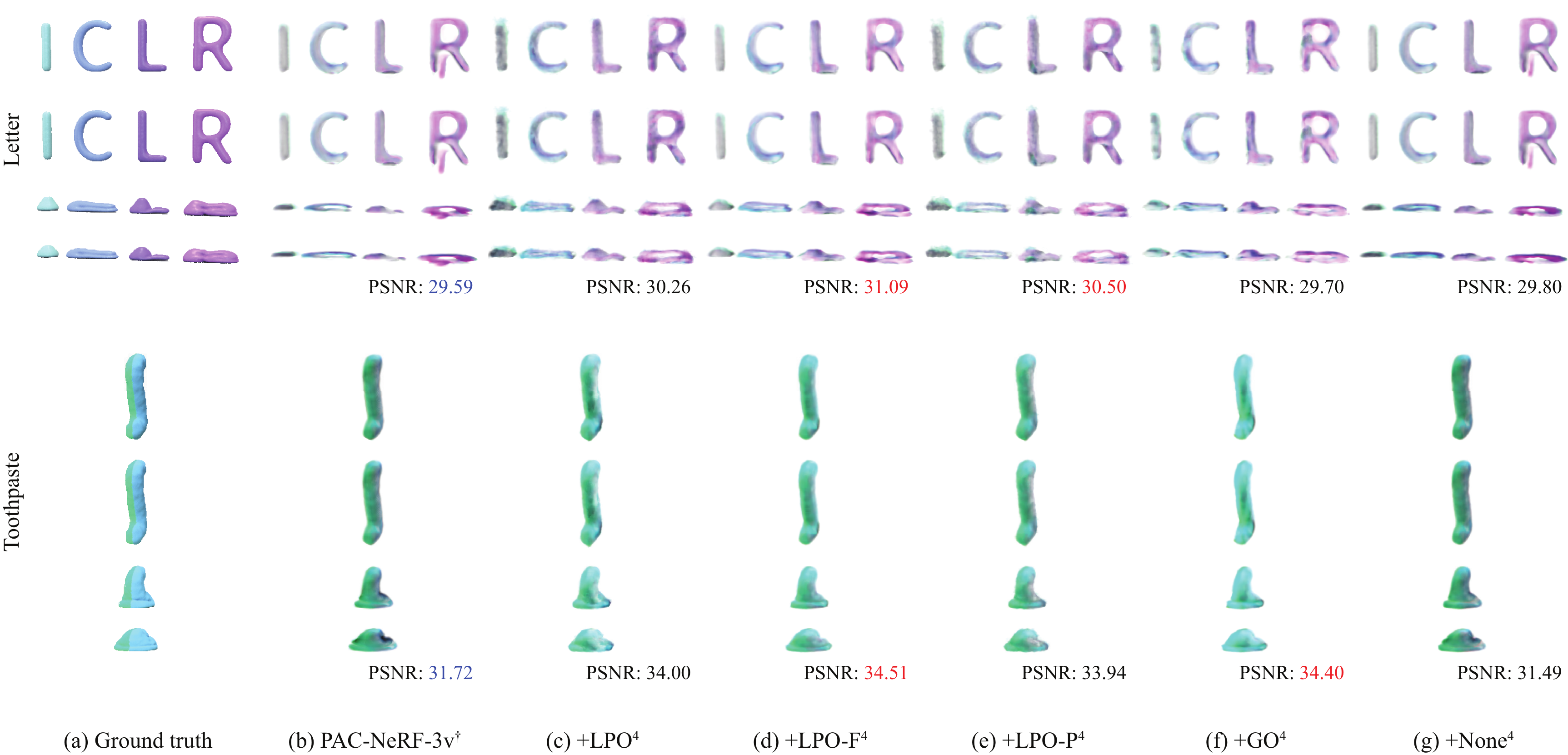}
  \end{center}
  \caption{Qualitative comparisons among +LPO$^4$, +LPO-F$^4$, +LPO-P$^4$, +GO$^4$, and +None$^4$ in the scenes where +LPO-F$^4$ outperformed +LPO-P$^4$ in terms of PSNR.
    In the +LPO-F$^4$, position (that is, shape) optimization was ablated, and only feature (that is, appearance) optimization was conducted.
    In the +LPO-P$^4$, feature (that is, appearance) optimization was ablated, and only position (that is, shape) optimization was conducted.
    As discussed in Figure~\ref{fig:examples_ptcl}, +LPO-F$^4$ is unsuitable for significantly correcting the geometric shape because it focuses on correcting appearance.
    However, as shown in the above scenes, when PAC-NeRF-3v$^\dag$ (b) captures the geometric structure relatively well, +LPO-F$^4$ (d) also works well because appearance correction is more important than shape correction.
    For example, in the Letter scene, +LPO-F$^4$ (d) succeeds in eliminating artifacts existing in the vicinity of the left line of the letter ``R.''
    In the Toothpaste scene, +LPO-F$^4$ (d) succeeds in making the color of the material brighter and closer to the ground truth (a).
    Notably, in both scenes, +None$^4$ (g) fails to do so and produces almost the same results as those in PAC-NeRF-3v$^\dag$ (b).
    These results indicate that appearance correction by +LPO-F$^4$ is essential (simple iterative updates in Algorithm~\ref{alg:iterative_optimization} are insufficient) to address these problems.
    Another interesting finding is that +LPO-P$^4$ (e), which focuses on correcting shape, also works well.
    For example, in the Letter scene, +LPO-P$^4$ (e) eliminates the artifacts existing in the vicinity of the left line of the letter ``R,'' and in the Toothpaste scene, it makes the material brighter.
    This is possible because moving correctly colored particles from other places allows for appearance changes.}
  \label{fig:examples_feat}
\end{figure*}

\begin{itemize}
\item Figure~\ref{fig:examples_ptcl}:\\
  Qualitative comparisons among +LPO$^4$, +LPO-F$^4$, +LPO-P$^4$, +GO$^4$, and +None$^4$ in the scenes where +LPO-P$^4$ outperformed +LPO-F$^4$.\\
\item Figure~\ref{fig:examples_feat}:\\
  Qualitative comparisons among +LPO$^4$, +LPO-F$^4$, +LPO-P$^4$, +GO$^4$, and +None$^4$ in the scenes where +LPO-F$^4$ outperformed +LPO-P$^4$.\\
\end{itemize}

\clearpage
\clearpage
\section{Implementation details}
\label{sec:implementation}

\subsection{Dataset}
\label{subsec:dataset}

We investigated the benchmark performance on the dataset provided by the original study on PAC-NeRF~\cite{XLiICLR2023}.
This dataset comprised nine scenes and various continuum materials, including the following:
\smallskip
\begin{itemize}
\item Newtonian fluids with fluid viscosity $\mu$ and bulk modulus $\kappa$:
  \begin{itemize}
  \item \textit{Droplet} with $\mu = 200$ and $\kappa = 10^5$
  \item \textit{Letter} with $\mu = 100$ and $\kappa = 10^5$
  \end{itemize}
\item Non-Newtonian fluids with shear modulus $\nu$, bulk modulus $\kappa$, yield stress $\tau_Y$, and plasticity viscosity $\eta$:
  \begin{itemize}
  \item \textit{Cream} with $\mu = 10^4$, $\kappa = 10^6$, $\tau_Y = 3 \times 10^3$, and $\eta = 10$
  \item \textit{Toothpaste} with $\mu = 5 \times 10^3$, $\kappa = 10^5$, $\tau_Y = 200$, and $\eta = 10$
  \end{itemize}
\item Elastic materials with Young's modulus $E$ and Poisson's ratio $\nu$:
  \begin{itemize}
  \item \textit{Torus} with $E = 10^6$ and $\nu = 0.3$
  \item \textit{Bird} with $E = 3 \times 10^5$ and $\nu = 0.3$
  \end{itemize}
\item Plasticine with Young's modulus $E$, Poisson's ratio $\nu$, and yield stress $\tau_Y$:
  \begin{itemize}
  \item \textit{Playdoh} with $E = 2 \times 10^6$, $\nu = 0.3$, and $\tau_Y = 1.54 \times 10^4$
  \item \textit{Cat} with $E = 10^6$, $\nu = 0.3$, and $\tau_Y = 3.85 \times 10^3$
  \end{itemize}
\item Granular media with friction angle $\theta_{fric}$:
  \begin{itemize}
  \item \textit{Trophy} with $\theta_{fric} = 40^\circ$
  \end{itemize}
\end{itemize}

\smallskip
In each scene, the objects fall freely under the influence of gravity and undergo collisions.
The ground-truth simulation data were generated using the MLS-MPM framework~\cite{YHuTOG2018}.
A photorealistic simulation engine rendered objects under diverse environmental lighting conditions and ground textures.
Each scene was captured from 11 viewpoints with cameras evenly spaced on the upper hemisphere, including the object.
To evaluate our method in sparse-view settings, three views were used for training, and the remaining eight views were used for testing in the main experiments (Section~\ref{sec:experiments}) and the experiments described in Appendix~\ref{subsec:robustness_selection_views}.
Six views were used for training, and the remaining five were used for testing in the experiments described in Appendix~\ref{subsec:robustness_number}.
Data were downloaded from the website\footnote{\label{foot:pacnerf_code}\url{https://github.com/xuan-li/PAC-NeRF}} provided by the authors of PAC-NeRF~\cite{XLiICLR2023}.

\subsection{Model}
\label{subsec:model}

The models were implemented based on the official PAC-NeRF source code.\footnoteref{foot:pacnerf_code}
For simplicity and fair comparison, we used the default model configurations provided in the source code for the experiments.
Specifically, the architecture of a discretized voxel-based NeRF followed direct voxel grid optimization~\cite{CSunCVPR2022}, in which a volume density field and color field were represented by voxel grids, and a 2-layer MLP with a hidden dimension of $128$ was applied to the color grid with positional embedding for a view direction $\mathbf{d}$.
Regarding a differentiable MPM, DiffTaichi~\cite{YHuICLR2020} was used.

\subsection{Training settings}
\label{subsec:training}

For simplicity and fair comparison, we conducted Eulerian static voxel grid optimization (Figure~\ref{fig:pipelines}(1)) and physical property optimization (Figure~\ref{fig:pipelines}(2)) using the default training settings provided in the source code,\footnoteref{foot:pacnerf_code} in PAC-NeRF, PAC-NeRF-3v, and PAC-NeRF-6v, except that the number of views was changed.
In PAC-NeRF-3v$^\dag$, we applied the three modifications described in Appendix~\ref{sec:pacnerf3vdag_detail} (scheduling of $\mathcal{L}_{surf}$, introduction of $\mathcal{L}_{pixel}^{VI}$, and adjustment of training length) to the Eulerian static voxel grid optimization and adopted the default training settings for physical property optimization.
For the LPO (Figure~\ref{fig:pipelines}(3)), we trained the features and positions of the particles for 100 iterations using the Adam optimizer~\cite{DPKingmaICLR2015}.
In particular, we optimized the features of the particles at a learning rate of $0.1$, which is the default value used for training the feature grids in the Eulerian static voxel grid optimization.
The positions of the particles were optimized at a learning rate of $\frac{\mathrm{d}x}{32}$, where $\mathrm{d}x$ indicates the voxel grid size and differs depending on the scene (set in the configuration files in the source code\footnoteref{foot:pacnerf_code}).
In a preliminary experiment, we found that careful setting of this learning rate is vital for stable training because the particles can diverge when the learning rate is exceptionally high.
Based on this observation, we used a learning rate adjusted according to $\mathrm{d}x$.
We set the momentum terms of the Adam optimizer, $\beta_1$ and $\beta_2$, to $0.9$ and $0.999$, respectively.

\subsection{Evaluation metrics}
\label{subsec:evaluation}

\textbf{Evaluation of geometric (re)correction.}
We evaluated the performance of the geometric (re)correction using metrics commonly used to assess the performance of novel view synthesis in NeRF studies: the peak signal-to-noise ratio (\textit{PSNR}), structural similarity index measure (\textit{SSIM})~\cite{ZWangTIP2004}, and learned perceptual image patch similarity (\textit{LPIPS})~\cite{RZhangCVPR2018}.
For PSNR and SSIM, the larger the values, the better the performance.
For LPIPS, the smaller the values, the better the performance.
In particular, we report the scores averaged over the video sequences in a test set.

\smallskip\noindent
\textbf{Evaluation of physical identification.}
To evaluate the performance of the physical identification, we measured the absolute distance between the ground truth and the estimated physical properties.
The values of the ground-truth physical properties are provided in Appendix~\ref{subsec:dataset}.
For an easy comparison, we calculated these distances after adjusting the scale (i.e., either a logarithmic scale or a linear scale) following the study of PAC-NeRF~\cite{XLiICLR2023}.
The smaller the values, the better the performance.

\end{document}